# Post-Regularization Inference for Time-Varying Nonparanormal Graphical Models

**Junwei Lu** JUNWEIL@PRINCETON.EDU
*Department of Operations Research and Financial Engineering*
*Princeton University*
*Princeton, NJ 08544, USA*

**Mladen Kolar** MKOLAR@CHICAGOBOOTH.EDU
*Booth School of Business*
*The University of Chicago*
*Chicago, IL 60637, USA*

**Han Liu** HANLIU@PRINCETON.EDU
*Department of Operations Research and Financial Engineering*
*Princeton University*
*Princeton, NJ 08544, USA*

## Abstract

We propose a novel class of time-varying nonparanormal graphical models, which allows us to model high dimensional heavy-tailed systems and the evolution of their latent network structures. Under this model we develop statistical tests for presence of edges both locally at a fixed index value and globally over a range of values. The tests are developed for a high-dimensional regime, are robust to model selection mistakes and do not require commonly assumed minimum signal strength. The testing procedures are based on a high dimensional, debiasing-free moment estimator, which uses a novel kernel smoothed Kendall's tau correlation matrix as an input statistic. The estimator consistently estimates the latent inverse Pearson correlation matrix uniformly in both the index variable and kernel bandwidth. Its rate of convergence is shown to be minimax optimal. Our method is supported by thorough numerical simulations and an application to a neural imaging data set.

## 1. Introduction

We consider the problem of inferring time-varying undirected graphical models from high dimensional non-Gaussian distributions. Undirected graphical models have been widely used as a powerful tool for exploring the dependency relationships between variables. We are interested in graphical models which have non-static graphical structures and can handle heavy-tail distributions as well as data contaminated with outliers. To that end, we develop a class of time-varying nonparanormal models, which can be used to explore Markov dependencies of a random vector $\boldsymbol{X}$ given the index variable $Z$. Specifically, we assume the random variables $(\boldsymbol{X}, Z)$ follow the following joint distribution: the conditional distribution of $\boldsymbol{X} \mid Z = z$ follows a nonparanormal distribution

$$\boldsymbol{X} \mid Z = z \sim \mathsf{NPN}_d(\mathbf{0}, \boldsymbol{\Sigma}(z), f) \tag{1}$$

where $f = \{f_1, \ldots, f_d\}$ is a set of $d$ univariate, strictly increasing functions and $Z$ is a random variable with a continuous density. A variable follows a nonparanormal distribution $\boldsymbol{Y} \sim \mathsf{NPN}_d(\boldsymbol{\mu}, \boldsymbol{\Sigma}, f)$ if $f(\boldsymbol{Y}) \sim N(\boldsymbol{\mu}, \boldsymbol{\Sigma})$ (Liu et al., 2009). Graphical modeling with nonparanormal distribution is studied in Liu et al. (2009), Liu et al. (2012a) and Xue and Zou (2012), however, their graph structure is static. Time-varying graphical models are studied in Talih and Hengartner (2005), Xuan and Murphy (2007), Zhou et al. (2010), Kolar and Xing (2011), Kolar and Xing (2012), Yin et al. (2010), Kolar et al. (2010b), Ahmed and Xing (2009), Kolar et al. (2010a) and Kolar and Xing (2009). However, these papers assume that conditionally on the index, the distribution of $\boldsymbol{X}$ is parametric, which is not adequate for applications to heavy-tailed data sets in finance, neuroscience and genomics (Qiu et al., 2016). Moreover, inferential methods for the time-varying graphical models have not been developed so far.

Primary motivation for proposing the model in (1) and developing corresponding estimation and inferential procedures comes from an application in neuroscience. Graphical models are widely used to estimate and explore functional connectivity between different brain regions from functional magnetic resonance imaging (fMRI) data (Wang et al., 2010; Bullmore and Bassett, 2011; Smith et al., 2011). There is evidence that that the brain connectivity network evolves over time (Bartzokis et al., 2001; Gelfand et al., 2003) and current techniques are not adequate for capturing evolving nature of brain networks. For example, work of Kolar et al. (2010a) assumes that data are Gaussian, which is rarely satisfied in practice. Qiu et al. (2016) need replicated observations at each time point, which are not available in most of the time-varying fMRI data sets. Furthermore, current procedures are solely focused on estimation of networks, while the question of inference and quantification of uncertainty is left unanswered. We address these drawbacks in the current work.

The focus of the paper is on the inferential analysis about parameters in the model given in (1), as well as the Markov dependencies between observed variables. The inference procedures we develop are uniformly valid in a high-dimensional regime and robust to model selection mistakes. In particular, the inference does not depend on the oracle support recovery properties of the estimator. As a foundation for inference, we develop an estimation procedure for the high-dimensional latent time-varying inverse correlation matrix based on a novel kernel smoothed Kendall's tau statistic. The estimator is uniformly consistent in both the bandwidth and the index variable, and furthermore is optimal in a minimax sense. Obtaining rate of convergence for the estimator is technically challenging and requires development of new uniform bounds for the $U$-processes, careful characterization of the leading terms in the expansion of the estimator in the presence of high-dimensionality and approximation errors arising from the local approximation of nonlinear curves. The proof that the rate of convergence is optimal involves application of Le Cam's method on a carefully chosen function valued high dimensional matrices class. These technical details are novel and of independent interest, as discussed in Section 4.2.

We consider three types of hypothesis tests: (1) the edge presence test for whether there is an edge in the Markov graph at $z_0$, (2) the super graph test for whether the true graph is a subgraph of a fixed graph at $z_0$, and (3) the uniform edge presence test for whether the true graph is a subgraph of a given graph over a range of index values. The first test was studied in the context of static Gaussian graphical models in Janková and van de Geer (2015, 2017) and Ren et al. (2015), however, their approach cannot handle time-varying models. The second test was considered in Wasserman et al. (2014), Yang et al. (2014b) and Gu et al. (2015) also in the context of static graphical models. Cai et al. (2013b) considered a statistical test for whether the correlation matrix is an identity, which is a special case of the super graph test. The third test is a generalization of the above two local tests to a global test over a range of index values, which allows for identifying whether certain connections in graphical models exist for a period of time. We illustrate the super graph test in an application to the ADHD-200 data set containing fMRI data from subjects with and without attention deficit hyperactive disorder (ADHD) (Biswal et al., 2010), which allows us to uncover how the brain networks change with age.

This paper makes two major contributions to the literature on statistical inference for graphical models. First, we develop a general inferential procedure for a wide family of high dimensional graphical model estimation methods. Many existing high dimensional inference methods are specifically designed for concrete estimators. For example, Zhang and Zhang (2013), van de Geer et al. (2014), Javanmard and Montanari (2014), and Ning and Liu (2017) design inference procedures for specific M-estimators, while Neykov et al. (2015) developed an inferential procedure for the method of moments estimators like Dantzig selector (Candés and Tao, 2007) and CLIME (Cai et al., 2011). Barber and Kolar (2018) design a procedure for constructing confidence intervals in high-dimensional elliptical copula models. In contrast to that, we propose a nonparametric score-type statistic, which uses any estimator of $\mathbf{\Sigma}(z)^{-1}$ with fast enough rate of convergence as an input. Therefore, our inference procedure does not depend on a particular estimate of $\mathbf{\Sigma}(z)^{-1}$ and can be applied to both $M$-estimators, like graphical Lasso, and method of moments estimators, like CLIME. Second, to the best of our knowledge, this paper considers for the first time presence of the edges test uniformly over the index $z$ for high dimensional graphical models. Computing quantiles of the test statistic requires development a new Gaussian multiplier bootstrap procedure for a $U$-process.

### 1.1 Related Literature

High-dimensional Gaussian graphical models are studied in Meinshausen and Bühlmann (2006), Yuan and Lin (2007), Rothman et al. (2008), Friedman et al. (2008), d'Aspremont et al. (2008), Fan et al. (2009), Lam and Fan (2009), Yuan (2010), Cai et al. (2011), Liu and Wang (2017), and Zhao and Liu (2014), with an extension to covariate-adjusted graphical models given in Dondelinger et al. (2010), Li et al. (2012), Cai et al. (2013a), Chen et al. (2016), and Yin and Li (2013). Semiparametric extensions using copulas are developed in Liu et al. (2009), Liu et al. (2012a), Xue and Zou (2012), and Liu et al. (2012b), and extended for mixed data in Fan et al. (2015). Guo et al. (2011a), Guo et al. (2011b), Lee and Hastie (2015), Cheng et al. (2017), Yang et al. (2012), and Yang et al. (2014a) study the mixed exponential family graphical models where a node conditional distribution is a member of an exponential family distribution. Danaher et al. (2014), Qiu et al. (2016), Mohan et al. (2014) consider joint estimation of multiple graphical models.

All of the above mentioned work assumes that the graphical structure is static. However, in analysis of many complex systems, such an assumption is not valid. There are two major types of time-varying graphical model: directed and undirected. The directed time-varying graphical models are mainly studied in the context of autoregressive models with time-varying parameters (Punskaya et al., 2002; Fujita et al., 2007; Rao et al., 2007; Grzegorczyk and Husmeier, 2011; Song et al., 2009; Robinson and Hartemink, 2010; Jia and Huan, 2010; Lébre et al., 2010; Husmeier et al., 2010; Wang et al., 2011; Grzegorczyk and Husmeier, 2012; Dondelinger et al., 2013; Lébre et al., 2010). For the time-varying undirected graphical models, Zhou et al. (2010), Kolar and Xing (2011), Yin et al. (2010), Kolar et al. (2010b) and Kolar et al. (2010a) consider the kernel- smoothed type estimator for graphical models. Kolar and Xing (2012) assume the graphical model evolves in a piecewise-constant fashion and estimate it by the temporally smoothed $\ell_1$ penalized regression. Talih and Hengartner (2005) and Xuan and Murphy (2007) consider a Bayesian framework to model the time-varying of graphs and estimate the graph by Markov chain Monte Carlo methods. All the above works show the statistical rates for graphical models for fixed time points. Our work contributes to this literature by studying the uniform properties of estimators and developing inferential procedures. For estimation, we prove a rate of convergence that is uniform over $z$ and show it matches the minimax rate. Instead of the kernel smoothed sample covariance, we propose the kernel smoothed $U$-statistics as a robust estimator. For inference, we study the presence of edges for a range of index values instead of the local tests in the literature.

Our paper also contributes to the literature on high dimensional inference. Hypothesis testing and confidence intervals for the high dimensional M-estimators are studied in Zhang and Zhang (2013), van de Geer et al. (2014); Javanmard and Montanari (2014), Belloni et al. (2013), Belloni et al. (2016), Javanmard and Montanari (2014) and Meinshausen (2015). Lu et al. (2015) considered the confidence bands for the high dimensional nonparametric models. Neykov et al. (2015) proposed the inference for high dimensional method of moments estimators. Lee et al. (2016) and Tibshirani et al. (2016) consider the conditional inference based on post-selection methods. Our work considers a new inferential framework involving both discrete graph structures and the nonparametric index variable, which provides more flexibility in the modeling of modern data sets.

Finally, we develop novel probabilistic tools to study the high dimensional $U$-statistics. Classical analysis for fixed dimensional $U$-statistics is built on the Hoeffding decomposition (Hoeffding, 1948). Concentration inequalities for high dimensional $U$-statistics are studied in Giné et al. (2000) and Adamczak (2006). However, existing methods based on uniform entropy numbers are too loose to be applicable (Nolan and Pollard, 1987). We develop a new peeling method to control suprema of our kernel smoothed $U$-process uniformly over three aspects: dimension, index variable and bandwidth. The uniform consistency over the bandwidth shown in the paper also generalizes a data-driven bandwidth-tuning method to $U$-statistics from the kernel-type estimator considered in Einmahl and Mason (2005). This provides more flexibility in the tuning procedure of our method. Moreover, to study the limiting distribution of the $U$-statistics, we generalize the Gaussian multiplier bootstrap proposed in Chernozhukov et al. (2013) and Chernozhukov et al. (2014a) to nonparametric $U$-process by considering a new type of nonlinear Gaussian multiplier $U$-processes.

### 1.2 Notation

Let $[n]$ denote the set $\{1, \ldots, n\}$ and let $\mathbb{1}\{\cdot\}$ denote the indicator function. For a vector $a \in \mathbb{R}^d$, we let $\text{supp}(a) = \{j \, : \, a_j \neq 0\}$ be the support set (with an analogous definition for matrices $A \in \mathbb{R}^{n_1 \times n_2}$), $||a||_q$, $q \in [1, \infty)$, the $\ell_q$-norm defined as $||a||_q = (\sum_{i \in [n]} |a_i|^q)^{1/q}$ with the usual extensions for $q \in \{0, \infty\}$, that is, $||\mathbf{a}||_0 = |\text{supp}(\mathbf{a})|$ and $||\mathbf{a}||_\infty = \max_{i \in [n]} |a_i|$. For a matrix $\mathbf{A} \in \mathbb{R}^{n_1 \times n_2}$, we use the notation $\text{vec}(A)$ to denote the vector in $\mathbb{R}^{n_1 n_2}$ formed by stacking the columns of $\mathbf{A}$. We write $\mathbf{A} = [\mathbf{A}_{jk}]$ if the $(j,k)$-th entry of $\mathbf{A}$ is $\mathbf{A}_{jk}$. Let $\boldsymbol{A}_{\setminus j,k}$ be the sub-vector of the $k$-th column $\mathbf{A}$ with $\mathbf{A}_{jk}$ removed. We denote the Frobenius norm of $\mathbf{A}$ by $||\mathbf{A}||_F^2 = \sum_{i \in [n_1], j \in [n_2]} \mathbf{A}_{ij}^2$, the max-norm $\|\mathbf{A}\|_{\max} = \max_{i \in [n_1], j \in [n_2]} |\mathbf{A}_{ij}|$, the $\ell_1$-norm $\|\mathbf{A}\|_1 = \max_{j \in [n_2]} \sum_{i \in [n_1]} |\mathbf{A}_{ij}|$, and the operator norm $||\mathbf{A}||_2 = \sup_{\|\mathbf{v}\|_2 = 1} \|\mathbf{A}\mathbf{v}\|_2$. The Hadamard product of two matrices is the matrix $\mathbf{A} \circ \mathbf{B}$ with elements $(\mathbf{A} \circ \mathbf{B})_{jk} = \mathbf{A}_{jk} \cdot \mathbf{B}_{jk}$. Given two functions $f$ and $g$, we denote their convolution as $(f * g)(t) = \int f(t-x) g(x) dx$. For $1 \leq p < \infty$, let $\|f\|_p = (\int f^p)^{1/p}$ denote the $L^p$-norm of $f$ and $\|f\|_\infty = \sup_x |f(x)|$. The total variation of $f$ is defined as $\text{TV}(f) = \|f'\|_1$. For two sequences of numbers $\{a_n\}_{n=1}^\infty$ and $\{b_n\}_{n=1}^\infty$, we use $a_n = O(b_n)$ or $a_n \lesssim b_n$ to denote that $a_n \leq C b_n$ for some finite positive constant $C$, and for all $n$ large enough. If $a_n = O(b_n)$ and $b_n = O(a_n)$, we use the notation $a_n \asymp b_n$. The notation $a_n = o(b_n)$ is used to denote that $a_n / b_n \to 0$ as $n$ goes infinity. We also define $a \vee b = \max(a, b)$ and $a \wedge b = \min(a, b)$ for any two scalars $a$ and $b$. We use $\xrightarrow{P}$ for convergence in probability and $\rightsquigarrow$ for convergence in distribution. Throughout the paper, we let $c, C$ be two generic absolute constants, whose values will vary at different locations.

We use the notation $\mathbb{E}_n[\cdot]$ to denote the empirical average, $\mathbb{E}_n[f] = n^{-1} \sum_{i \in [n]} f(X_i)$. We also use $\mathbb{G}_n[f] = \sqrt{n} \left( \mathbb{E}_n[f(X_i)] - \mathbb{E}[f(X_i)] \right) = n^{-1/2} \sum_{i \in [n]} \left( f(X_i) - \mathbb{E}[f(X_i)] \right)$. For a bivariate function $H(x, x')$, we define the $U$-statistic $\mathbb{U}_n[H] = [n(n-1)]^{-1} \sum_{i \neq i'} H(X_i, X_{i'})$.

Appendix J collects all the notation in a table format with reference to where they appear first.

## 2. Preliminaries

We start by providing background on the nonparanormal distribution and discuss how it relates to the time-varying nonparanormal graphical model in (1). The nonparanormal distribution was introduced in Liu et al. (2009). A random variable $\boldsymbol{X} = (X_1, \ldots, X_d)^T$ is said to follow a nonparanormal distribution if there exists a set of monotone univariate functions $f = \{f_1, \ldots, f_d\}$ such that $f(\boldsymbol{X}) := (f_1(X_1), \ldots, f_d(X_d))^T \sim N(\mathbf{0}, \boldsymbol{\Sigma})$, where $\boldsymbol{\Sigma}$ is a latent correlation matrix satisfying $\text{diag}(\boldsymbol{\Sigma}) = 1$. We denote $\boldsymbol{X} \sim \mathsf{NPN}_d(\mathbf{0}, \boldsymbol{\Sigma}, f)$.

Given $n$ independent copies of $\boldsymbol{X} \sim \mathsf{NPN}_d(\mathbf{0}, \boldsymbol{\Sigma}, f)$, Liu et al. (2012a) study how to estimate the latent correlation matrix $\boldsymbol{\Sigma}$. The key idea lies in relating the Kendall's tau correlation matrix with the Pearson correlation. The Kendall's tau correlation between $X_j$ and $X_k$, two coordinates of $\boldsymbol{X}$, is defined as

$$\tau_{jk} = \mathbb{E} \left[ \text{sign} \left( (X_j - \widetilde{X}_j)(X_k - \widetilde{X}_k) \right) \right],$$

where $(\widetilde{X}_j, \widetilde{X}_k)$ is an independent copy of $(X_j, X_k)$. It can be related to the latent correlation matrix using the fact that $\tau_{jk} = (2/\pi) \arcsin \left( \boldsymbol{\Sigma}_{jk} \right)$ when $\boldsymbol{X}$ follows a nonparanormal distribution (Fang et al., 1990). The inverse covariance matrix $\boldsymbol{\Omega} = \boldsymbol{\Sigma}^{-1}$ encodes the graph structure of a nonparanormal distribution (Liu et al., 2009). Specifically $\mathbf{\Omega}_{jk} = 0$ if and only if $X_j$ is independent of $X_k$ conditionally on $\boldsymbol{X}_{\setminus\{j,k\}}$.

The above observations lead naturally to the following estimation procedure for $\mathbf{\Omega}$. We estimate the Kendall's tau correlation matrix $\widehat{\mathbf{T}} = [\widehat{\tau}_{jk}] \in \mathbb{R}^{d\times d}$ elementwise using the following $U$-statistic

$$\widehat{\tau}_{jk} = \frac{2}{n(n-1)} \sum_{1\le i<i'\le n} \operatorname{sign}(X_{ij} - X_{i'j}) \operatorname{sign}(X_{ik} - X_{i'k}).$$

An estimate of the latent correlation matrix is given as $\widehat{\mathbf{\Sigma}} = \sin\left(\pi\widehat{\mathbf{T}}/2\right)$, where $\sin(\cdot)$ is applied elementwise. Finally, the estimate of the latent correlation matrix $\widehat{\mathbf{\Sigma}}$ is used as a plug-in statistic in the CLIME estimator (Cai et al., 2011), or calibrated CLIME estimator (Zhao and Liu, 2014), to obtain the inverse covariance estimator $\widehat{\mathbf{\Omega}}$.

The CLIME estimator solves the following optimization program

$$\widehat{\mathbf{\Omega}}_j^{\text{CLIME}} = \operatorname*{argmin}_{\boldsymbol{\beta}\in\mathbb{R}^d} \|\boldsymbol{\beta}\|_1 \qquad \text{subject to} \qquad \|\widehat{\mathbf{\Sigma}}\boldsymbol{\beta} - \mathbf{e}_j\|_\infty \le \lambda, \tag{2}$$

where $\mathbf{e}_j$ is the $j$-th canonical basis in $\mathbb{R}^d$ and the penalty parameter $\lambda$ that controls the sparsity of the resulting estimator is commonly chosen as $\lambda \asymp \|\mathbf{\Omega}\|_1\sqrt{\log d/n}$ (Cai et al., 2011). Note that the tuning parameter depends on the unknown $\mathbf{\Omega}$ through $\|\mathbf{\Omega}\|_1$, which makes practical selection of $\lambda$ difficult. The calibrated CLIME is a tuning-insensitive estimator, which alleviates this problem. The calibrated CLIME estimator solves

$$\left(\widehat{\mathbf{\Omega}}_j^{\text{CCLIME}}, \widehat{\kappa}_j\right) = \operatorname*{argmin}_{\boldsymbol{\beta}\in\mathbb{R}^d,\kappa\in\mathbb{R}} \|\boldsymbol{\beta}\|_1 + \gamma\kappa \qquad \text{subject to} \qquad \|\widehat{\mathbf{\Sigma}}\boldsymbol{\beta} - \mathbf{e}_j\|_\infty \le \lambda\kappa,\ \|\boldsymbol{\beta}\|_1 \le \kappa, \tag{3}$$

where $\gamma$ is any constant in $(0, 1)$ and the tuning parameter can be chosen as $\lambda = C\sqrt{\log d/n}$ with $C$ being a universal constant independent of the problem parameters. In what follows, we will adapt the calibrated clime to estimation of the parameters of the model in (1).

### 2.1 Time-Varying Nonparanormal Graphical Model

The time-varying nonparanormal graphical model in (1) is an extension of the nonparanormal distribution. For every fixed value of the index variable $Z = z$, we have a static nonparanormal distribution $\boldsymbol{X} \mid Z = z \sim \mathsf{NPN}_d(\mathbf{0}, \mathbf{\Sigma}(z), f)$ that can be easily interpreted. However, as the index variable changes, the conditional distribution of $\boldsymbol{X} \mid Z$ can change in an unspecified way. In this sense, time-varying nonparanormal graphical models extend nonparanormal graphical models in the same way varying coefficient models extend linear regression models.

Let $Y = (\boldsymbol{X}, Z)$ denote a random pair distributed according to the time-varying nonparanormal distribution. Specifically $Z \sim f_Z(z)$ with $f_Z(\cdot)$ being a continuous density function supported on $[0, 1]$ and $\boldsymbol{X} \mid Z = z \sim \mathsf{NPN}_d(\mathbf{0}, \mathbf{\Sigma}(z), f)$ for all $z \in [0, 1]$. For any fixed $z \in [0, 1]$, we denote the inverse of the correlation matrix as $\mathbf{\Omega}(z) = \mathbf{\Sigma}^{-1}(z)$. Both $f_Z(z)$ and each entry of $\mathbf{\Omega}(z)$ are second-order differentiable (we will formalize assumptions in Section 4). We denote the undirected graph encoding the conditional independence of $\boldsymbol{X} \mid Z = z$ as $G^*(z) = (V, E^*(z))$, with $(j, k) \in E^*(z)$ when $\mathbf{\Omega}_{jk}(z) \ne 0$. As in the static case, we relate the Kendall's tau correlation matrix with the latent correlation matrix. Let $\mathbf{T}(z) = [\tau_{jk}(z)]_{jk}$ be the Kendall's tau correlation matrix corresponding to $\boldsymbol{X} \mid Z = z$ with elements defined as

$$\tau_{jk}(z) = \mathbb{E}\left[\operatorname{sign}\left((X_j - \widetilde{X}_j)(X_k - \widetilde{X}_k)\right) \mid Z = z\right],$$

where $\widetilde{\boldsymbol{X}}$ is an independent copy of $\boldsymbol{X}$ conditionally on $Z = z$. Given $n$ independent copies of $Y = (\boldsymbol{X}, Z)$, $\{Y_i = (\boldsymbol{X}_i, Z_i)\}_{i\in[n]}$, we estimate an element of the Kendall's tau correlation matrix using the following kernel estimator

$$\widehat{\tau}_{jk}(z) = \frac{n^{-2}\sum_{i<i'} \omega_z(Z_i, Z_{i'}) \operatorname{sign}(X_{ij} - X_{i'j}) \operatorname{sign}(X_{ik} - X_{i'k})}{n^{-2}\sum_{i<i'} \omega_z(Z_i, Z_{i'})}, \text{ where} \tag{4}$$

$$\omega_z(Z_i, Z_{i'}) = K_h\left(Z_i - z\right) K_h\left(Z_{i'} - z\right) \tag{5}$$

with the kernel function $K(\cdot)$ being a symmetric density function, $K_h(\cdot) = h^{-1}K(\cdot/h)$ and $h > 0$ is the bandwidth parameter. We can choose the kernel function as long as it satisfies some regularity conditions, which will be specified in Assumption 4.3. The kernel $U$-statistic in (4) is a generalization of classical kernel regression (Opsomer and Ruppert, 1997; Fan and Jiang, 2005). For example, given i.i.d. samples $\{Y_i, Z_i\}_{i=1}^n$ from the model $Y = f(Z) + \epsilon$, the Nadaraya-Watson estimator (Bierens, 1988) is

$$\widehat{f}(z) = \frac{n^{-1}\sum_{i=1}^n K_h(Z_i - z)Y_i}{n^{-1}\sum_{i=1}^n K_h(Z_i - z)}, \tag{6}$$

where we take the weighted average of $Y_i$'s and the weight $K_h(Z_i - z)$ is related to the distance between $Z_i$ and $z$. In order to normalize the weights, we add the denominator in (6), which is the kernel density estimator for the density of $Z$. Comparing (4) with the Nadaraya-Watson estimator, since the kernel $U$-statistic involves both $\boldsymbol{X}_i$ and $\boldsymbol{X}_{i'}$, we need to multiply the weights as (5) to ensure that both $Z_i$ and $Z_{i'}$ are in the neighborhood of $z$. We also normalize the weights by dividing the denominator $n^{-2}\sum_{i<i'} \omega_z(Z_i, Z_{i'})$ which is the estimator of $f_Z^2(z)$. The denominator $n^{-2}\sum_{i<i'} \omega_z(Z_i, Z_{i'})$ is also related to the density $f_Z(z)$. In fact, it is the estimator of $f_Z^2(z)$. Intuitively, we can see it from

$$\mathbb{E}\Big[\frac{1}{n^2}\sum_{i<i'} \omega_z(Z_i, Z_{i'})\Big] = \iint K(t_1)K(t_2)f_Z(z + t_1 h)f_Z(z + t_2 h)dt_1 dt_2 = f_Z^2(z) + O(h^2).$$

Lemma 14 provides the details and is given in the supplementary material.

Based on the above estimator, we obtain the corresponding latent correlation matrix for any index value $z \in (0, 1)$ as

$$\widehat{\boldsymbol{\Sigma}}(z) = \sin\left(\frac{\pi}{2}\widehat{\mathbf{T}}(z)\right), \tag{7}$$

where $\widehat{\mathbf{T}}(z) = [\widehat{\tau}_{jk}(z)] \in \mathbb{R}^{d\times d}$.

Finally, similar to the static case, we can plug $\widehat{\boldsymbol{\Sigma}}(z)$ into a procedure that gives an estimate of the inverse correlation matrix, such as the CLIME in (2) or calibrated CLIME in (3). Due to practical advantages of the calibrated CLIME, we will use it for our simulations. However, at this point we note that the inferential framework only requires the estimator of the inverse correlation matrix to converge at a fast enough rate. Therefore, in what follows, we denote $\widehat{\boldsymbol{\Omega}}(z)$ a generic estimator of $\boldsymbol{\Omega}(z)$. Concrete statistical properties required of the calibrated CLIME will be discussed in details in Section 4.2.

## 3. Inferential Methods

In this section, we develop a framework for statistical inference about the parameters in a time-varying nonparanormal graphical model. We focus on the following three testing problems:

- **Edge presence test:** $H_0 : \mathbf{\Omega}_{jk}(z_0) = 0$ for a fixed $z_0 \in (0,1)$ and $j,k \in [d]$;
- **Super-graph test:** $H_0 : G^*(z_0) \subset G$ for a fixed $z_0 \in (0,1)$ and a fixed graph $G$;
- **Uniform edge presence test:** $H_0 : G^*(z) \subset G$ for all $z \in [z_L, z_U] \subset (0,1)$ and a fixed graph $G$.

For all the testing problems, the alternative hypotheses is the negation of the null. The edge presence test is concerned with a local hypothesis that $X_j$ and $X_k$ are conditionally independent given $\boldsymbol{X}_{\setminus\{j,k\}}$ for a particular value of the index $z_0$. Equivalently, under the null hypothesis of the edge presence test, the nodes $j$ and $k$ are not connected at a particular index value $z_0$. The null hypothesis under the super-graph test postulates that the true graph is a subgraph of a given graph $G$ for $Z = z_0$. It can also be interpreted as multiple-edge presence tests, since

$$H_0 : \mathbf{\Omega}_{jk}(z_0) = 0, \qquad \text{for all } (j,k) \in E^c, \tag{8}$$

where $E$ is the edge set of the graph $G$. The null hypothesis under the uniform edge presence test postulates that the true graph is a subgraph of $G$ for all index values in the range $[z_L, z_U]$. It is a generalization of the first two local tests to a global test over a range of index values. Similar to the super-graph test, this hypothesis is equivalent to the following

$$H_0 : \sup_{z \in [z_L, z_U]} |\mathbf{\Omega}_{jk}(z)| = 0, \qquad \text{for all } (j,k) \in E^c.$$

If the graph $G$ consists of the edge set $E = \{(a,b) \in V \times V \mid (a,b) \neq (j,k)\}$, the uniform edge presence test becomes a uniform single-edge test $H_0 : \sup_{z \in [z_L, z_U]} |\mathbf{\Omega}_{jk}(z)| = 0$.

Next, we provide details on how to construct tests for the above three hypothesis.

### 3.1 Edge Presence Testing

We consider the hypothesis $H_0 : \mathbf{\Omega}_{jk}(z_0) = 0$, for a fixed $z_0 \in (0,1)$ and $j,k \in [d]$. In order to construct a test for this hypothesis, we introduce the score function

$$\widehat{S}_{z|(j,k)}(\boldsymbol{\beta}) = \widehat{\mathbf{\Omega}}_j^T(z)\big(\widehat{\mathbf{\Sigma}}(z)\boldsymbol{\beta} - \mathbf{e}_k\big). \tag{9}$$

The argument $\boldsymbol{\beta}$ of the score function corresponds to the $k$-th column of $\mathbf{\Omega}(z)$. Our test is based on the score function evaluated at $\widehat{\mathbf{\Omega}}_{k\setminus j}$ which is an estimator of $\mathbf{\Omega}_k(z)$ under the null hypothesis, defined as

$$\widehat{\mathbf{\Omega}}_{k\setminus j}(z) = \big(\widehat{\mathbf{\Omega}}_{1k}(z), \ldots, \widehat{\mathbf{\Omega}}_{(j-1)k}(z), 0, \widehat{\mathbf{\Omega}}_{(j+1)k}(z), \ldots, \widehat{\mathbf{\Omega}}_{dk}(z)\big)^T \in \mathbb{R}^d,$$

where $\widehat{\mathbf{\Omega}}(z)$ is an estimator of $\mathbf{\Omega}(z)$. That is, we use $\widehat{S}_{z|(j,k)}\big(\widehat{\mathbf{\Omega}}_{k\setminus j}(z)\big)$ as the score statistic. We establish statistical properties of this statistic later. We first develop intuition for

why $\widehat{S}_{z|(j,k)}\big(\widehat{\mathbf{\Omega}}_{k\setminus j}(z)\big)$ is a good testing statistic for $H_0 : \mathbf{\Omega}_{jk}(z_0) = 0$. If we replace $\widehat{\mathbf{\Omega}}$ and $\widehat{\mathbf{\Sigma}}$ by the truth $\mathbf{\Omega}$ and $\mathbf{\Sigma}$ in (9), we can find the score statistic $\widehat{S}_{z|(j,k)}\big(\widehat{\mathbf{\Omega}}_{k\setminus j}(z)\big) \approx \mathbf{\Omega}_j^T(z)\big(\mathbf{\Sigma}(z)\mathbf{\Omega}_{k\setminus j}(z) - \mathbf{e}_k\big) = \mathbf{\Omega}_{jk}(z)$. Therefore, the score statistic is close to zero under the null, and $\widehat{S}_{z|(j,k)}\big(\widehat{\mathbf{\Omega}}_{k\setminus j}(z_0)\big) \approx \widehat{\mathbf{\Omega}}_{jk}(z_0)$ under the alternative. In specific, under $H_0$, the testing statistic is close to

$$\widehat{S}_{z|(j,k)}\big(\widehat{\mathbf{\Omega}}_{k\setminus j}\big) \approx \mathbf{\Omega}_j^T(z)\big(\widehat{\mathbf{\Sigma}}(z) - \mathbf{\Sigma}(z)\big)\mathbf{\Omega}_k(z) \approx \mathbf{\Omega}_j^T(z)\big[\dot{\mathbf{\Sigma}}(z) \circ \big(\widehat{\mathbf{T}}(z) - \mathbf{T}(z)\big)\big]\mathbf{\Omega}_k(z), \quad (10)$$

where $\dot{\mathbf{\Sigma}}_{jk}(z) = (\pi/2)\cos\big((\pi/2)\tau_{jk}(z)\big)$ is the derivative of $\mathbf{\Sigma}_{jk}(\cdot)$ and "$\approx$" denotes equality up to a smaller order term. The first approximation in (10) is due to replacing $\widehat{\mathbf{\Omega}}$ with the truth and the second approximation is due to the Taylor expansion. A rigorous derivation of this argument can be found in Appendix B.1. We note that the right-hand side of (10) is a linear function of $\widehat{\mathbf{T}}(z)$, which is a $U$-statistic. By applying the central limit theorem for $U$-statistics, we will show the asymptotic normality of $\widehat{S}_{z|(j,k)}\big(\widehat{\mathbf{\Omega}}_{k\setminus j}\big)$. See Theorem 1 for details.

Under the null, we have that

$$\sqrt{nh} \cdot \sigma_{jk}^{-1}(z_0)\widehat{S}_{z_0|(j,k)}\big(\widehat{\mathbf{\Omega}}_{k\setminus j}(z_0)\big) \rightsquigarrow N(0,1),$$

where $\sigma_{jk}^2(z_0) = f_Z^{-4}(z_0)\mathrm{Var}(\mathbf{\Omega}_j^S(z_0)\mathbf{\Theta}_{z_0}\mathbf{\Omega}_k(z_0))$, and $\mathbf{\Theta}_z$ is a random matrix with elements

$$(\mathbf{\Theta}_z)_{jk} = \pi\cos\big((\pi/2)\tau_{jk}(z)\big)\tau_{jk}^{(1)}(Y), \qquad \text{where} \quad (11)$$

$$\tau_{jk}^{(1)}(\boldsymbol{x}, z) = \sqrt{h} \cdot \mathbb{E}\left[K_h\left(z - z_0\right) K_h\left(Z - z_0\right)\left(\mathrm{sign}(X_j - x_j)\,\mathrm{sign}(X_k - x_k) - \tau_{jk}(z_0)\right)\right], \quad (12)$$

where the expectation is taken for $Z$ and $\boldsymbol{X}$. For simplicity, we denote $y = (\boldsymbol{x}^T, z)^T$ and write $\tau_{jk}^{(1)}(y) := \tau_{jk}^{(1)}(\boldsymbol{x}, z)$. The form of the asymptotic variance comes from the Hoeffding decomposition of the $U$-statistics in (4), with $\tau_{jk}^{(1)}$ being the leading term of the decomposition. Technical details will be provided in Section 4.1 and Section B.1. The score statistic is a generalization of Rao's score tests in fixed dimensional parametric models. We can apply a one-step debiased estimator of $\mathbf{\Omega}_{jk}(z)$ from the score statistic $\widehat{S}_{z|(j,k)}\big(\widehat{\mathbf{\Omega}}_{k\setminus j}(z)\big)$ following the procedure similar to (Ning and Liu, 2017). They show that the one-step estimation procedure is asymptotically equivalent to the score statistic and we choose to use score statistic in this paper. Janková and van de Geer (2017) considered a similar debiasing procedure specific for the nodewise regression estimator. On the other hand, we will show in Theorem 1 that the score statistic $\widehat{S}_{z|(j,k)}\big(\widehat{\mathbf{\Omega}}_{k\setminus j}(z)\big)$ can be applied to any estimator $\widehat{\mathbf{\Omega}}$ has sharp enough statistical rate.

In order to use the score function as a test statistic, we need to estimate its asymptotic variance $\sigma_{jk}^2(z_0)$. For any $1 \le s \le n$ and $1 \le j, k \le d$, let

$$q_{s,jk}(z) = \frac{\sqrt{h}}{n-1}\sum_{s' \neq s}\left[\omega_z(Z_s, Z_{s'})\left(\mathrm{sign}\big((X_{sj} - X_{s'j})(X_{sk} - X_{s'k})\big) - \widehat{\tau}_{jk}(z)\right)\right], \quad (13)$$

$$\widehat{\mathbf{\Theta}}_{jk}^{(s)}(z) = \pi\cos\left((\pi/2)\widehat{\tau}_{jk}(z)\right) q_{s,jk}(z).$$

With this notation, the leave-one-out Jackknife estimator for $\sigma_{jk}^2(z_0)$ is given as

$$\widehat{\sigma}_{jk}^2(z_0) = [\mathbb{U}_n(\omega_{z_0})]^{-2} \cdot \frac{1}{n} \sum_{s=1}^{n} \Big( \widehat{\mathbf{\Omega}}_j^S(z_0) \widehat{\mathbf{\Theta}}^{(s)}(z_0) \widehat{\mathbf{\Omega}}_k(z_0) \Big)^2, \tag{14}$$

where the matrix $\widehat{\mathbf{\Theta}}^{(s)}(z) = [\widehat{\mathbf{\Theta}}_{jk}^{(s)}(z)]$. As we have remarked in Section 2.1, $\widehat{\sigma}_{jk}^2(z_0)$ is an estimator for $f_Z^2(z)$. We divide $[\mathbb{U}_n(\omega_{z_0})]^{-2}$ in (14) in order to normalized the weights $\omega_z(Z_s, Z_{s'})$ in the $U$-statistics $q_{s,jk}(z)$ defined in (13). The Jackknife estimator is widely used when estimating the variance of a $U$-statistics, which is not an average of independent random variables. The leave-one-out statistic $q_{s,jk}(z)$ in (13) is estimates the expectation in (12) by leaving $Y_s$ out of the summation in $q_{s,jk}(z)$.

Finally, a level $\alpha$ test for $H_0 : \mathbf{\Omega}_{jk}(z_0) = 0$ is given as

$$\psi_{z_0|(j,k)}(\alpha) = \begin{cases} 1 & \text{if } \sqrt{nh} \cdot \big| \widehat{S}_{z_0|(j,k)} \big( \widehat{\mathbf{\Omega}}_{k\setminus j}(z_0) \big) / \widehat{\sigma}_{jk}(z_0) \big| > \Phi^{-1}(1-\alpha/2); \\ 0 & \text{if } \sqrt{nh} \cdot \big| \widehat{S}_{z_0|(j,k)} \big( \widehat{\mathbf{\Omega}}_{k\setminus j}(z_0) \big) / \widehat{\sigma}_{jk}(z_0) \big| \le \Phi^{-1}(1-\alpha/2), \end{cases}$$

where $\Phi(\cdot)$ is the cumulative distribution function of a standard normal distribution. The single-edge presence test is a cornerstone of more general hypothesis tests described in the next two sections. The properties of the test are given in Theorem 1.

### 3.2 Super-Graph Testing

In this section, we discuss super-graph testing. Recall that for a fixed $z_0$ and a predetermined graph $G = (V, E)$, the null hypothesis is

$$H_0 : G^*(z_0) \subset G. \tag{15}$$

From (8), we have that the super-graph test can be seen as a multiple test for presence of several edges. Therefore, we propose the following testing statistic based on the score function in (9):

$$S(z_0) = \sqrt{nh} \cdot \mathbb{U}_n(\omega_{z_0}) \max_{(j,k)\in E^c} \widehat{S}_{z_0|(j,k)} \big( \widehat{\mathbf{\Omega}}_{k\setminus j}(z_0) \big). \tag{16}$$

In order to estimate the quantile of $S(z_0)$, we develop a novel Gaussian multiplier bootstrap for $U$-statistics. Let $\{\xi_i\}_{i\in[n]}$ be $n$ independent copies of $N(0,1)$. Let $\widehat{\mathbf{T}}^B(z) = \big[\widehat{\tau}_{jk}^B(z)\big]$ where

$$\widehat{\tau}_{jk}^B(z) = \frac{\sum_{i\neq i'} K_h\left(Z_i - z\right) K_h\left(Z_{i'} - z\right) \operatorname{sign}\big((X_{ij} - X_{i'j})(X_{ik} - X_{i'k})\big) \left(\xi_i + \xi_{i'}\right)}{\sum_{i\neq i'} K_h\left(Z_i - z\right) K_h\left(Z_{i'} - z\right) \left(\xi_i + \xi_{i'}\right)}, \tag{17}$$

$$\widehat{\mathbf{\Sigma}}^B(z) = \sin\Big( \frac{\pi}{2} \widehat{\mathbf{T}}^B(z) \Big). \tag{18}$$

The Gaussian multiplier bootstrap statistic in (19) is motivated by the method developed in Chernozhukov et al. (2013), who proposed a bootstrap procedure to estimate a quantile of the supremum of high dimensional empirical processes. Their method, however, cannot be directly applied to our kernel Kendall's tau estimator in (4), which is a ratio of two $U$-statistics. If we compare (17) with (4), we add $\xi_i + \xi_{i'}$ into the bootstrap estimator in order to simulate the distribution of $\widehat{\tau}_{jk}(z)$ in (4).

The bootstrap estimator of the test statistic $S(z_0)$ in (16) is

$$S^B(z_0) = \sqrt{nh} \cdot \mathbb{U}_n[\omega_{z_0}^B] \max_{(j,k)\in E^c} \widehat{\mathbf{\Omega}}_j^S(z_0)\big(\widehat{\mathbf{\Sigma}}^B(z_0)\widehat{\mathbf{\Omega}}_{k\setminus j} - \mathbf{e}_k^T\big), \text{ where} \tag{19}$$

$$\mathbb{U}_n[\omega_{z_0}^B] = \frac{2}{n(n-1)} \sum_{i\neq i'} K_h\left(Z_i - z_0\right) K_h\left(Z_{i'} - z_0\right)(\xi_i + \xi_{i'}). \tag{20}$$

Here we multiply $\mathbb{U}_n[\omega_{z_0}^B]$ in (19) in order to eliminate the denominator in (17) such that the leading term of the bootstrap statistic $S^B(z_0)$ is a Gaussian multiplier bootstrap $U$-statistic. The correlation estimator in (7) has an additional sin transform. Therefore, a new nonlinear type of multiplier bootstraps in (17) and (19) are introduced to overcome these problems and novel technical tools are then developed to study statistical properties. See Theorem 3 for the statement of statistical properties.

Denote the conditional $(1-\alpha)$-quantile of $S^B(z_0)$ given $\{Y_i\}_{i=1}^n$ as $\widehat{c}_T(1-\alpha,\{Y_i\}_{i=1}^n)$. The level-$\alpha$ super-graph test is constructed as

$$\psi_{z_0|G}(\alpha) = \begin{cases} 1 & \text{if } S(z_0) > \widehat{c}_T(1-\alpha,\{Y_i\}_{i=1}^n); \\ 0 & \text{if } S(z_0) \leq \widehat{c}_T(1-\alpha,\{Y_i\}_{i=1}^n). \end{cases} \tag{21}$$

Note that the quantile $\widehat{c}_T(1-\alpha,\{Y_i\}_{i=1}^n)$ can be estimated by a Monte-Carlo method.

An alternative approach for the super-graph test in (15) is the multiple hypothesis testing. We can apply the Holm's multiple testing procedure (Holm, 1979) to control the family-wise error for the hypotheses set $\{H_{0,(jk)}\}_{(j,k)\in E^c}$ where $H_{0,(jk)} : \mathbf{\Omega}_{jk}(z_0) = 0$ and $G^*(z_0) \subset G = (V,E)$. However, it is not straightforward to obtain the nominal probability for the family-wise error in Holm's method. Moreover, we will show in Theorem 3 that our testing procedure is nominal.

### 3.3 Uniform Edge Presence Testing

In this section, we develop the uniform presence test for which the null hypothesis is given as

$$H_0 : G^*(z) \subset G \text{ for all } z \in [z_L, z_U].$$

This test is a generalization of the edge presence test to the uniform version over both edges and index. We again use the score function in (9) to construct the test statistic

$$W_G = \sqrt{nh} \sup_{z\in[z_L,z_U]} \max_{(j,k)\in E^c} \mathbb{U}_n[\omega_z]\widehat{S}_{z|(j,k)}\big(\widehat{\mathbf{\Omega}}_{k\setminus j}(z)\big) \tag{22}$$

and estimate a quantile of $W_G$ by developing a Gaussian multiplier bootstrap. Let

$$W_G^B = \sqrt{nh} \sup_{z\in[z_L,z_U]} \max_{(j,k)\in E^c} \mathbb{U}_n[\omega_z^B] \cdot \widehat{\mathbf{\Omega}}_j^T(z)\big(\widehat{\mathbf{\Sigma}}^B(z)\widehat{\mathbf{\Omega}}_{k\setminus j}(z) - \mathbf{e}_k^T\big), \tag{23}$$

where $\widehat{\mathbf{\Sigma}}^B(z)$ is defined in (18). Let $\widehat{c}_W(1-\alpha,\{Y_i\}_{i=1}^n)$ denote the conditional $(1-\alpha)$-quantile of $W_G^B$ given $\{Y_i\}_{i=1}^n$. Similar to (21), the level $\alpha$ uniform edge presence test is constructed as

$$\psi_G(\alpha) = \begin{cases} 1 & \text{if } W_G > \widehat{c}_W(1-\alpha,\{Y_i\}_{i=1}^n); \\ 0 & \text{if } W_G \leq \widehat{c}_W(1-\alpha,\{Y_i\}_{i=1}^n). \end{cases} \tag{24}$$

Theorem 4 provides statistical properties of the test. The statistics $W_G$ in (22) and $W_G^B$ in (23) involve taking supreme over $z \in [z_L, z_U]$. In practice, we approximate the suprema by evenly dividing $[z_L, z_U]$ into discrete grids and taking the maximum of the statistic over these discrete values in $[z_L, z_U]$.

## 4. Theoretical Properties

In this section, we establish the validity of tests proposed in the previous sections. Validity of tests rely on existence of estimators for the latent inverse correlation matrix with fast enough convergence. We show that the calibrated CLIME satisfy the testing requirements and, in addition, show that it achieves the minimax rate of convergence for a large class of models.

To facilitate the argument, we need the regularity and smoothness of the density function of $Z$ and the time-varying correlation matrix $\mathbf{\Sigma}(z)$. Let us first introduce the Hölder class $\mathcal{H}(\gamma, L)$ of smooth functions. The Hölder class $\mathcal{H}(\gamma, L)$ on $(0,1)$ is the set of $\ell = \lfloor \gamma \rfloor$ times differentiable functions $g : \mathcal{X} \mapsto \mathbb{R}$ whose derivative $g^{(\ell)}$ satisfies

$$|g^{(\ell)}(x) - g^{(\ell)}(y)| \leq L|x-y|^{\gamma-\ell}, \quad \text{for any } x, y \in \mathcal{X}$$

and $\lfloor \gamma \rfloor$ denotes the largest integer smaller than $\gamma$. In this paper, we need some regularity conditions for the functions in our model.

**Assumption 4.1 (Density function of $Z$)** *There exist constants $0 < \underline{f}_Z < \bar{\mathrm{f}}_Z < \infty$ such that the marginal density $f_Z$ of the index variable $Z$ has its image in $[\underline{f}_Z, \bar{\mathrm{f}}_Z]$ and $f_Z \in \mathcal{H}(2, \bar{\mathrm{f}}_Z)$.*

**Assumption 4.2 (Regularization of $\mathbf{\Sigma}_{jk}(\cdot)$)** *The correlations $\mathbf{\Sigma}_{jk}(\cdot) \in \mathcal{H}(2, M_\sigma)$ for some constant $M_\sigma < \infty$ given any $1 \leq j, k \leq d$.*

The above two assumptions are standard assumptions on the marginal distribution of $Z$ (Pagan and Ullah, 1999) and time-varying graphical models (see, for example, Kolar et al., 2010a).

**Assumption 4.3 (Kernel function)** *Through this paper, we assume the kernel function $K$, used in (4), is a symmetric density function supported on $[-1, 1]$ with bounded variation, i.e., $||K||_\infty \vee \mathrm{TV}(K) < \infty$,*

$$\int_{-1}^{1} K(u)du = 1 \text{ and } \int_{-1}^{1} uK(u)du = 0.$$

These properties are also required in Zhou et al. (2010). Many widely used kernels, including the uniform kernel $K(u) = 0.5\mathbb{1}(|u| < 1)$, the triangular kernel $K(u) = (1 - |u|)\mathbb{1}(|u| < 1)$, and the Epanechnikov kernel $K(u) = 0.75(1 - u^2)\mathbb{1}(|u| < 1)$, satisfy this assumption.

Finally, we list a generic assumption on the properties of $\widehat{\mathbf{\Sigma}}(z)$ in (7) and the an inverse correlation matrix estimator $\widehat{\mathbf{\Omega}}(z)$.

**Assumption 4.4 (Statistical rates)** *There are sequences* $r_{1n}, r_{2n}, r_{3n} = o(1)$ *such that*

$$\sup_{z\in(0,1)} \|\widehat{\mathbf{\Sigma}}(z) - \mathbf{\Sigma}(z)\|_{\max} \le r_{1n}, \quad \sup_{z\in(0,1)} \|\widehat{\mathbf{\Omega}}(z) - \mathbf{\Omega}(z)\|_1 \le r_{2n}, \text{ and}$$
$$\sup_{z\in(0,1)} \max_{j\in[d]} \|\widehat{\mathbf{\Sigma}}(z)\widehat{\mathbf{\Omega}}_j(z) - \mathbf{e}_j\|_\infty \le r_{3n},$$

*with probability at least* $1 - 1/d$.

Assumption 4.4 is a generic condition on the consistency of $\widehat{\mathbf{\Sigma}}(z)$ and $\widehat{\mathbf{\Omega}}(z)$. We aim to show that our testing methods are independent to any specific procedure to estimate $\mathbf{\Omega}(\cdot)$. The three rates in Assumption 4.4 are sufficient for the validity of our tests. Our inferential framework can thus be easily generalized to other inverse correlation matrix estimators as long as their rates satisfy Assumption 4.4. Under this assumption, the score statistic used for testing can be approximated by an asymptotically normal leading term. For our estimators $\widehat{\mathbf{\Sigma}}(\cdot)$ in (7) and $\widehat{\mathbf{\Omega}}(\cdot)$ in (2) or (3), we will show in Theorems 5 and 6 that

$$r_{1n} = O(\sqrt{\log(d/h)/(nh)}), \quad r_{2n} = O(s\sqrt{\log(d/h)/(nh)}) \text{ and } r_{3n} = O(\sqrt{\log(d/h)/(nh)}). \tag{25}$$

See Section 4.2 for more details.

### 4.1 Validity of Tests

In this section, we state theorems on asymptotic validity of the tests considered in Section 3. We first define the parameter space

$$\mathcal{U}_s(M,\rho) = \Big\{\mathbf{\Omega} \in \mathbb{R}^{d\times d} \,\big|\, \mathbf{\Omega} \succ 1/\rho, \|\mathbf{\Omega}\|_2 \le \rho, \max_{j\in[d]} \|\mathbf{\Omega}_j\|_0 \le s, \|\mathbf{\Omega}\|_1 \le M\Big\}. \tag{26}$$

This matrix class was considered in the literature on inverse covariance matrix estimation (Cai et al., 2016) and time-varying covariance estimation (Chen and Leng, 2016).

The following theorem gives us the limiting distribution of the score function defined in (9).

**Theorem 1 (Edge presence test)** *For a fixed* $z_0 \in (0,1)$*, suppose* $\mathbf{\Omega}(z_0) \in \mathcal{U}_s(M,\rho)$*, Assumption 4.4 holds with* $\sqrt{nh}\cdot(r_{2n}(r_{1n}+r_{3n})) = o(1)$ *and the bandwidth* $h$ *satisfies*

$$\sqrt{nh}\big(\log(dn)/(nh) + h^2\big) + s^3/\sqrt{nh} = o(1). \tag{27}$$

*Furthermore, for a fixed and* $j,k \in [d]$*, suppose there exists* $\theta_{\min} > 0$ *such that*

$$\mathbb{E}(\mathbf{\Omega}_j^T(z_0)\mathbf{\Theta}_{z_0}\mathbf{\Omega}_k(z_0))^2 \ge \theta_{\min}\|\mathbf{\Omega}_j(z_0)\|_2^2\|\mathbf{\Omega}_k(z_0)\|_2^2,$$

*then under* $H_0 : \mathbf{\Omega}_{jk}(z_0) = 0$*, we have that*

$$\sqrt{nh}\cdot\sigma_{jk}^{-1}(z_0)\widehat{S}_{z_0|(j,k)}\big(\widehat{\mathbf{\Omega}}_{k\setminus j}(z_0)\big) \rightsquigarrow N(0,1),$$

*where* $\sigma_{jk}^2(z_0) = f_Z^{-4}(z_0)\mathrm{Var}(\mathbf{\Omega}_j^T(z_0)\mathbf{\Theta}_{z_0}\mathbf{\Omega}_k(z_0))$ *and* $\mathbf{\Theta}_{z_0}$ *is defined in* (11).

We have two sets of scaling conditions in the above theorem. Under the condition $\sqrt{nh}(r_{2n}(r_{1n}+r_{3n})) = o(1)$, the first heuristic approximation "$\approx$" in (10) is valid. The condition in (27) guarantees that the leading term on the right hand side of (10) is asymptotically normal. In particular, the first term of (27) makes the second heuristic approximation in (10) valid and allows for control of the higher order term of the Hoeffding decomposition of the $U$-statistics in (4). If $r_{1n}, r_{2n}$ and $r_{3n}$ have the rates as in (25) (see Theorems 5 and 6 for more details), this condition becomes $\sqrt{nh} \cdot s(h^2 + \sqrt{\log(dn)/(nh)})^2 = o(1)$. We choose $h \asymp n^{-\nu}$, where $\nu > 1/5$ in order to remove the bias. The two scaling conditions in Theorem 1 can be replaced by $(s^3 + s\log(dn))/n^{(1-\nu)/2} = o(1)$. This is similar to the condition $s^2 \log d/\sqrt{n}$ in the inference for the Lasso estimator (Zhang and Zhang, 2013; van de Geer et al., 2014; Javanmard and Montanari, 2014). Here, the slower $n^{-(1-\nu)/2}$ term originates from the nonparametric relationship between the index and correlation matrix. The additional $s^2$ term comes from the matrix structure and is ignorable if $s = o\left(\sqrt{\log(dn)}\right)$.

The following lemma shows that the asymptotic variance of the score function can be consistently estimated.

**Lemma 2** *Suppose the conditions of Theorem 1 hold. If* $r_{2n}/h = o(1)$ *and* $\log(dn)/(nh^3) = o(1)$, *then the variance estimator* $\widehat{\sigma}^2_{jk}(z_0)$ *in* (14) *has* $\widehat{\sigma}^2_{jk}(z_0) \xrightarrow{P} \sigma^2_{jk}(z_0)$.

The proofs of Theorem 1 and Lemma 2 are deferred to Appendix B.1 and G.4 respectively. Stronger scaling conditions are needed for consistent estimation of variance as its estimator in (14) relies on controlling higher moments. Under the rates in (25) with $h \asymp n^{-\nu}$, for some $\nu > 1/5$, the scaling $(s^3 + s^2\log(dn))/n^{(1-\nu)/2} = o(1)$ suffices for the estimator to consistently estimate the variance.

We also have the following theorems on the asymptotic validity of the super-graph test.

**Theorem 3 (Super-graph test)** *Let* $z_0 \in (0,1)$ *and* $j,k \in [d]$ *be fixed. Assume the conditions of Theorem 1 hold. Suppose there exists* $\theta_{\min} > 0$ *such that for all* $j \neq k \in [d]$, $\mathbb{E}(\mathbf{\Omega}_j^T(z_0)\mathbf{\Theta}_{z_0}\mathbf{\Omega}_k(z_0))^2 \geq \theta_{\min}\|\mathbf{\Omega}_j(z_0)\|_2^2\|\mathbf{\Omega}_k(z_0)\|_2^2$, *and there exists a constant* $\epsilon > 0$ *such that* $\sqrt{nh}(r_{2n}(r_{1n}+r_{3n})) = O(n^{-\epsilon})$ *and*

$$\sqrt{nh}\big(\log(dn)/(nh) + h^2\big) + \log d/(nh^2) + (\log(dn))^7/(nh) = O(n^{-\epsilon}). \tag{28}$$

*Let* $G = (V,E)$ *be any fixed graph and* $G^*(z_0)$ *is the Markov graph corresponding to the index value* $z_0$. *Under the null hypothesis* $H_0 : G^*(z_0) \subset G$, *the test* $\psi_{z_0|G}(\alpha)$ *defined in* (21) *satisfies*

$$\sup_{\alpha\in(0,1)} \Big|\mathbb{P}_{H_0}\big(\psi_{z_0|G}(\alpha) = 1\big) - \alpha\Big| = O(n^{-c}) \tag{29}$$

*for some universal constant* $c$.

The next theorem shows the asymptotic validity of the uniform edge presence test.

**Theorem 4 (Uniform edge presence test)** *Assume that* $\mathbf{\Omega}(z) \in \mathcal{U}(M,\rho)$ *for any* $z \in (0,1)$ *and Assumption 4.4 is true. Suppose there exists* $\theta_{\min} > 0$ *such that for all* $j \neq k \in [d]$ *and* $z \in (0,1)$, $\mathbb{E}(\mathbf{\Omega}_j^T(z)\mathbf{\Theta}_z\mathbf{\Omega}_k(z))^2 \geq \theta_{\min}\|\mathbf{\Omega}_j(z)\|_2^2\|\mathbf{\Omega}_k(z)\|_2^2$, *and there exists a constant* $\epsilon > 0$ *such that* $\sqrt{nh}(r_{2n}(r_{1n}+r_{3n})) = O(n^{-\epsilon})$ *and*

$$\sqrt{nh}\big(\log(dn)/(nh) + h^2\big) + \log d/(nh^2) + (\log(dn))^8/(nh) = O(n^{-\epsilon}). \tag{30}$$

*Under the null hypothesis* $H_0 : G^*(z) \subset G$ *for all* $z \in [z_L, z_U]$*, the test* $\psi_G(\alpha)$ *defined in* (24) *satisfies*

$$\sup_{\alpha \in (0,1)} \Big| \mathbb{P}_{H_0}\big(\psi_G(\alpha) = 1\big) - \alpha \Big| = O(n^{-c}) \tag{31}$$

*for some universal constant* $c > 0$.

We defer the proof of Theorem 3 to Appendix B.2 and the proof of Theorem 4 to Appendix F.

Theorems 3 and 4 only depend on the estimation rates of $\widehat{\mathbf{\Sigma}}(z)$ and $\widehat{\mathbf{\Omega}}(z)$ through Assumption 4.4. This implies that our inferential framework does not rely on exact model selection. We have $O(n^{-\epsilon})$ in (28) and (30) instead of $o(1)$ in (27) to achieve the polynomial convergence rate for type I error in (29) and (31). Comparing the scaling condition in (28) with the one in (27), the second term in (28) is dominated by the first term under a mild bandwidth rate $h = o(n^{-1/3})$. The third term $(\log(dn))^7/(nh)$ in (28) comes from a Berry-Essen bound on the suprema of increasing dimensional $U$-processes. Such a scaling condition is similar to the one in Chernozhukov et al. (2013). They showed that for the empirical process $\boldsymbol{W} = (W_1, \ldots, W_d)^T$ having the same covariance as the centered Gaussian vector $\boldsymbol{U} = (U_1, \ldots, U_d)^T$, the following Berry-Essen bound holds

$$\sup_{t \in \mathbb{R}} \Big| \mathbb{P}\big(\max_j W_j \le t\big) - \mathbb{P}\big(\max_j U_j \le t\big) \Big| = O\left( \left( (\log(dn))^7/n \right)^{1/6} \right).$$

Comparing with our condition that $(\log(dn))^7/(nh) = O(n^{-\epsilon})$, the additional term $nh$ in the denominator comes from the nonparametric part of our estimator. Furthermore, Theorem 4 requires a stronger scaling condition in (30), where the term $(\log(dn))^8/(nh) = O(n^{-\epsilon})$ arises from the additional supremum over $z \in [z_L, z_U]$ in the uniform edge presence test.

## 4.2 Consistency of Estimation

In this section, we show that the Assumption 4.4 holds under mild conditions on the data generating process. We give explicit rates for $r_{1n}, r_{2n}$ and $r_{3n}$ under concrete estimation procedures. We first show the estimation rate of $\widehat{\mathbf{\Sigma}}$ given in (7). Next, we give the rate of convergence for $\mathbf{\Omega}(z)$ when using the (calibrated) CLIME estimator.

We establish rates of convergence that are uniform in bandwidth $h$. Uniform in bandwidth results are important as they ensure consistency of our estimators even when the bandwidth is chosen in a data-driven way, which is the case in practice, including the cross-validation over integrated squared error Hall (1992) and other risks (Muller and Stadtmuller, 1987; Ruppert et al., 1995; Fan and Gijbels, 1995). See Jones et al. (1996) for a survey of other methods. Existing literature on uniform in bandwidth consistency focuses on low dimensional problems (see, for example, Einmahl and Mason, 2005). High dimensional statistical methods usually have more tuning parameters and it is hard to guarantee that the selected bandwidth satisfies an optimal scaling condition. To the best of our knowledge this is the first result established in a high-dimensional regime that shows uniform consistency for a wide range of possible bandwidths.

The following theorem shows the rate of the covariance matrix estimator.

**Theorem 5** *Assume* $\log d/n = o(1)$ *and the bandwidths* $0 < h_l < h_u < 1$ *satisfy*

$$h_l n/\log(dn) \to \infty \quad \text{and} \quad h_u = o(1).$$

*There exists a universal constant* $C_{\mathbf{\Sigma}} > 0$ *such that for any* $\delta \in (0,1)$*, we have*

$$\sup_{h\in[h_l,h_u]} \sup_{z\in(0,1)} \frac{\big\|\widehat{\mathbf{\Sigma}}(z) - \mathbf{\Sigma}(z)\big\|_{\max}}{h^2 + \sqrt{(nh)^{-1}\big[\log(d/h) \vee \log\big(\delta^{-1}\log\big(h_u h_l^{-1}\big)\big)\big]}} \le C_{\mathbf{\Sigma}} \tag{32}$$

*with probability* $1-\delta$.

The proof of this theorem is deferred to Appendix A.1. Using (32), we can determine the rate of $r_{1n}$ in Assumption 4.4. The supremum over the bandwidth $h$ in (32) implies that if a data-driven bandwidth $\widehat{h}_n$ satisfies $\mathbb{P}(h_l \le \widehat{h}_n \le h_u) \longrightarrow 1$, then with high probability,

$$\sup_{z\in(0,1)} \big\|\widehat{\mathbf{\Sigma}}(z) - \mathbf{\Sigma}(z)\big\|_{\max} \le C_{\mathbf{\Sigma}} \left(\widehat{h}_n^2 + \sqrt{\frac{\log(d/\widehat{h}_n)}{n\widehat{h}_n}}\right). \tag{33}$$

The first term in the rate is the bias and the second is the variance. Our result is sharper than the rate $O\left(h^2\sqrt{\log d} + \sqrt{\log d/(nh)}\right)$ established in Lemma 9 of Chen and Leng (2016). The uniform consistency result $\sup_{h_l\le h\le h_u}\sup_{z\in(0,1)}\big\|\widehat{\mathbf{\Sigma}}(z) - \mathbf{\Sigma}(z)\big\|_{\max} = o_P(1)$ holds for a wide range of bandwidths satisfying $h_l n/\log(dn) \to \infty$ and $h_u = o(1)$, which allows flexibility for data-driven methods. In fact, such $h_l$ is the smallest to make the variance (33) converge and $h_u$ is the largest for the convergence of bias. When $d = 1$, the range $[h_l, h_u]$ is the same as for the kernel-type function estimators (Einmahl and Mason, 2005).

Due to the large capacity of the estimator $\widehat{\mathbf{\Sigma}}(z)$ in (32), which varies with both the bandwidth $h$ and the index $z$, the routine proof based on uniform entropy numbers does not easily apply here. We use the peeling method (Van de Geer, 2000) by slicing the range of $h$ into smaller intervals, for which the uniform entropy number is controllable. Finally, we assemble the interval specific bounds to obtain (32). See Section E.1 for more details.

Next, we give a result on the estimation consistency of the inverse correlation matrix. Let $\widehat{\mathbf{\Omega}}(z) = (\widehat{\mathbf{\Omega}}_1(z), \dots, \widehat{\mathbf{\Omega}}_d(z))$ where each column $\widehat{\mathbf{\Omega}}_j(z)$ is constructed either by using the CLIME in (2) or calibrated CLIME in (3) We recommend using the calibrated CLIME in practice due to the tuning issues discussed in Section 2.1.

**Theorem 6** *Suppose* $\mathbf{\Omega}(z) \in \mathcal{U}_s(M,\rho)$ *for all* $z \in (0,1)$*. Assume* $\log d/n = o(1)$*, the bandwidths* $0 < h_l < h_u < 1$ *satisfy* $h_l n/\log(dn) \to \infty$ *and* $h_u = o(1)$*. The regularization parameter* $\lambda$ *is chosen to satisfy* $\lambda \ge \lambda_{n,h} := C_{\mathbf{\Sigma}}\big(h^2 + \sqrt{\log(d/h)/(nh)}\big)$*, where* $C_{\mathbf{\Sigma}}$ *is the constant in* (32)*, for the calibrated CLIME and* $\lambda \ge M\lambda_{n,h}$ *for the CLIME estimator. Then there exists a universal constant* $C > 0$ *such that*

$$\sup_{h\in[h_l,h_u]} \sup_{z\in(0,1)} \frac{1}{\lambda M^2}\big\|\widehat{\mathbf{\Omega}}(z) - \mathbf{\Omega}(z)\big\|_{\max} \le C; \tag{34}$$

$$\sup_{h\in[h_l,h_u]} \sup_{z\in(0,1)} \frac{1}{\lambda s M}\big\|\widehat{\mathbf{\Omega}}(z) - \mathbf{\Omega}(z)\big\|_1 \le C;$$

$$\sup_{z\in(0,1)} \max_{j\in[d]} \frac{1}{\lambda M} \cdot \|\widehat{\mathbf{\Omega}}_j^T\widehat{\mathbf{\Sigma}} - \mathbf{e}_j\|_\infty \le C, \tag{35}$$

*with probability* $1 - 1/d$.

The proof is deferred to Appendix A.2. From this theorem, we can see that (34) determines $r_{2n}$ and (35) determines $r_{3n}$ in Assumption 4.4. We can plug the $r_{1n}$, $r_{2n}$ and $r_{3n}$ given in (32), (34) and (35) into the condition $\sqrt{nh}(r_{2n}(r_{1n} + r_{3n})) = O(n^{-\epsilon})$ stated in Theorems 1, 3 and 4 to get an explicit condition for $n, h, s, d$.

**Corollary 7** *Let* $\widehat{\mathbf{\Omega}}(z)$ *be the calibrated CLIME estimator with* $\lambda \geq \lambda_{n,h} = C_{\mathbf{\Sigma}}\big(h^2 + \sqrt{\log(d/h)/(nh)}\big)$ *or the CLIME estimator with* $\lambda \geq M\lambda_{n,h}$. *Then the Assumption 4.4 and all conditions on* $r_{1n}$, $r_{2n}$ *and* $r_{3n}$ *in Theorems 1, 3 and 4 can be replaced with* $\sqrt{nh} \cdot s\left(h^2 + \sqrt{\log(dn)/(nh)}\right)^2 = o(1)$.

Theorem 6 implies that if the bandwidth satisfies $h \asymp (\log(dn)/n)^{1/5}$, then

$$\sup_{z\in(0,1)} \big\|\widehat{\mathbf{\Omega}}(z) - \mathbf{\Omega}(z)\big\|_{\max} \leq CM^2\left(\frac{\log d + \log n}{n}\right)^{2/5};$$

$$\sup_{z\in(0,1)} \big\|\widehat{\mathbf{\Omega}}(z) - \mathbf{\Omega}(z)\big\|_1 \leq CMs\left(\frac{\log d + \log n}{n}\right)^{2/5}$$

with probability $1 - 1/d$. When $\log d \gg \log n$, the optimal bandwidth for selection is larger than the standard scaling $h \asymp (\log n/n)^{1/5}$ for univariate nonparametric regression (Tsybakov, 2009). This is because we need to over-regularize each entry of $\widehat{\mathbf{\Sigma}}(z)$ to reduce the variance of entire matrix. The optimal bandwidth is also larger than the scaling for inference $h \asymp n^{-\nu}$ for some $\nu > 1/5$ in (27), since we also need to over-regularize to remove bias for inference.

#### 4.2.1 Optimality of Estimation Rate

The following theorem shows that the rate in (34) is minimax optimal.

**Theorem 8** *Consider the following class of the inverse correlation matrices*

$$\overline{\mathcal{U}}_s(M,\rho,L) := \{\mathbf{\Omega}(\cdot) \mid \mathbf{\Omega}(z) \in \mathcal{U}_s(M,\rho) \text{ for any } z \in (0,1), \\ \text{and } \mathbf{\Omega}_{jk}(\cdot) \in \mathcal{H}(2,L) \text{ for } j,k \in [d]\},$$

*where* $\mathcal{U}_s(M,\rho)$ *is defined in* (26). *We have the following two results on the minimax risk:*

1. *If* $\log(dn)/n = o(1)$, *then*

$$\inf_{\widehat{\mathbf{\Omega}}(z)} \sup_{\mathbf{\Omega}(\cdot)\in\overline{\mathcal{U}}_s(M,\rho,L)} \mathbb{E}\Big[\sup_{z\in(0,1)} \|\widehat{\mathbf{\Omega}}(z) - \mathbf{\Omega}(z)\|_{\max}\Big] \geq c\left(\frac{\log d + \log n}{n}\right)^{2/5}. \tag{36}$$

2. *If* $s^2\log(dn)/n = o(1)$ *and* $s^{-v}d \leq 1$ *for some* $v > 2$, *then*

$$\inf_{\widehat{\mathbf{\Omega}}(z)} \sup_{\mathbf{\Omega}(\cdot)\in\overline{\mathcal{U}}_s(M,\rho,L)} \mathbb{E}\Big[\sup_{z\in(0,1)} \|\widehat{\mathbf{\Omega}}(z) - \mathbf{\Omega}(z)\|_1\Big] \geq cs\left(\frac{\log d + \log n}{n}\right)^{2/5}. \tag{37}$$

The proof is deferred to Appendix C. We prove it by applying Le Cam's lemma (Le Cam, 1973) and constructing a finite collection $\mathbf{\Omega}(\cdot)$ from the function value matrices space $\overline{\mathcal{U}}_s(M,\rho,L)$. If we take the dimension $d=1$, our problem degenerates to the univariate twice differentiable function estimation. The risk on the left hand side of (36) becomes to the supreme norm between the estimated function and truth. The right hand side of (36) degenerates to $O((\log n/n)^{2/5})$ which matches the typical minimax rate for nonparametric regression in $\|\cdot\|_\infty$ risk (Tsybakov, 2009). This indicates the reason why we have the power $2/5$ in the rates.

## 5. Numerical Experiments

In this section, we explore finite sample performance of our estimation procedure and test using simulations. Furthermore, we also apply the super graph test to the brain image network.

### 5.1 Synthetic Data

We illustrate finite sample properties of the estimator in (3) on synthetic data. We consider three other competing methods: (1) the kernel Pearson CLIME estimator (Zhou et al., 2010; Yin et al., 2010), (2) the kernel graphical Lasso estimator (Zhou et al., 2010), and (3) the kernel neighborhood selection estimator (Kolar et al., 2010a). Zhou et al. (2010) and Yin et al. (2010) consider the correlation matrix estimator at $z$ as a weighted summation of Pearson's correlations

$$\widehat{\mathbf{\Sigma}}_P(z) = \frac{\sum_{i=1}^n K_h(Z_i - z)\boldsymbol{X}_i\boldsymbol{X}_i^T}{\sum_{i=1}^n K_h(Z_i - z)}.$$

This estimate can be plugged into the calibrated CLIME estimator in (3) to obtain the kernel Pearson inverse correlation estimator $\widehat{\mathbf{\Omega}}_{(1)}(z)$. Zhou et al. (2010) proposed to estimate the inverse correlation matrix by plugging $\widehat{\mathbf{\Sigma}}_P$ into the graphical lasso (Yuan and Lin, 2007) objective

$$\widehat{\mathbf{\Omega}}_{(2)}(z) = \underset{\mathbf{\Omega}\succ 0}{\operatorname{argmax}} \Big\{ \log\det\mathbf{\Omega} - \operatorname{tr}\big(\mathbf{\Omega}\widehat{\mathbf{\Sigma}}_P(z)\big) - \lambda\|\mathbf{\Omega}\|_1 \Big\},$$

resulting in the kernel graphical Lasso estimator. Kolar et al. (2010a) proposed to combine the neighborhood selection procedure Meinshausen and Bühlmann (2006) with local kernel smoothing. To estimate the $j$-th column of $\mathbf{\Omega}(z)$ they use

$$\widehat{\boldsymbol{\beta}}_j(z) = \underset{\boldsymbol{\beta}\in\mathbb{R}^{d-1}}{\operatorname{argmin}} \frac{1}{nh}\sum_{i=1}^n K\left(\frac{Z_i - z}{h}\right)(\boldsymbol{X}_{ij} - \boldsymbol{X}_{i\backslash j}\boldsymbol{\beta})^2 + \lambda\|\boldsymbol{\beta}\|_1,$$

where $\boldsymbol{X}_{i\backslash j}$ is a $(d-1)$-dimensional vector with the $j$-th entry of $\boldsymbol{X}_i$ removed. The kernel neighborhood selection estimator $\widehat{\mathbf{\Omega}}_{(3)}(z)$ is then obtained as as

$$\big[\widehat{\mathbf{\Omega}}_{(3)}(z)\big]_{jj} = 1 \text{ and } \big[\widehat{\mathbf{\Omega}}_{(3)}(z)\big]_{\backslash j,j} = \widehat{\boldsymbol{\beta}}_j(z), \text{ for all } j\in[d].$$

We describe a procedure to generate the graph $G^*(z)$ and inverse correlation matrix $\mathbf{\Omega}(z)$ at each $z\in(0,1)$. At any index $z\in(0,1)$, we generate a graph with $d=50$ nodes and $e = 25$ by selecting edge from the 4-nearest neighbor graph illustrated in Figure 1(a). We first generate $Z_1, \ldots, Z_n$ independently from Uniform$(0, 1)$ with the sample size $n \in \{200, 500, 800\}$ and then generate $\boldsymbol{\Omega}(z)$ at each index value $Z_i$ as follows.

1. Randomly select 25 edges from the 4-nearest neighbor graph and generate the initial graph $G^{(0)}$. The structure of the underlying graph will change at the following anchor points $z = 0.2, 0.4, 0.6, 0.8$. At the $\ell$-th anchor point, we randomly remove 10 edges from the previous graph $G^{(\ell-1)}$ and randomly add 10 edges from the 4-nearest neighbor graph that did not belong to the previous graph $G^{(k-1)}$. We therefore generate 5 anchor graphs $G^{(0)}, \ldots, G^{(4)}$. On the $\ell$-th interval $[(\ell-1)/5, \ell/5]$, the graph structure stays constant and equal to $G^{(\ell-1)}$ for $\ell = 1, \ldots, 5$.

2. Given the graph structure we generate $\boldsymbol{\Omega}(z)$. At the middle of each interval, that is, for index values $z = 0.1, 0.3, 0.5, 0.7, 0.9$, if the edge $(j, k)$ belongs to the graph at that time, we randomly generate $\boldsymbol{\Omega}_{jk}(z)$ from Uniform$[\mu, 0.9]$, where $\mu$ is the minimal signal strength, otherwise we set $\boldsymbol{\Omega}_{jk}(z)$ to be zero. For $1 \leq \ell \leq 4$, if the edge $(j, k)$ belongs to both $G^{(\ell)}$ and $G^{(\ell+1)}$, we let $\boldsymbol{\Omega}_{jk}(z_\ell)$ be generated from Uniform$[\mu, 0.9]$, otherwise we set it to be zero. We also generate $\boldsymbol{\Omega}_{jk}(0)$ from Uniform$[\mu, 0.9]$ if the edge $(j, k)$ is in $G^{(0)}$ and similarly for $\boldsymbol{\Omega}_{jk}(1)$ using $G^{(4)}$. The signal strength is set to $\mu = 0.5$.

3. For any $z \in (0, 1)$, if $z \in [\ell/10, (\ell+1)/10]$ for some $\ell = 0, \ldots, 9$, $\boldsymbol{\Omega}_{jk}(z)$ is set to be an linear interpolation of $\boldsymbol{\Omega}_{jk}(\ell/10)$ and $\boldsymbol{\Omega}_{jk}((\ell+1)/10)$. We first rescale the diagonal of $\boldsymbol{\Omega}(z)$ by $\boldsymbol{\Omega}(z) + (1 - \lambda_{\min}(\boldsymbol{\Omega}(z)))\mathbf{I}_d$, where $\lambda_{\min}(\boldsymbol{\Omega}(z))$ is the minimum eigenvalue of $\boldsymbol{\Omega}(z)$ in order to make its minimum eigenvalue equal to one for each time $z \in (0, 1)$. We then get the covariance matrix $\boldsymbol{\Sigma}(z) = \boldsymbol{\Omega}^{-1}(z)$ and normalize its diagonal to all ones. An illustration of the time-varying graphical model is shown in Figure 1(b) - Figure 1(f).

Given the inverse correlation matrix $\boldsymbol{\Omega}(z)$, we consider four data generation schemes:

1. Gaussian, where $\boldsymbol{X} \mid Z = z \sim N(\mathbf{0}, \boldsymbol{\Sigma}(z))$;

2. Gaussian Copula, where $\boldsymbol{X} \mid Z = z \sim (\Phi(\boldsymbol{Y}_1(z)), \ldots, \Phi(\boldsymbol{Y}_d(z)))^T$, $\boldsymbol{Y} \mid Z = z \sim N(\mathbf{0}, \boldsymbol{\Sigma}(z))$ and $\Phi(\cdot)$ is the cumulative density function of standard normal distribution;

3. Gaussian with 2% contamination, where $\boldsymbol{X} \mid Z = z \sim N(\mathbf{0}, \boldsymbol{\Sigma}(z))$ and then 2% of locations are randomly chosen and replaced by $\pm N(3, 3)$;

4. Gaussian with 5% contamination, is uses the same data generating mechanism as before, but more of the observations are contaminated.

All the methods under consideration require a specification of the kernel function satisfying Assumption 4.3. We choose the Epanechnikov kernel $K(u) = 0.75(1 - u^2)\mathbb{1}(|u| < 1)$ which can be easily checked that it satisfies Assumption 4.3. The bandwidth parameter is set as $h = 0.35/n^{1/5}$ for all four methods in order to facilitate easier comparison. For the calibrated CLIME, we choose $\gamma = 0.5$ in (3). We then estimate a sequence of inverse

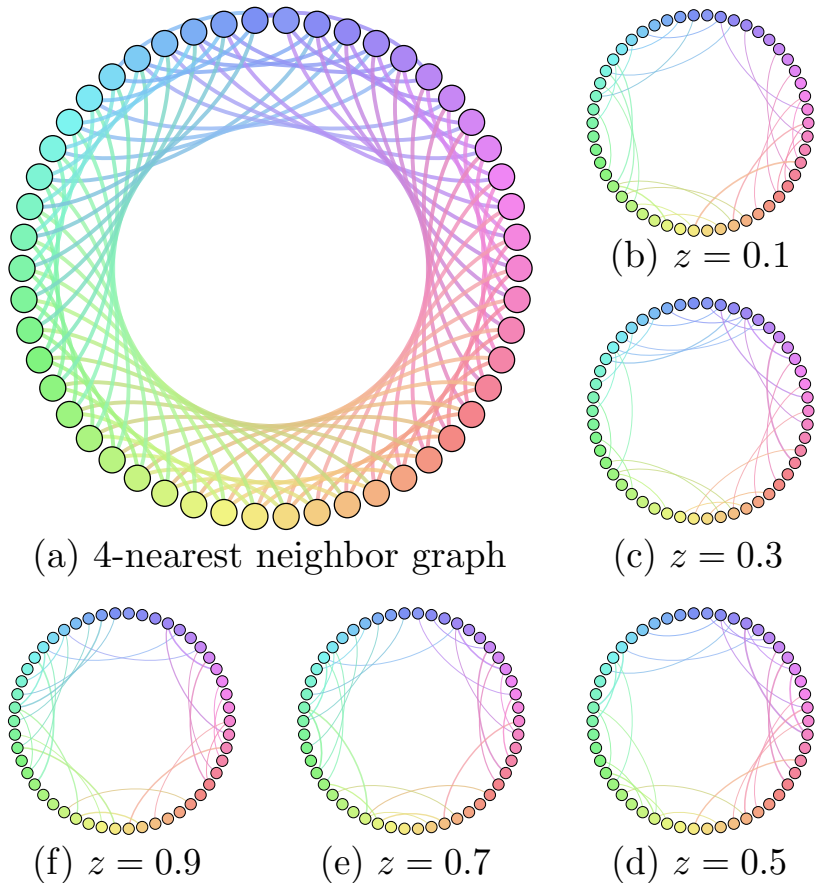

Figure 1: The 4-nearest neighbor graph and the time-varying graph for various index values $z$. The thickness of edges represents the magnitude of the inverse correlation matrix corresponding to that edge for a specific value $z$.

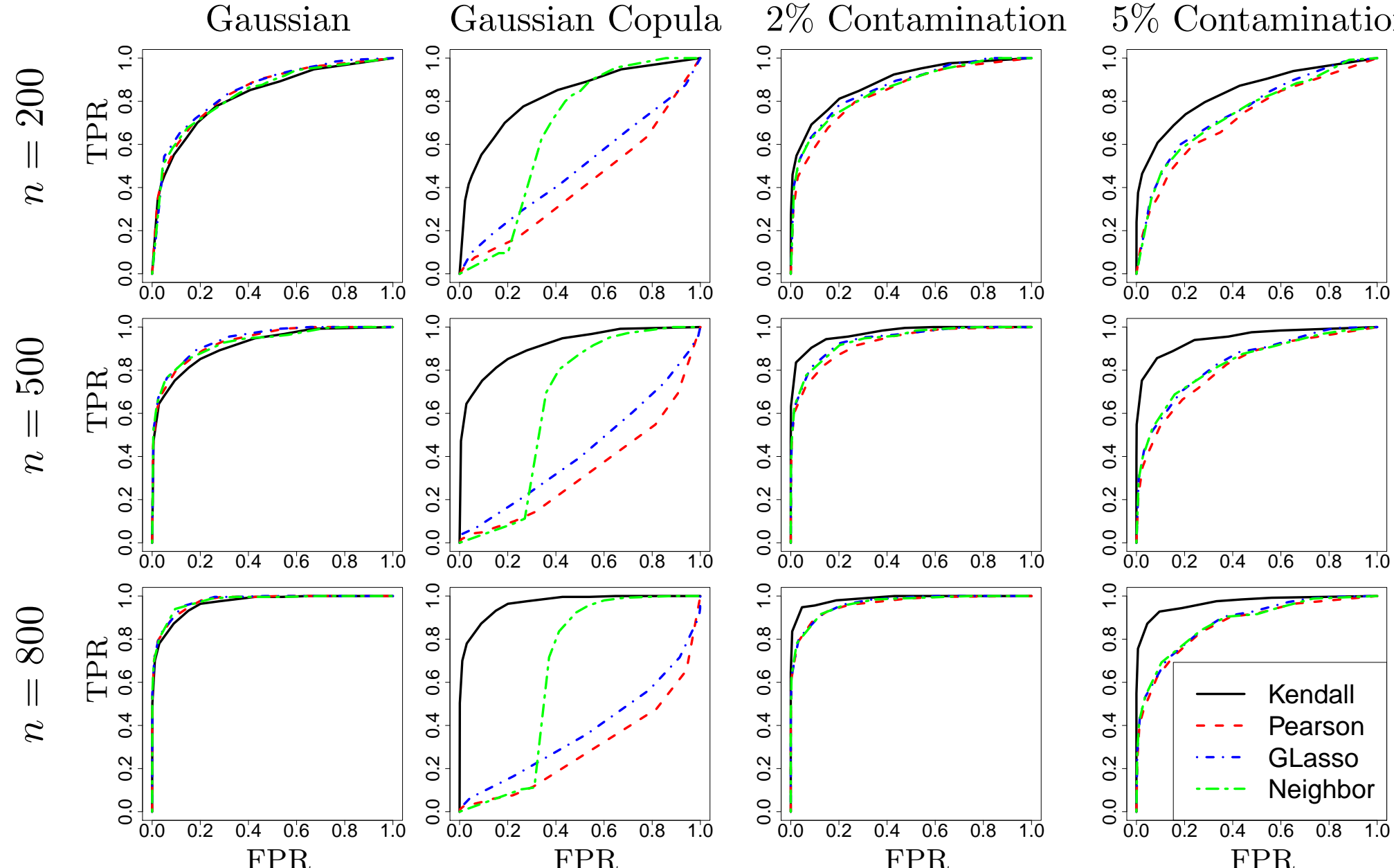

Figure 2: ROC curves for different estimators of the inverse correlation matrix estimator. The number of nodes is $d = 50$, the number of edges for each index value is $e = 25$ and the sample size is varied in $n \in \{200, 500, 800\}$. Each curve is obtained by averaging 100 simulation runs.

correlation matrices and graphs by changing the penalty parameter $\lambda$ over a large range of values. Let $\widehat{E}(z)$ be the estimated edge set and $E^*(z)$ is the true edge set at the index value

| $n$ | $\mu_0 = 0$ | $\mu_0 = 0.4$ | $\mu_0 = 0.6$ | $\mu_0 = 0.9$ |
|---|---|---|---|---|
| 600 | **0.108** | 0.722 | 0.892 | 0.994 |
| 800 | **0.090** | 0.822 | 0.900 | 0.998 |
| 1,000 | **0.056** | 0.736 | 0.968 | 1.000 |

Table 1: Size (bold) and power of the local edge presence test $H_0 : \mathbf{\Omega}_{12}(0.5) = 0$ at significance level 95% with various signal strength $\mu_0$. Results reported based on 500 simulation runs.

$z$. The true positive rate (TPR) and false positive rate (FPR) are defined as

$$\text{TPR}(z) = \frac{|\widehat{E}(z) \bigcap E^*(z)|}{|E^*(z)|} \qquad \text{and} \qquad \text{FPR}(z) = \frac{|\widehat{E}(z) \backslash E^*(z)|}{d(d-1)/2 - |E^*(z)|},$$

where $|S|$ denotes cardinality of the set $S$. To measure the quality of graph estimation for $z \in (0,1)$, we randomly choose 10 data points from $\{Z_i\}_{i\in[n]}$ and compute the averaged TPR and FPR on these 10 points as the overall TPR and FPR of the graph estimation. Figure 2 illustrates the ROC curves of the overall TPR and FPR for the four competing methods under four schemes for $n = 200, 500$ and 800. We can observe that the proposed estimator is slightly worse compared to other three estimators when the data are Gaussian conditionally on the index value. In this setting the data generating assumptions are satisfied for the other three procedures. On the other hand, when the data are generated according to the Gaussian copula distribution, the Gaussian assumption is violated and our estimator performs better compared to the other three estimators. The third and fourth columns of Figure 2 further illustrate that our estimator is more robust to corruption of data.

In addition to graph estimation accuracy, we consider the numerical performance of testing procedures proposed in the paper. We first focus on the local edge presence test introduced in Section 3.1. Consider the following null hypothesis $H_0 : \mathbf{\Omega}_{12}(0.5) = 0$. We choose the bandwidth $h = 0.9n^{-1/5}$, $\gamma = 0.5$ in (3) and the penalty parameter $\lambda = 0.2\big(h^2 + \sqrt{\log(d/h)/(nh)}\big)$ for the sample size $n \in \{600, 800, 1000\}$. The data generating process is the same as before, except that we set $\mathbf{\Omega}_{21}(z) = \mu_0$ for all $z \in (0,1)$ where $\mu_0 \in \{0, 0.4, 0.6, 0.9\}$. We use (21) to test the null hypothesis at significance level 95% and estimate the power based on 500 repetitions. The Q-Q plots of the testing statistic for various sample sizes are shown in Figure 3. The empirical distribution of the testing statistic is close to standard normal distribution. The results on the power of these tests are summarized in Table 5.1. When the signal strength $\mu_0 = 0$, the size of the test approaches the nominal value 0.05. When $\mu \geq 0.4$, the power approaches 1, showing validity of the proposed test.

| | $n = 600$ | | | $n = 800$ | | | $n = 1,000$ | | |
|---|---|---|---|---|---|---|---|---|---|
| | $\mu = 0$ | $\mu = 0.4$ | $\mu = 0.9$ | $\mu = 0$ | $\mu = 0.4$ | $\mu = 0.9$ | $\mu = 0$ | $\mu = 0.4$ | $\mu = 0.9$ |
| $k = 0$ | **0.118** | 0.954 | 0.958 | **0.088** | 0.958 | 0.976 | **0.068** | 0.977 | 0.992 |
| $k = 2$ | **0.128** | 0.900 | 0.926 | **0.090** | 0.944 | 0.964 | **0.070** | 0.970 | 0.982 |
| $k = 4$ | **0.098** | **0.089** | **0.058** | **0.082** | **0.068** | **0.054** | **0.066** | **0.052** | **0.048** |
| $k = 8$ | **0.090** | **0.074** | **0.054** | **0.066** | **0.058** | **0.048** | **0.054** | **0.052** | **0.050** |

Table 2: Size (bold) and power of local edge present test $H_0 : G^*(z) \subset G$ for all $z \in (0,1)$ at significance level 95% with the super graph $G$ being a $k$-nearest neighbor graph for $k \in \{0, 2, 4, 8\}$. Results reported based on 500 simulation runs.

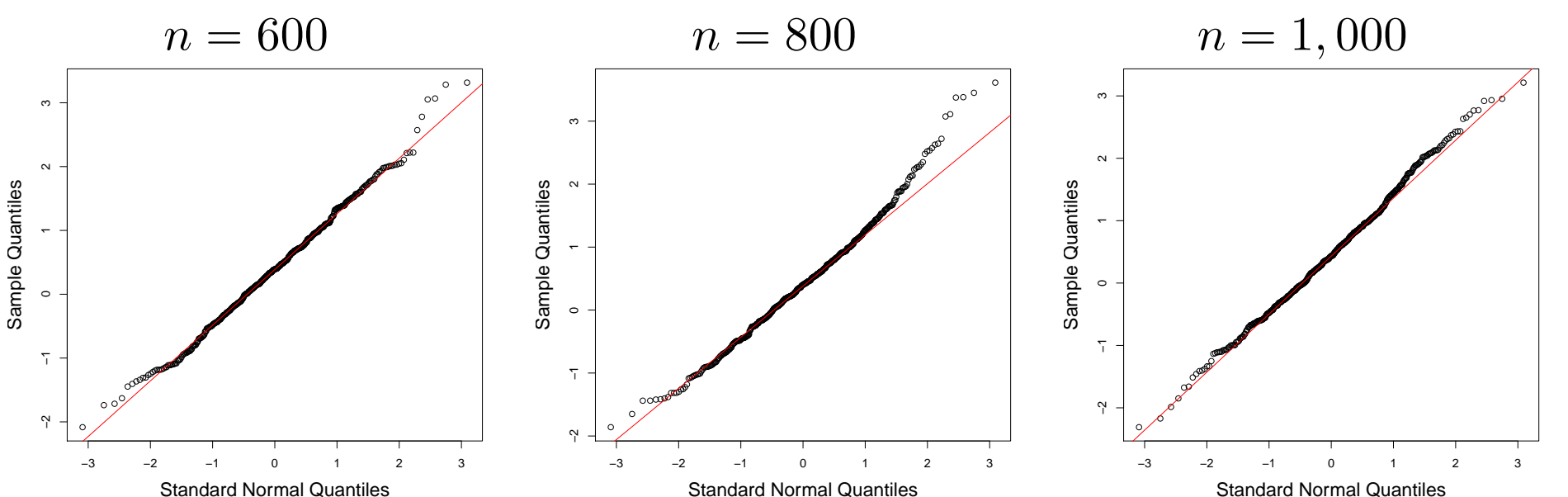

Figure 3: Q-Q plot of the testing statistic under the null hypothesis $H_0 : \mathbf{\Omega}_{12}(0.5) = 0$.

We also conduct a simulation for the uniform edge presence test introduced in Section 3.3. We consider the null hypothesis $H_0 : G^*(z) \subset G$ for all $z \in (0, 1)$, where the super graph $G$ is a $k$-nearest neighbor graph for $k \in \{0, 2, 4, 8\}$. Here a $k$-nearest neighbor graph has edges only connecting a vertex to its closest $k$ nodes. See Figure 1(a) for an illustration of a 4-nearest neighbor graph. When $k = 0$, it is a null graph whose adjacency matrix is identity. In order to illustrate the power of our test, we generate the nonzero entries of the inverse correlation matrix at anchor points from Uniform$[\mu, \max(\mu + 0.2, 0.9)]$ for $\mu = 0.4$ and 0.9. We also consider the setting where $\mathbf{\Omega}(z) = \mathbf{I}$ for all $z \in (0, 1)$. The sample size is varied as $n \in \{600, 800, 1000\}$. The bandwidth and tuning parameter are set in the same way as for the local edge presence test. When computing the test statistic in (22) the suprema over $z \in (0, 1)$ is approximated by taking the maximum of the statistic over 100 evenly spaced values in $(0, 1)$, and similarly for the bootstrapped statistic in (23). The size and power of the test are summarized in Table 5.1. From the data generating mechanism, we see that the true graph is always a subgraph of the 4-nearest graph. Therefore, the null hypothesis is true when the number of nearest neighbors is $k \in \{4, 8\}$ or $\mathbf{\Omega}(z) = \mathbf{I}$. From the results reported in Table 5.1, we observe that the uniform edge presence test has the correct size as the sample size increases. The alternative hypothesis is true when $k \in \{0, 2\}$ and $\mathbf{\Omega}(z) \neq \mathbf{I}$. In this setting, the less edges the super graph has, the more powerful the test is. That is because we take maximum over more edges in (16).

## 5.2 Super Brain Test

We apply the super graph test in Section 3.2 to the ADHD-200 brain imaging data set (Biswal et al., 2010). The ADHD-200 data set is a collection of resting-state functional MRI (R-fMRI) of subjects with and without attention deficit hyperactive disorder (ADHD) (Eloyan et al., 2012; The ADHD-200 Consortium, 2012). The data set contains 776 subjects: 491 controls, 195 cases diagnosed with ADHD of various types and 88 subjects with withheld diagnosis for the purpose of a prediction competition. For each subject there are between 76 and 276 R-fMRI scans. We focus on 264 voxels as the regions of interest (ROI) extracted by Power et al. (2011). These voxels are representative of the functional areas corresponding

| | $H_0 : G(8) \subset G(11.75)$ | $H_0 : G(11.75) \subset G(20)$ |
|---|---|---|
| Statistics | 2.793 | 3.548 |
| Quantiles | 1.899 | 2.154 |

Table 3: Results on the super brain hypothesis tests at 95% significant level. The row named "Statistics" are the values of the testing statistic in (16). The row named "Quantiles" are the 95%-quantile estimators obtained by the bootstrap estimator in (19).

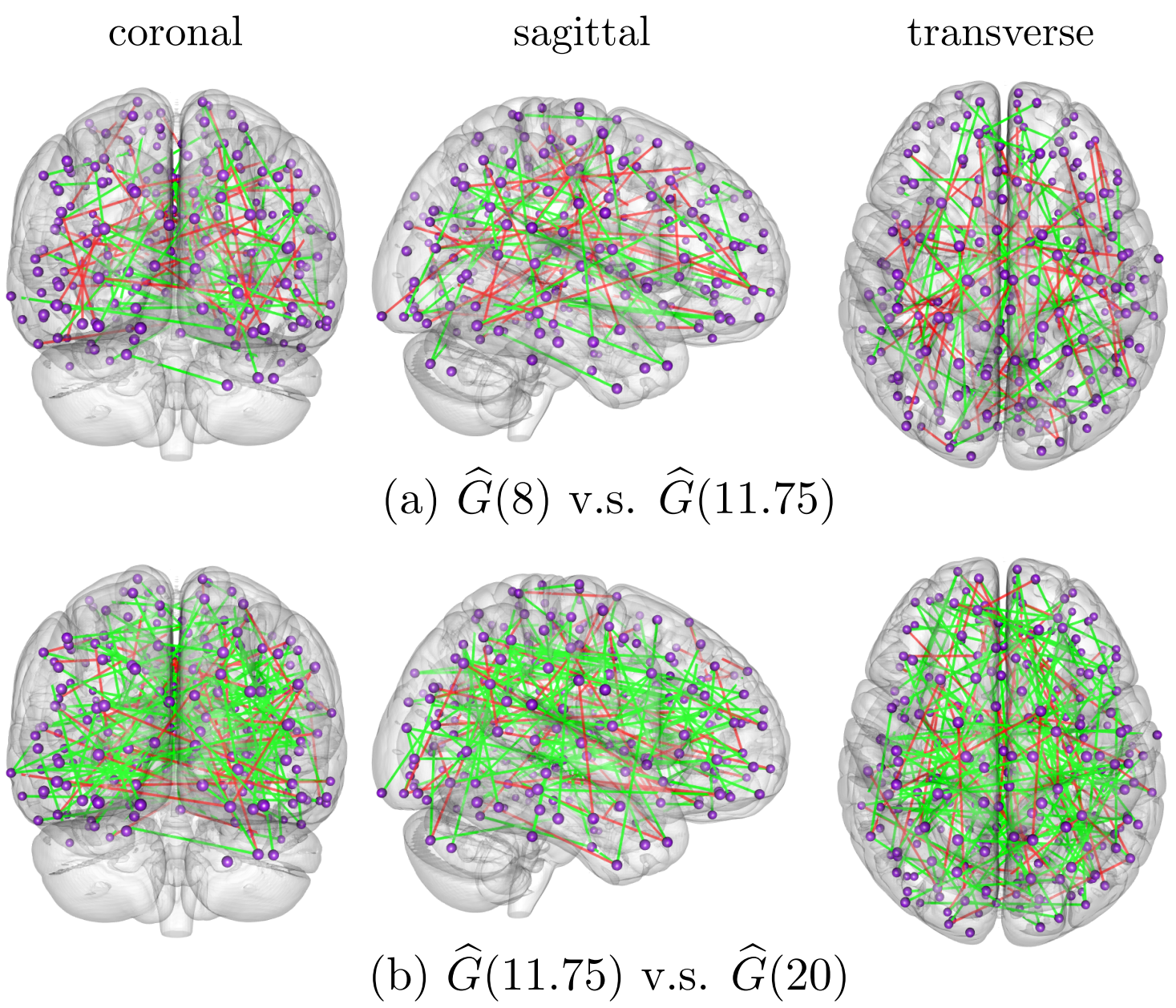

(a) $\widehat{G}(8)$ v.s. $\widehat{G}(11.75)$

(b) $\widehat{G}(11.75)$ v.s. $\widehat{G}(20)$

Figure 4: The differences between junior, median and senior neural networks. $\widehat{G}(8)$, $\widehat{G}(11.75)$ and $\widehat{G}(20)$ denote the estimated graphs at age 8, 11.75 and 20. The first row is the difference between $\widehat{G}(8)$ and $\widehat{G}(11.75)$. The green lines are the edges existing in $\widehat{G}(11.75)$ but not in $\widehat{G}(8)$ and the red edges only exist in $\widehat{G}(8)$ but not $\widehat{G}(11.75)$. The second row is the difference between $\widehat{G}(11.75)$ and $\widehat{G}(20)$. The green edges only exist in $\widehat{G}(20)$ and the red edges only exist in $\widehat{G}(11.75)$.

to the cerebral cortex and cerebellum. For each subject we also have covariates that include age, IQ, gender and handedness. For illustration purposes, we focus on 491 controls and explore how the structure of the neural network varies with age of subjects, which range from 7.08 to 21.83. Using the kernel smoothing technique to estimate the varying neural networks, Qiu et al. (2016) found that the connections tend to be denser at the age 21.83 than at ages 11.75 and 7.09. Such a discovery is also supported by Bartzokis et al. (2001) and Gelfand et al. (2003).

Based on these discoveries, a more interesting conjecture is whether the neural network is always growing with age. We can use our testing framework to investigate the claim that a neural network in a younger subject is a subgraph of a neural network in an older subject. Specifically, we choose three ages: 8, 11.75 and 20, and we are interested in the graphs $G(8)$, $G(11.75)$ and $G(20)$, where $G(z)$ is the true neural network at age $z$. The age 11.75 is chosen as the median age of subjects. We call the neural networks at ages 8, 11.75 and 20 as the junior, median and senior neural network, respectively. As we do not have access to the true networks, we will estimate them using the calibrated CLIME in (3). We first map the ages onto the interval $(0, 1)$ and set the tuning parameters for our procedure as $\gamma = 0.5$, the bandwidth $h = 0.002$ and the penalty parameter $\lambda = 0.03\big(h^2 + \sqrt{d/h}/(nh)\big)$, where the last two parameters are chosen through cross-validation. We estimate the neural networks $\widehat{G}(8)$, $\widehat{G}(11.75)$ and $\widehat{G}(20)$ by (3) at ages 8, 11.75 and 20. The estimated graphs are used as super graphs in defining the null hypothesis. In particular, we have the following two tests: $H_0 : G(8) \subset \widehat{G}(11.75)$ and $H_0 : G(11.75) \subset \widehat{G}(20)$. Although we use the random estimators as the super graphs in these hypotheses, the testing procedure is still valid. This is because of the small bandwidth $h = 0.002$ making the data used in network estimation independent to the data used in these two tests. Therefore, the test can still be reduced as a super graph test.

Figure 4 shows the differences among the junior, median and senior brains. We observe that even if a later neural network has more edges compared to the earlier one, many edges existing at an earlier stage disappear later in the development. This implies that the conjecture that neural networks grow with age is not supported by the data. Table 5.2 quantifies evidence against the null hypothesis.

## 6. Discussion

In this paper, we consider the time-varying graphical model under the framework of non-paranormal distribution. Although it contains many examples of heavy-tailed distributions, there are many other cases uncovered in this family. It will be interesting to explore the estimation and inference methods for the time-varying graphical model under general heavy-tailed distributions with certain moment conditions. It is possible to incorporate existing methods including the Catoni's estimator (Catoni, 2012) and the median-of-means estimator (Hsu and Sabato, 2014) into the framework of this paper and conduct inference for the general heavy-tailed time-varying graphical models.

## Acknowledgments

The authors are grateful for the support of NSF CAREER Award DMS1454377, NSF IIS1408910, NSF IIS1332109, NIH R01MH102339, NIH R01GM083084, and NIH R01HG06841. This work is also supported by an IBM Corporation Faculty Research Fund at the University of Chicago Booth School of Business. This work was completed in part with resources provided by the University of Chicago Research Computing Center.

## Appendix A. Proof of Estimation Consistency

In the appendix, we use $c, C, c_1, C_1, c_2, C_2, \ldots$ to denote universal constants, independent of $n$ and $d$, whose values may change from line to line.

In this section, we prove uniform rates of convergence for the covariance matrix estimator $\widehat{\boldsymbol{\Sigma}}(z)$ and the inverse covariance estimator $\widehat{\boldsymbol{\Omega}}(z)$. These rates are uniformly valid over both the index $z$ and the kernel bandwidth $h$ used for the estimator $\widehat{\tau}_{jk}(z)$ in (4).

### A.1 Proof of Theorem 5

We apply the bias-variance decomposition for the kernel smoothed Kendall's tau statistic in the following two lemmas. The first lemma controls the variance term $|\widehat{\tau}_{jk}(z) - \mathbb{E}[\widehat{\tau}_{jk}(z)]|$ uniformly in $j, k \in [d]$, $z \in (0, 1)$, and $h \in [h_l, h_u]$. The second lemma controls the bias term $|\mathbb{E}[\widehat{\tau}_{jk}(z)] - \tau_{jk}(z)|$.

**Lemma 9 (Variance of Kendall's tau estimator)** *Assume that $n^{-1}\log d = o(1)$ and the bandwidths $0 < h_l < h_u < 1$ satisfy $h_l n/\log(dn) \to \infty$ and $h_u = o(1)$. There exists a universal constant $C > 0$ such that, with probability $1-\delta$, for sufficiently large $n$,*

$$\sup_{j,k\in[d]} \sup_{h\in[h_l,h_u]} \sup_{z\in(0,1)} \frac{\sqrt{nh}}{\sqrt{\log(d/h) \vee \log\left(\delta^{-1}\log\left(h_u h_l^{-1}\right)\right)}} \left|\widehat{\tau}_{jk}(z) - \mathbb{E}\left[\widehat{\tau}_{jk}(z)\right]\right| \le C.$$

**Lemma 10 (Bias of Kendall's tau estimator)** *Assume that the bandwidths $0 < h_l < h_u < 1$ satisfy $h_u = o(1)$. There exists a constant $C > 0$ such that*

$$\sup_{j,k\in[d]} \sup_{h\in[h_l,h_u]} \sup_{z\in(0,1)} \frac{|\mathbb{E}[\widehat{\tau}_{jk}(z)] - \tau_{jk}(z)|}{h^2 + 1/(nh)} \le C.$$

We defer the proof of these two lemmas to Appendix D.

By the definition of $\widehat{\boldsymbol{\Sigma}}(z)$ in (7), for any $j, k \in [d]$ and $z \in (0,1)$, we have

$$\left|\widehat{\boldsymbol{\Sigma}}_{jk}(z) - \boldsymbol{\Sigma}_{jk}(z)\right| = \left|\sin\left(\pi\widehat{\tau}_{jk}(z)/2\right) - \sin\left(\pi\tau_{jk}(z)/2\right)\right| \le \frac{\pi}{2}\left|\widehat{\tau}_{jk}(z) - \tau_{jk}(z)\right|, \tag{38}$$

where the last inequality is due to $|\sin(x) - \sin(y)| \le |x - y|$ for any $x, y \in [-\pi/2, \pi/2]$. Therefore, the rate of $\widehat{\boldsymbol{\Sigma}}(z)$ can be bounded by the rate of $\widehat{\tau}_{jk}(z)$ up to a constant. Recall that $\mathbf{T} = [\tau_{jk}]_{jk}$ and $\widehat{\mathbf{T}} = [\widehat{\tau}_{jk}]_{jk}$. We have

$$\|\widehat{\mathbf{T}}(z) - \mathbf{T}(z)\|_{\max} \le \sup_{j,k\in[d]} |\widehat{\tau}_{jk}(z) - \mathbb{E}\widehat{\tau}_{jk}(z)| + \sup_{j,k\in[d]} |\mathbb{E}\widehat{\tau}_{jk}(z) - \tau_{jk}(z)|.$$

Lemma 9 and Lemma 10 together with (38) give us

$$\begin{aligned}
&\sup_{h\in[h_l,h_u]} \sup_{z\in(0,1)} \frac{\left\|\widehat{\boldsymbol{\Sigma}}(z) - \boldsymbol{\Sigma}(z)\right\|_{\max}}{h^2 + \sqrt{(nh)^{-1}\left[\log(d/h) \vee \log\left(\delta^{-1}\log\left(h_u h_l^{-1}\right)\right)\right]}} \\
&\le \sup_{h\in[h_l,h_u]} \sup_{z\in(0,1)} \frac{\left\|\widehat{\mathbf{T}}(z) - \mathbf{T}(z)\right\|_{\max}}{h^2 + \sqrt{(nh)^{-1}\left[\log(d/h) \vee \log\left(\delta^{-1}\log\left(h_u h_l^{-1}\right)\right)\right]}} \le 2\pi(C_1 + C_2),
\end{aligned}$$

with probability $1-\delta$, since $1/(nh) = o(1)$. We complete the proof of the theorem by setting $C_{\boldsymbol{\Sigma}} = 2\pi(C_1 + C_2)$.

### A.2 Proof of Theorem 6

Using the uniform rate of convergence of $\widehat{\mathbf{\Sigma}}(z)$ established in Theorem 5, we establish the corresponding rate for $\widehat{\mathbf{\Omega}}(z)$ when estimated using the CLIME (Cai et al., 2011) or the calibrated CLIME (Zhao and Liu, 2014). Modifying the proofs in the above two papers, we can establish a bound on $\big\|\widehat{\mathbf{\Omega}}-\mathbf{\Omega}\big\|_{\max}$, if $\big\|\widehat{\mathbf{\Sigma}}-\mathbf{\Sigma}\big\|_{\max}$ is controlled. For simplicity, we recall the results for the calibrated CLIME estimator. Similar results for the CLIME estimator can be found in the proof of Theorem 6 in Cai et al. (2011).

**Theorem 11 (Adapted from Zhao and Liu 2014)** *Suppose $\mathbf{\Omega} \in \mathcal{U}_s(M,\rho)$ and the tuning parameter satisfies $s\lambda = o(1)$. On the event $\big\{\|\widehat{\mathbf{\Sigma}}-\mathbf{\Sigma}\|_{\max} \le \lambda\big\}$, there exist universal constants $C_1, C_2, C_3$ such that the output of the calibrated CLIME satisfies*

$$\big\|\widehat{\mathbf{\Omega}}-\mathbf{\Omega}\big\|_{\max} \le C_1 M^2\lambda, \quad \big\|\widehat{\mathbf{\Omega}}-\mathbf{\Omega}\big\|_1 \le C_2 sM\lambda \quad \textit{and} \quad \max_{j\in[d]} \|\widehat{\mathbf{\Sigma}}\widehat{\mathbf{\Omega}}_j - \mathbf{e}_j\|_\infty \le C_3\lambda M.$$

The formal statements of Theorem IV.1 and Theorem IV.2 in Zhao and Liu (2014) are not the same as the statement above. The result of Theorem 11 is more general and directly follows from the proofs of Theorem IV.1 and Theorem IV.2 in Zhao and Liu (2014). For example, the last inequality in Theorem 11 follows from Equation (E.12) of Zhao and Liu (2014).

Let $\lambda_{n,h} = C_{\mathbf{\Sigma}}\big(h^2 + \sqrt{\log(dn)/(nh)}\,\big)$ and define the event

$$\mathcal{E} = \left\{ \sup_{h\in[h_l,h_u]} \sup_{z\in(0,1)} \lambda_{n,h}^{-1} \big\|\widehat{\mathbf{\Sigma}}(z) - \mathbf{\Sigma}(z)\big\|_{\max} \le 1 \right\}.$$

Since the constants $C_1, C_2$ and $C_3$ in Theorem 11 are universal and the penalty parameter $\lambda \ge \lambda_{n,h}$, it follows that

$$\sup_{h\in[h_l,h_u]} \sup_{z\in(0,1)} \lambda^{-1}\big\|\widehat{\mathbf{\Omega}}(z) - \mathbf{\Omega}(z)\big\|_{\max} \le C_1 M^2;$$
$$\sup_{h\in[h_l,h_u]} \sup_{z\in(0,1)} \lambda^{-1}\big\|\widehat{\mathbf{\Omega}}(z) - \mathbf{\Omega}(z)\big\|_1 \le C_2 sM;$$
$$\sup_{z\in(0,1)} \max_{j\in[d]} \lambda^{-1} \cdot \|\widehat{\mathbf{\Omega}}_j^T \widehat{\mathbf{\Sigma}} - \mathbf{e}_j\|_\infty \le C_3 M,$$

on the event $\mathcal{E}$. Theorem 5 gives us $\mathbb{P}(\mathcal{E}) \ge 1 - 1/d$, which completes the proof.

## Appendix B. Asymptotic Properties of Testing Statistics

In this section, we prove asymptotic properties of the testing statistics for three kinds of hypothesis tests: (1) edge presence test in Theorem 1, (2) super-graph test in Theorem 3 and (3) uniform edge presence test in Theorem 4.

Let $\mathcal{S}_{j,z} = \{k \in [d] \mid \mathbf{\Omega}_{kj}(z) \ne 0\}$. For any index sets $\mathcal{S}, \mathcal{S}' \subset [d]$, we define $\mathbf{\Omega}_{\mathcal{S}\mathcal{S}'} \in \mathbb{R}^{|\mathcal{S}|\times|\mathcal{S}'|}$ to be the submatrix of $\mathbf{\Omega}$ obtained from rows indexed by $\mathcal{S}$ and columns indexed by $\mathcal{S}'$. For any function $f(x)$, we define

$$\mathbb{G}_n^\xi[f] = \frac{1}{\sqrt{n}} \sum_{i=1}^n f(X_i)\cdot \xi_i, \tag{39}$$

where $\xi_1, \ldots, \xi_n \sim N(0,1)$. We also define $\mathbb{P}_\xi(\cdot) := \mathbb{P}(\cdot | \{Y_i\}_{i\in[n]})$ and $\mathbb{E}_\xi[\cdot] := \mathbb{E}[\cdot|\{Y_i\}_{i\in[n]}]$.

At a high-level, we establish the asymptotic results by carefully studying the Hoeffding decomposition of the kernel smoothed Kendall's tau estimator. From (4), we observe that $\widehat{\tau}_{jk}(z)$ is a quotient of two $U$-statistics. To study this quotient, we introduce some additional notation. For a fixed $j,k \in [d]$, we define the following bivariate function

$$g_{z|(j,k)}(y_i, y_{i'}) = \omega_z(z_i, z_{i'}) \operatorname{sign}(x_{ij} - x_{i'j}) \operatorname{sign}(x_{ik} - x_{i'k}), \tag{40}$$

for $y_i = (z_i, \boldsymbol{x}_i)$ and $y_{i'} = (z_{i'}, \boldsymbol{x}_{i'})$. Recalling the definition of $\omega_z(z_i, z_{i'})$ in (5), we have that

$$\widehat{\tau}_{jk}(z) = \mathbb{U}_n[g_{z|(j,k)}]/\mathbb{U}_n[\omega_z].$$

Let us define the Hoeffding decomposition of $g_{z|(j,k)}$ as

$$g^{(1)}_{z|(j,k)}(y) = \mathbb{E}[g_{z|(j,k)}(y, Y)] - \mathbb{E}\left[\mathbb{U}_n[g_{z|(j,k)}]\right], \tag{41}$$

$$g^{(2)}_{z|(j,k)}(y_1, y_2) = g_{z|(j,k)}(y_1, y_2) - g^{(1)}_{z|(j,k)}(y_1) - g^{(1)}_{z|(j,k)}(y_2) - \mathbb{E}\left[\mathbb{U}_n[g_{z|(j,k)}]\right]. \tag{42}$$

Then we can reformulate the centered $U$-statistic as

$$\mathbb{U}_n\left[g_{z|(j,k)}\right] - \mathbb{E}\left[\mathbb{U}_n\left[g_{z|(j,k)}\right]\right] = 2\mathbb{E}_n\left[g^{(1)}_{z|(j,k)}(Y_i)\right] + \mathbb{U}_n\left[g^{(2)}_{z|(j,k)}\right]. \tag{43}$$

In the above display, we decomposed the centered $U$-statistic into an empirical process and a higher order $U$-statistic. Our proof strategy is to study the asymptotic property of the leading empirical process term $2\mathbb{E}_n\left[g^{(1)}_{z|(j,k)}(Y_i)\right]$ and show that the higher order term can be ignored. Similarly, let

$$\omega^{(1)}_z(s) = \mathbb{E}[\omega_z(s, Z)] - \mathbb{E}\left[\mathbb{U}_n[\omega_z]\right], \tag{44}$$

$$\omega^{(2)}_z(s,t) = \omega_z(s,t) - \omega^{(1)}_z(s) - \omega^{(1)}_z(t) - \mathbb{E}\left[\mathbb{U}_n[\omega_z]\right], \tag{45}$$

which leads to the following Hoeffding decomposition

$$\mathbb{U}_n[\omega_z] - \mathbb{E}\left[\mathbb{U}_n[\omega_z]\right] = 2\mathbb{E}_n\left[\omega^{(1)}_z\right] + \mathbb{U}_n\left[\omega^{(2)}_z\right], \tag{46}$$

According to the heuristic approximation of $\widehat{S}_{z|(j,k)}\big(\widehat{\boldsymbol{\Omega}}_{k\setminus j}(z_0)\big)$ in (10), we have

$$\begin{aligned}
&\widehat{S}_{z|(j,k)}\big(\widehat{\boldsymbol{\Omega}}_{k\setminus j}(z)\big) \approx \boldsymbol{\Omega}_j^T(z)\big[\dot{\boldsymbol{\Sigma}}(z) \circ \big(\widehat{\mathbf{T}}(z) - \mathbf{T}(z)\big)\big]\boldsymbol{\Omega}_k(z) \\
&\approx [\mathbb{U}_n[\omega_z]]^{-1} \sum_{u,v\in[d]} \boldsymbol{\Omega}_{ju}(z)\boldsymbol{\Omega}_{kv}(z)\pi\cos\left(\tau_{uv}(z)\frac{\pi}{2}\right) \cdot \left[\mathbb{U}_n[g_{z|(u,v)}] - \mathbb{E}[\mathbb{U}_n[g_{z|(u,v)}]] \cdot \mathbb{U}_n[\omega_z]\right],
\end{aligned} \tag{47}$$

where the last "$\approx$" comes from (4) and Lemma 10. Combining (43) and (46) with (47),

$$\sqrt{nh}\cdot\widehat{S}_{z|(j,k)} \approx [\mathbb{U}_n[\omega_z]]^{-1}\mathbb{G}_n\big[J_{z|(j,k)}\big],$$

where the leading term of the Hoeffding decomposition is defined as

$$J_{z|(j,k)}(y') := \sum_{u,v\in[d]} \boldsymbol{\Omega}_{ju}(z)\boldsymbol{\Omega}_{kv}(z)\pi\cos\left(\tau_{uv}(z)\frac{\pi}{2}\right)\sqrt{h}\cdot\left[g^{(1)}_{z|(u,v)}(y') - \tau_{uv}(z)\omega^{(1)}_z(z')\right], \tag{48}$$

for any $y' = (z', \boldsymbol{x}')$. We find the asymptotic distribution of $[\mathbb{U}_n[\omega_z]]^{-1}\mathbb{G}_n\big[J_{z|(j,k)}\big]$ in the next part.

### B.1 Proof of Theorem 1

Let the operator $u_n[\cdot]$ be defined as

$$u_n[H] = \sqrt{n} \cdot \big(\mathbb{U}_n[H] - \mathbb{E}[\mathbb{U}_n[H]]\big) \tag{49}$$

for any bivariate function $H(x, x')$. In order to prove Theorem 1, we need the convergence rate of terms related to the Kendall's tau estimator, especially the two lemmas below.

**Lemma 12** *Suppose that $n^{-1}\log d = o(1)$ and the bandwidths $0 < h_l < h_u < 1$ satisfy $h_l n/\log(dn) \to \infty$ and $h_u = o(1)$. There exists a universal constant $C > 0$ such that with probability $1-\delta$,*

$$\sup_{j,k\in[d]} \sup_{h\in[h_l,h_u]} \sup_{z\in(0,1)} \left| \frac{\sqrt{h}}{\sqrt{\log(d/h) \vee \log\big(\delta^{-1}\log\big(h_u h_l^{-1}\big)\big)}} \Big( u_n\left[\omega_z\right] \vee u_n\left[g_{z|(j,k)}\right] \Big) \right| \le C, \tag{50}$$

*for large enough $n$.*

**Lemma 13** *Suppose the bandwidths $0 < h_l < h_u < 1$ satisfy $h_u = o(1)$. Then*

$$\sup_{z\in(0,1)} \left|\mathbb{E}\left[\mathbb{U}_n\left[g_{z|(j,k)}\right]\right] - f_Z^2(z)\tau_{jk}(z)\right| = O(h^2), \tag{51}$$

$$\sup_{z\in(0,1)} \left|\mathbb{E}\left[\mathbb{U}_n\left[\omega_z\right]\right] - f_Z^2(z)\right| = O(h^2), \tag{52}$$

$$\sup_{z\in(0,1)} n^{-1}\mathbb{E}\left[u_n\left[g_{z|(j,k)}\right] \cdot u_n\left[\omega_z\right]\right] = O\big((nh)^{-1}\big), \tag{53}$$

$$\sup_{z\in(0,1)} n^{-1}\mathbb{E}\big[\left(u_n\left[\omega_z\right]\right)^2\big] = O\big((nh)^{-1}\big). \tag{54}$$

We defer the proof of the above two lemmas to Appendix E in the supplementary material.

Using Lemma 12 and Lemma 13, we have

$$\begin{aligned} \inf_{z\in(0,1)} \mathbb{U}_n[\omega_z] &\ge \inf_{z\in(0,1)} \mathbb{E}[\mathbb{U}_n[\omega_z]] - \sup_{z\in(0,1)} n^{-1/2}|u_n[\omega_z]| \ge \underline{\mathrm{f}}_Z^2/2, \\ \sup_{z\in(0,1)} \mathbb{U}_n[\omega_z] &\le \sup_{z\in(0,1)} \mathbb{E}[\mathbb{U}_n[\omega_z]] + \sup_{z\in(0,1)} n^{-1/2}|u_n[\omega_z]| \le 2\overline{\mathrm{f}}_Z^2, \end{aligned} \tag{55}$$

with probability $1 - 1/d$ for sufficiently large $n$. The last inequality is due to the fact that $f_Z$ is bounded from above and below, $h = o(1)$ and $\log(1/h)/nh = o(1)$.

Combining the above display with Lemma 19, we have

$$\begin{aligned} &\left|\sqrt{nh} \cdot \widehat{S}_{z|(j,k)}\big(\widehat{\mathbf{\Omega}}_{k\setminus j}(z)\big) - [\mathbb{U}_n[\omega_z]]^{-1}\mathbb{G}_n\left[J_{z|(j,k)}\right]\right| \\ &\qquad \le \left|\sqrt{nh} \cdot \mathbb{U}_n[\omega_z]\widehat{S}_{z|(j,k)}\big(\widehat{\mathbf{\Omega}}_{k\setminus j}(z)\big) - \mathbb{G}_n\left[J_{z|(j,k)}\right]\right| \cdot \Big(\inf_{z\in(0,1)} \mathbb{U}_n[\omega_z]\Big)^{-1} \le 2\underline{\mathrm{f}}_Z^{-2} n^{-c}. \end{aligned} \tag{56}$$

Therefore, it suffices to derive the limiting distribution of $\mathbb{G}_n\left[J_{z|(j,k)}\right]$.

In order to apply central limit theorem to $\mathbb{G}_n\big[J_{z|(j,k)}\big]$ we check Lyapunov's condition. By the definition of $g^{(1)}_{z|(j,k)}$ in (41) and $\omega^{(1)}_z$ in (44), we have $\mathbb{E}[J_{z|(j,k)}(Y_i)] = 0$ for all $i \in [n]$. The matrix $\mathbf{\Theta}_z$ defined in (11) can be rewritten as

$$(\mathbf{\Theta}_z)_{jk} = \pi\cos\left(\tau_{jk}(z)\frac{\pi}{2}\right)\sqrt{h}\cdot\left[g^{(1)}_{z|(j,k)}(Y) - \tau_{jk}(z)\omega^{(1)}_z(Z)\right]. \tag{57}$$

In order to apply the Lyapunov condition to show the asymptotic normality, we begin to control the third moments of $\omega^{(1)}_z$ and $g^{(1)}_{z|(j,k)}$. We bound the third moment of $\omega^{(1)}_z(Z)$ by

$$\begin{aligned}
\sup_z \mathbb{E}\Big[\big|\omega^{(1)}_z(Z)\big|^3\Big] &= \sup_z |\mathbb{E}[K_h(z-Z)]|^3\mathbb{E}\Big[\big|K_h(z-Z) - \mathbb{E}[K_h(z-Z)]\big|^3\Big] \\
&\le \sup_z 8|\mathbb{E}[K_h(z-Z)]|^3\mathbb{E}\Big[\big|K_h(z-Z)\big|^3\Big] \\
&= \sup_z 8\big|f_Z(z) + O(h^2)\big|^3\cdot h^{-2}\int |K^3(t)|f_Z(z+th)dt \\
&\lesssim h^{-2}\cdot \bar{\mathrm{f}}_Z^4\int |K^3(t)|dt,
\end{aligned} \tag{58}$$

where we used that $(1+x)^3 \le 4(1+x^3)$ for $x>0$ and $|\mathbb{E}[K_h(z-Z)]|^3 \le \mathbb{E}[|K_h(z-Z)|^3]$. By the definition of $g^{(1)}_{z|(j,k)}$, we also have

$$\mathbb{E}\Big[\big|g^{(1)}_{z|(j,k)}(Y)\big|^3\Big] \le 4\mathbb{E}\Big[\big|\mathbb{E}[g_{z|(j,k)}(Y',Y)\mid Y']\big|^3\Big] + 4\Big|\mathbb{E}\big[g_{z|(j,k)}(Y',Y)\big]\Big|^3, \tag{59}$$

where $Y'$ is an independent copy of $Y$. Using (99),

$$\sup_z \Big|\mathbb{E}\big[g_{z|(j,k)}(Y',Y)\big]\Big|^3 = \sup_z \big|f_Z^2(z)\tau_{jk}(z) + O(h^2)\big|^3 \lesssim \bar{\mathrm{f}}_Z^6. \tag{60}$$

We now bound the conditional expectation in (59). From (A.3) of Mitra and Zhang (2014), denoting $\boldsymbol{x}' = (x'_1,\ldots,x'_d)^T$ and $y' = (z',\boldsymbol{x}')$, we have

$$\begin{aligned}
\mathbb{E}\left[g_{z|(j,k)}(y',Y)|Z=s\right] &= K_h(z'-z)K_h(s-z)\varphi\left(x'_j,x'_k,\mathbf{\Sigma}_{jk}(s)\right), \text{ where} \\
\varphi(u,v,\rho) &= 2\int \operatorname{sign}(u-x)\phi(x)\cdot\Phi\left(\frac{v-\rho x}{\sqrt{1-\rho^2}}\right)dx,
\end{aligned} \tag{61}$$

with $\phi(\cdot)$ and $\Phi(\cdot)$ being the probability density and cumulative distribution function of a standard normal variable, respectively. From (41), we have

$$\begin{aligned}
\mathbb{E}\left[g_{z|(j,k)}(y',Y)\right] &= \mathbb{E}\big[\mathbb{E}\left[g_{z|(j,k)}(y',Y)|Z\right]\big] \\
&= K_h(z'-z)\int K_h(s-z)\varphi\left(x'_j,x'_k,\mathbf{\Sigma}_{jk}(s)\right)f_Z(s)ds.
\end{aligned} \tag{62}$$

Let $\phi_\rho(x,y)$ be the density function of bivariate normal distribution with mean zero, variance one and correlation $\rho$. Notice that $\sup_{x,y,\rho}|\varphi(x,y,\rho)| \le 2$. Since the minimum eigenvalue of $\mathbf{\Sigma}(z)$ is strictly positive for any $z$, there exists a $\gamma_\sigma < 1$ such that $\sup_z |\mathbf{\Sigma}_{jk}(z)| \le \gamma_\sigma < 1$ for any $j \ne k$. We also have $\sup_{x,y,\rho} |\phi_\rho(x,y)| \le (2\pi\sqrt{1-\gamma_\sigma^2})^{-1}$. By (62), for any $z \in (0,1)$

$$
\begin{aligned}
&\mathbb{E}\Big[\big|\mathbb{E}[g_{z|(j,k)}(Y',Y) \mid Y']\big|^3\Big] \\
&= \frac{2}{\pi}\int h^{-2}K^3(t_1)f_Z(z+t_1h)\times \\
&\qquad \times \bigg|\int K(t_2)f_Z(z+t_2h)\varphi\left(u,v,\mathbf{\Sigma}_{jk}(z+t_2h)\right)dt_2\bigg|^3 \phi_{\mathbf{\Sigma}_{jk}(z+t_1h)}(u,v)dt_1 \\
&\lesssim \frac{1}{h^2\pi\sqrt{1-\gamma_\sigma^2}}\int K^3(t_1)f_Z(z+t_1h)|f_Z(z)+O(h^2)|^3dt_1 \le h^{-2}\cdot\frac{\bar{\mathrm{f}}_Z^4\|K\|_3^3}{\pi\sqrt{1-\gamma_\sigma^2}}.
\end{aligned} \tag{63}
$$

Combining (60), (63) with (59), we have

$$
\mathbb{E}\Big[\big|g^{(1)}_{z|(j,k)}(Y)\big|^3\Big] \lesssim h^{-2}\left(\frac{\bar{\mathrm{f}}_Z^4\|K\|_3^3}{\pi\sqrt{1-\gamma_\sigma^2}}\right). \tag{64}
$$

By the assumption of Theorem 4, there exists a $\theta_{\min} > 0$ such that

$$
\mathrm{Var}(J_{z_0|(j,k)}(Y)) = \mathbb{E}(\mathbf{\Omega}_j^T(z_0)\mathbf{\Theta}_{z_0}\mathbf{\Omega}_k(z_0))^2 \ge \theta_{\min}\|\mathbf{\Omega}_j(z_0)\|_2^2\|\mathbf{\Omega}_k(z_0)\|_2^2. \tag{65}
$$

We are now ready to check the Lyapunov's condition. We have

$$
\begin{aligned}
\frac{\sum_{i=1}^n \mathbb{E}|J_{z_0|(j,k)}(Y_i)|^3}{n^{3/2}\mathrm{Var}^{3/2}(J_{z_0|(j,k)}(Y))} &\overset{(65)}{\le} \frac{(\theta_{\min}n)^{-3/2}}{\|\mathbf{\Omega}_j^T\|_2^3\|\mathbf{\Omega}_k\|_2^3}\sum_{i=1}^n \mathbb{E}|J_{z_0|(j,k)}(Y_i)|^3 \\
&\lesssim \frac{1}{n^{3/2}}\sum_{i=1}^n \mathbb{E}\|\operatorname{Vec}((\mathbf{\Theta}_{z_0})_{\mathcal{S}_{j,z_0},\mathcal{S}_{k,z_0}})\|_2^3.
\end{aligned} \tag{66}
$$

Since $|\mathbf{\Theta}_{jk}^{(i)}| \le \pi\sqrt{h}|g^{(1)}_{z_0|(j,k)}(Y)| + \pi\sqrt{h}|\tau_{jk}(z_0)\omega^{(1)}_{z_0}(Z)|$, we have

$$
\begin{aligned}
&\mathbb{E}\|\operatorname{Vec}((\mathbf{\Theta}_{z_0})_{\mathcal{S}_{j,z_0},\mathcal{S}_{k,z_0}})\|_2^3 \\
&\qquad \le |\mathcal{S}_{j,z_0}|^{3/2}|\mathcal{S}_{k,z_0}|^{3/2}\pi^3h^{3/2}\Big(\mathbb{E}\Big[\big|g^{(1)}_{z_0|(j,k)}(Y)\big|^3\Big] + |\tau_{jk}(z_0)|^3\mathbb{E}\Big[\big|\omega^{(1)}_{z_0}(Z)\big|^3\Big]\Big).
\end{aligned}
$$

Using (58) and (64), together with $s^3/\sqrt{nh} = o(1)$,

$$
\frac{\sum_{i=1}^n \mathbb{E}|J_{z_0|(j,k)}(Y_i)|^3}{n^{3/2}\mathrm{Var}^{3/2}(J_{z_0|(j,k)}(Y))} \lesssim \frac{s^3}{\sqrt{nh}} = o(1),
$$

which implies that the Lyapunov's condition is satisfied. Moreover, by Lemma 12, for any $z_0 \in (0,1)$, $\mathbb{U}_n[\omega_{z_0}] - \mathbb{E}[\mathbb{U}_n[\omega_{z_0}]]$ converges to 0 in probability. Combining this with (52) and $h = o(1)$, we have that $\mathbb{U}_n[\omega_{z_0}]$ converges to $f_Z^2(z_0)$ in probability. Therefore, by the central limit theorem and Slutsky's theorem, for any $j,k \in [d]$,

$$
\frac{[\mathbb{U}_n[\omega_{z_0}]]^{-1}\mathbb{G}_n\left[J_{z_0|(j,k)}\right]}{f_Z^{-2}(z_0)\left\{\mathbb{E}\left[\left(\mathbf{\Omega}_j^T\mathbf{\Theta}_{z_0}\mathbf{\Omega}_k\right)^2\right]\right\}^{1/2}} \rightsquigarrow N(0,1).
$$

Combining with (56), the proof is complete.

### B.2 Proof of Theorem 3

The strategy is to apply the theory for multiplier bootstrap developed in Chernozhukov et al. (2013) to the score function in (16). A similar strategy is applied to prove Theorem 4, whose proof is deferred to Section F.

Let $T_0(z) := \max_{(j,k)\in E^c} \mathbb{G}_n[J_{z|(j,k)}]$ and

$$S_0^B(z_0) = \max_{(j,k)\in E^c} \mathbb{G}_n^{\xi}\left[J_{z_0|(j,k)}\right] = \max_{(j,k)\in E^c} \frac{1}{\sqrt{n}} \sum_{i=1}^{n} J_{z_0|(j,k)}(Y_i)\cdot \xi_i$$

be the bootstrap counterpart to $T_0(z_0)$. Recall that $S^B(z)$ is defined in (19). We denote

$$\Delta_z := \max_{(j,k),(j',k')\in E^c} \left| \frac{1}{n}\sum_{i=1}^{n} \Big( J_{z|(j,k)}(Y_i) J_{z|j'k'}(Y_i) - \mathbb{E}[J_{z|(j,k)}(Y_i) J_{z|j'k'}(Y_i)] \Big) \right|.$$

In order to Use Theorem 3.2 in Chernozhukov et al. (2013), we check four conditions.

1. With probability $1-1/d$,

$$|S(z_0) - T_0(z_0)| \le \sup_{j,k\in[d]} \left| \sqrt{nh}\cdot \mathbb{U}_n[\omega_{z_0}] \widehat{S}_{z_0|(j,k)}\left(\widehat{\mathbf{\Omega}}_{k\setminus j}(z_0)\right) - \mathbb{G}_n\left[J_{z_0|(j,k)}\right] \right| \le n^{-c}.$$

2. With probability $1-1/d$, $\mathbb{P}_{\xi}\big(|S^B(z_0) - S_0^B(z_0)| \le n^{-c}\big) \ge 1-1/d$.

3. There exists a constant $c>0$, such that $\mathrm{Var}(\mathbb{G}_n\big[J_{z_0|(j,k)}\big]) > c$.

4. There exists a constant $c>0$, such that $\mathbb{P}(\Delta_{z_0} > n^{-c}) \le n^{-c}$.

The first condition is proven in Lemma 19. We defer the proof of the second condition in Lemma 20. The third condition is due to (65). The following of the proof verifies the last condition.

Define

$$\gamma_{z|(j,k,j',k')}(Y_i) = J_{z|(j,k)}(Y_i) J_{z_0|(j',k')}(Y_i) - \mathbb{E}[J_{z|(j,k)}(Y_i) J_{z|(j',k')}(Y_i)]. \tag{67}$$

We will apply Lemma A.1 in van de Geer (2008) on the concentration of empirical processes. For the self-consistence, we has restated the lemma in Lemma 34. In order to apply Lemma 34, we need to bound $\|\gamma_{z|(j,k,j',k')}\|_\infty$ and $n^{-1}\sum_{i=1}^n \mathbb{E}[\gamma^2_{z|(j,k,j',k')}(Z_i)]$. By the definition of $J_{z_0|(j,k)}$ in (48), we have for any $z_0\in(0,1)$,

$$\begin{aligned}
&\max_{(j,k),(j',k')\in E^c} \|\gamma_{z_0|(j,k,j',k')}\|_\infty \\
&\qquad \le \max_{i\in[n]} \max_{(j,k)\in E^c} 2|J_{z_0|(j,k)}(Y_i)|^2 \\
&\qquad \le \max_{(j,k)\in E^c} 2\|\mathbf{\Omega}_j(z_0)\|_1^2 \|\mathbf{\Omega}_k(z_0)\|_1^2 \cdot \pi^2 h\cdot \Big( \|g^{(1)}_{z_0|(j,k)}(Y_i)\|_\infty + \|\omega^{(1)}_{z_0}(Z_i)\|_\infty \Big)^2 \\
&\qquad \le CM^2 h^{-1},
\end{aligned} \tag{68}$$

where the second inequality follows from Hölder's inequality, similar to (66), and the final inequality is due to (84) and (94). Since the right hand size of (68) does not depend on $z_0$, we also have

$$\max_{z\in(0,1)} \max_{(j,k),(j',k')\in E^c} \|\gamma_{z|(j,k,j',k')}\|_\infty \le CM^2h^{-1} \text{ and } \tag{69}$$

$$\max_{z\in(0,1)} \max_{(j,k),(j',k')\in E^c} \mathbb{E}[\gamma^2_{z|(j,k,j',k')}(Z_i)] \le \max_{z\in(0,1)} \max_{(j,k),(j',k')\in E^c} \|\gamma_{z|(j,k,j',k')}\|^2_\infty \le CM^4h^{-2}. \tag{70}$$

According to Lemma 34, the expectation of $\Delta_{z_0}$ is bounded by

$$\mathbb{E}[\Delta_{z_0}] \lesssim \sqrt{\frac{2M^4\log(2d)}{nh^2}} + \frac{M^2\log(2d)}{nh}.$$

Since $\log d/(nh^2) = o(n^{-\epsilon})$, there exists $c_1 > 0$ such that $\mathbb{E}[\Delta_{z_0}] \le n^{-2c_1}$ for sufficiently large $n$. By Markov's inequality, $\mathbb{P}(\Delta_{z_0} > n^{-c_1}) \le n^{c_1}\mathbb{E}[\Delta_{z_0}] \le n^{-c_1}$ for sufficiently large $n$, which verifies the last condition.

By Theorem 3.2 in Chernozhukov et al. (2013),

$$\sup_{\alpha\in(0,1)} \Big|\mathbb{P}_{H_0}\big(\psi_{z_0|G}(\alpha) = 1\big) - \alpha\Big| \lesssim n^{-c},$$

for some constant $c > 0$, which completes the proof.

## Appendix C. Proof of Theorem 8

In this section, we prove the minimax rate of convergence for estimating time-varying inverse covariance matrices. Section C.1 proves the minimax rate in terms of $\|\cdot\|_{\max}$ norm, while Section C.2 establishes the minimax rate for the $\|\cdot\|_1$ norm.

At a high-level, both results will use Le Cam's lemma applied to a finite collection of time-varying inverse covariance matrices. Given a time-varying inverse covariance matrix $\boldsymbol{\Omega}(\cdot)$, let $\mathbb{P}_{\boldsymbol{\Omega}}$ be the joint distribution of $(\boldsymbol{X}_1, Z_1), \ldots, (\boldsymbol{X}_n, Z_n)$ where $(\boldsymbol{X}_i, Z_i)$ are independent copies of $(\boldsymbol{X}, Z)$ with $Z \sim \text{Unif}((0,1))$ and $\boldsymbol{X} \mid Z \sim N(0, \boldsymbol{\Omega}(z)^{-1})$. Let $\mathcal{U}_0 = \{\boldsymbol{\Omega}_0(\cdot), \boldsymbol{\Omega}_1(\cdot), \ldots, \boldsymbol{\Omega}_m(\cdot)\}$ be a collection of time-varying inverse covariance matrices, which are going to be defined later. With these we define the mixture distribution $\overline{\mathbb{P}} = m^{-1}\sum_{\ell=1}^m \mathbb{P}_{\boldsymbol{\Omega}_\ell}$. For two measures $\mathbb{P}$ and $\mathbb{Q}$, the total variation is given as $\|\mathbb{P} \wedge \mathbb{Q}\| := \int (d\mathbb{P}/d\mu) \wedge (d\mathbb{Q}/d\mu)d\mu$, where $d\mu$ is the Lebesgue measure. Now, Le Cam's lemma (Le Cam, 1973) gives us the following lower bound.

**Lemma 14** *Let $\widehat{\boldsymbol{\Omega}}(\cdot)$ be any estimator of $\boldsymbol{\Omega}(\cdot)$ based on the data generated from the distribution family $\{\mathbb{P}_{\boldsymbol{\Omega}} \mid \boldsymbol{\Omega} \in \mathcal{U}_0\}$. Then*

$$\max_{1\le\ell\le m} \mathbb{E}\Big[\sup_{z\in(0,1)} \|\widehat{\boldsymbol{\Omega}}(z) - \boldsymbol{\Omega}_\ell(z)\|_{\max}\Big] \ge r_{\min}\|\overline{\mathbb{P}} \wedge \mathbb{P}_{\boldsymbol{\Omega}_0}\|,$$

*where* $r_{\min} = \min_{1\le\ell\le m}\sup_{z\in(0,1)} \|\boldsymbol{\Omega}_0(z) - \boldsymbol{\Omega}_\ell(z)\|_{\max}$.

We will use the above lemma in the following two subsections.

### C.1 Proof of Maximum Norm in (36)

We start by constructing the collection of inverse covariance matrices $\mathcal{U}_0$. Let $\mathbf{\Omega}_0(\cdot) \equiv \mathbf{I}$. Let $M_0 = \lceil c_0(n/\log(dn))^{1/5}\rceil$ where $c_0$ is some constant to be determined. Then

$$\mathcal{U}_0 = \Big\{\mathbf{\Omega}_{(j,m)}(\cdot) \mid \mathbf{\Omega}^{-1}_{(j,m)}(z) = \mathbf{\Sigma}_{(j,m)}(z) = \mathbf{I} + \tau_m(z)\mathbf{E}_{jj}, z \in (0,1), j \in [d-1], m \in [M_0]\Big\},$$

where $\mathbf{E}_{jj} = \mathbf{e}_j\mathbf{e}_{j+1}^T + \mathbf{e}_{j+1}\mathbf{e}_j^T$, $\mathbf{e}_j$ is the $j$-th canonical basis in $\mathbb{R}^d$ and for any $m \in [M_0]$,

$$\tau_m(z) = Lh^2 K_0\Big(\frac{z - z_m}{h}\Big), \quad z_m = \frac{m - 1/2}{M_0}, \quad h = 1/M_0. \tag{71}$$

Here, $K_0(\cdot)$ is any function supported on $(-1/2, 1/2)$ and satisfying $\|K_0\|_{\sup} \le K_{\max}$. For example, consider $K_0(\cdot) = a\psi(2z)$, where $\psi(z) = \exp(-1/(1-z^2))\mathbb{1}(|z| \le 1)$, for some sufficiently small $a$ (Tsybakov, 2009). It is easy to check that $\mathcal{U}_0 \subset \overline{\mathcal{U}}_s(M, \rho, L)$ if $n^{-1}\log(dn) = o(1)$.

For any $j \in [d-1]$, $m \in [M_0]$, we have

$$\sup_{z\in(0,1)} \|\mathbf{\Omega}_{(j,m)}(z) - \mathbf{\Omega}_0(z)\|_{\max} \ge \Big\|\frac{\tau_m(\cdot)}{1 - \tau_m^2(\cdot)}\Big\|_{\sup} \ge \|\tau_m\|_{\sup} \ge Lh^2 K_0(0)$$

by direct calculation, which gives us $r_{\min} \ge Lh^2K_0(0) \asymp (\log(dn)/n)^{2/5}$.

In the remainder of the proof we show that for $\overline{\mathbb{P}} = ((d-1)M_0)^{-1}\sum_{j\in[d],m\in[M_0]}\mathbb{P}_{\mathbf{\Omega}_{(j,m)}}$, we have $\|\mathbb{P}_{\mathbf{\Omega}_0} \wedge \overline{\mathbb{P}}\| \ge 1/2$. Let $f_{jm}$ be the density of $\mathbb{P}_{\mathbf{\Omega}_{(j,m)}}$ for $j \in [d-1], m \in [M_0]$ and $f_0$ the density of $\mathbb{P}_0$. Under our setting,

$$f_{jm}((\boldsymbol{x}_i, z_i)_{i=1}^n) = \prod_{i=1}^n g_{jm}(\boldsymbol{x}_i)\mathbb{1}\{z_i \in (0,1)\},$$

where $g_{jm}$ is the density function of $N(0, \mathbf{\Sigma}_{(j,m)})$. Note that for any two densities $f$ and $f'$

$$\int (f \wedge f')d\mu = 1 - \frac{1}{2}\int |f - f'|d\mu \ge 1 - \frac{1}{2}\Big(\int \frac{f^2}{f'}d\mu - 1\Big)^{1/2}.$$

Therefore, it suffices to show that

$$\Delta := \int \Big(((d-1)M_0)^{-1}\sum_{j,m} f_{jm}\Big)^2 / f_0 \, d\mu - 1 \to 0. \tag{72}$$

Expanding the square of the mixture in (72), we have

$$\frac{1}{((d-1)M_0)^2}\Bigg\{\sum_{j=1}^{d-1}\sum_{m=1}^{M_0}\int \frac{f_{jm}^2}{f_0}d\mu + \sum_{j=1}^{d-1}\sum_{m_1\ne m_2}\int \frac{f_{jm_1}f_{jm_2}}{f_0}d\mu + \sum_{j_1\ne j_2}\sum_{m_1,m_2\in[M_0]}\int \frac{f_{j_1m_1}f_{j_2m_2}}{f_0}d\mu\Bigg\} - 1.$$

To proceed, we recall the following result of Ren et al. (2015) given in their equation (94). Let $g_i$ be the density function of $N(0, \mathbf{\Sigma}_i)$ for $i = 0, 1, 2$. Then

$$\int \frac{g_1 g_2}{g_0} = \Big[\det\Big(\mathbf{I} - \mathbf{\Sigma}_0^{-1}(\mathbf{\Sigma}_1 - \mathbf{\Sigma}_0)\mathbf{\Sigma}_0^{-1}(\mathbf{\Sigma}_2 - \mathbf{\Sigma}_0)\Big)\Big]^{-1/2}. \tag{73}$$

Using the above display, for $j_1 \neq j_2$ and $m_1, m_2 \in [M_0]$, since $(\mathbf{\Sigma}_{j_1} - \mathbf{I})(\mathbf{\Sigma}_{j_2} - \mathbf{I}) = 0$, we have $\int f_{j_1 m_1} f_{j_2 m_2}/f_0 d\mu = 1$. For $|j_1 - j_2| = 1$, we have $\int f_{j_1 m_1} f_{j_2 m_2}/f_0 d\mu = [\det(\mathbf{I}_3 - \mathbf{M}_3)]^{-n/2} = 1$, where $\mathbf{M}_3 \in \mathbb{R}^{3\times 3}$ with $(\mathbf{M}_3)_{13} = \tau_{m_1}(z)\tau_{m_2}(z)$ and the other entries are zero. For any $m_1, m_2 \in [M_0]$ and $j \in [d-1]$, we have

$$\int \frac{f_{jm_1} f_{jm_2}}{f_0} d\mu = \prod_{1\le i\le n} \int_0^1 \big(1 - \tau_{m_1}(z_i)\tau_{m_2}(z_i)\big)^{-1} dz_i \quad = \Big[\int_0^1 \big(1 - \tau_{m_1}(z)\tau_{m_2}(z)\big)^{-1} dz\Big]^n. \tag{74}$$

If $m_1 \neq m_2$, since the supports of $\tau_{m_1}(\cdot)$ and $\tau_{m_2}(\cdot)$ are disjoint, (74) implies

$$\int f_{jm_1} f_{jm_2}/f_0 d\mu = 1.$$

Finally, if $m_1 = m_2 = m$, we have

$$\begin{aligned}\int \frac{f_{jm}^2}{f_0} d\mu &\le \Big[\int_0^1 \big(1 - \tau_m^2(z)\log^{4/5}(dn)\big)^{-1} dz\Big]^n \\ &\le \Big[\frac{M_0 - 1}{M_0} + \frac{1}{M_0}(1 - L^2 h^4 K_{\max}^2)^{-1}\Big]^n = \Big[1 + \frac{L^2 h^4 K_{\max}^2}{M_0(1 - L^2 h^4 K_{\max}^2)}\Big]^n.\end{aligned} \tag{75}$$

In summary,

$$\begin{aligned}\Delta &= \frac{1}{((d-1)M_0)^2} \sum_{j=1}^{d-1} \sum_{m=1}^{M} \Big(\int \frac{f_{jm}^2}{f_0} d\mu - 1\Big) \\ &\le \exp\Big[-\log((d-1)M_0) + n\log\Big(1 + \frac{L^2 h^4 K_{\max}^2}{M_0(1 - L^2 h^4 K_{\max}^2)}\Big)\Big] - \frac{1}{(d-1)M_0}.\end{aligned}$$

Recall that $M_0 = \lceil c_0 (n/\log(dn))^{1/5} \rceil$ and $h = 1/M_0$. We choose $c_0$ sufficiently large such that $c_0^5 > 1 - L^2 K_{\max}^2$. Using $\log(1 + x) \le x$ and $x/(1 - x) \le 2x$ for $x \in (0, 1/2)$, we have

$$\Delta \le \exp\Big[-\log(d(c_0 n/\log(dn))^{1/5}) + c_0^{-5} L^2 K_{\max}^2 \log(dn)\Big] - \frac{1}{c_0 d n^{1/5}} \to 0.$$

Since $r_{\min} \ge L h^2 K_0(0) \asymp (\log(dn)/n)^{2/5}$, the proof of (36) is complete by Lemma 14.

### C.2 Proof of $\ell_1$ Norm in (37)

In order to show the lower bound for $\|\cdot\|_1$, we need to construct a different $\mathcal{U}_0$. We still choose $\mathbf{\Omega}_0(\cdot) \equiv \mathbf{I}$. Let $\mathcal{B}$ be the set of matrices defined as

$$\mathcal{B} = \Big\{\mathbf{B} \in \mathbb{R}^{d\times d} \,\Big|\, B = \begin{pmatrix} 0 & \boldsymbol{v} \\ \boldsymbol{v} & 0 \end{pmatrix} \text{ s.t. } \boldsymbol{v} \in \{0,1\}^{d-1}, \|\boldsymbol{v}\|_0 = s, \boldsymbol{v}_2 = 0\Big\}.$$

With this, we define

$$\mathcal{U}_0' = \big\{\mathbf{\Omega}_{(j,m)}(\cdot) \mid \mathbf{\Omega}^{-1}_{(j,m)}(z) = \mathbf{\Sigma}_{(j,m)}(z) = \mathbf{I} + \tau_m(z)B_j, \text{ for } z \in (0,1), B_j \in \mathcal{B}, m \in [M_0]\big\},$$

where the index $j$ corresponds to an element in the set $\mathcal{B}$ and the function $\tau_m(\cdot)$ is defined in (71). Let $D := |\mathcal{B}| = \binom{d-2}{s}$. We still choose $M_0 = \lceil c_0(n/\log(dn))^{1/5}\rceil$ for some constant $c_0$. It can be easily shown that $\mathcal{U}_0 \subset \overline{\mathcal{U}}_s(M, \rho, L)$ if $s^2\log(dn)/n^{4/5} = o(1)$.

The rest of the proof is similar to the proof in Section C.1. For any $j \in [D]$, $m \in [M_0]$,

$$\sup_{z\in(0,1)} \|\mathbf{\Omega}_{(j,m)}(z) - \mathbf{\Omega}_0(z)\|_1 \geq s\|\tau_m\|_{\sup} \geq sLh^2K_0(0),$$

giving $r_{\min} \geq sLh^2K_0(0) \asymp s(\log(dn)/n)^{2/5}$. We proceed to show $\|\mathbb{P}_{\mathbf{\Omega}_0} \wedge \overline{\mathbb{P}}\| \geq 1/2$, where $\overline{\mathbb{P}} = (DM_0)^{-1}\sum_{j\in[D],m\in[M_0]}\mathbb{P}_{\mathbf{\Omega}_{(j,m)}}$, by proving

$$\Delta := \int \Big(\frac{1}{DM_0}\sum_{j,m} f_{jm}\Big)^2 / f_0 \, d\mu - 1 \to 0.$$

We establish the above display by modifying the proof of Lemma 2 in Ren et al. (2015).

Let $J(j_1, j_2) = \mathrm{Vec}(B_{j_1})^T\mathrm{Vec}(B_{j_2})/2$, where $\mathrm{Vec}(\mathbf{M}) = (\mathbf{M}_1^T, \ldots, \mathbf{M}_d^T)^T$ for any matrix $\mathbf{M} = [\mathbf{M}_1, \ldots, \mathbf{M}_d]$. From (73) and the definition of $\mathcal{B}$, we have

$$\begin{aligned}\int \frac{f_{j_1m_1}f_{j_1m_2}}{f_0}d\mu &= \Big[\int_0^1 \Big[\det\Big(\mathbf{I} - \tau_{m_1}(z)\tau_{m_2}(z)B_{j_1}B_{j2}\Big)\Big]^{-1/2}dz\Big]^n \\ &= \Big[\int_0^1 (1 - J(j_1,j_2)\tau_{m_1}(z)\tau_{m_2}(z))^{-1}dz\Big]^n.\end{aligned} \tag{76}$$

Similar to (74), when $m_1 \neq m_2$, (76) yields that $\int f_{j_1m_1}f_{j_2m_2}/f_0 d\mu = 1$. Similar to (75), we also have

$$\int \frac{f_{j_1m}f_{j_2m}}{f_0}d\mu \leq \Big[1 + \frac{2J(j_1,j_2)L^2h^4K_{\max}^2}{M_0}\Big]^n.$$

Since

$$\big|\{(j_1,j_2) \in [D]^2 \mid J(j_1,j_2) = j\}\big| = \binom{d-2}{s}\binom{s}{j}\binom{d-2-s}{s-j},$$

we have

$$\begin{aligned}\Delta &= \frac{1}{D^2M_0^2}\sum_{m=1}^{M_0}\sum_{0\leq j\leq s}\sum_{J(j_1,j_2)=j}\Big(\int\frac{f_{j_1m}f_{j_2m}}{f_0}d\mu - 1\Big) \\ &\leq \frac{1}{D^2M_0^2}\sum_{m=1}^{M_0}\sum_{0\leq j\leq s}\sum_{J(j_1,j_2)=j}\Big(\Big[1 + \frac{2jL^2h^4K_{\max}^2}{M_0}\Big]^n - 1\Big) \\ &\leq \frac{1}{D^2M_0^2}\sum_{m=1}^{M_0}\sum_{0\leq j\leq s}\binom{d-2}{s}\binom{s}{j}\binom{d-2-s}{s-j}\Big[1 + \frac{2jL^2h^4K_{\max}^2}{M_0}\Big]^n.\end{aligned}$$

Let $a_0 = 2c_0^{-5}L^2K_{\max}^2$. Similar to the proof of Lemma 6 in Ren et al. (2015), we have

$$\Big[1 + \frac{2jL^2h^4K_{\max}^2\log(dn)}{M_0}\Big]^n \leq \exp\big(2nM^{-1}jL^2h^4K_{\max}^2\big) \leq (nd)^{a_0j}.$$

This further gives us

$$\Delta \le \frac{1}{M_0^2} \sum_{m=1}^{M_0} \sum_{1 \le j \le s} \binom{s}{j} \binom{d-2-s}{s-j} (nd)^{a_0 j} \Big/ \binom{d-2}{s}$$
$$= \frac{1}{M_0^2} \sum_{m=1}^{M_0} \sum_{0 \le j \le s} \frac{1}{j!} \left( \frac{s!(d-s)!}{(s-j)!} \right)^2 \frac{(nd)^{a_0 j}}{(d-2)!(d-2s+j-2)!}$$
$$\le \frac{1}{M_0} \sum_{1 \le j \le s} \left( \frac{s^2 (nd)^{a_0}}{d-s} \right)^j,$$

where the last inequality is due to $\frac{s!}{(s-j)!} \le s^j$ and $\frac{(d-2)!(d-2s+j-2)!}{[(d-s)!]^2} \ge (d-s)^j$. By assumption $s^{-v} d \ge 1$ for some $v > 2$, and therefore

$$\Delta \le \frac{1}{c_0 (n/\log n)^{1/5}} \sum_{0 \le j \le s} (2d^{2/v-1}(dn)^{a_0})^j \le \frac{2c_0 (n/\log(dn))^{-1/5} d^{2/v-1+a_0} n^{a_0}}{(1-2d^{2/v-1+2a_0})},$$

for sufficiently large $d$, $d-s \ge d/2$, $d \ge n$. By choosing $c_0$ in $a_0 = 2c_0^{-5} L^2 K_{\max}^2$ sufficiently small, so that $a_0 < \min(1/5, 1/v - 1/2)$, we have $\Delta \to 0$. This completes the proof.

## Appendix D. Convergence Rate of Kendall's Tau Estimator

In this section, we study the rate of convergence of kernel Kendall's tau estimator $\widehat{\tau}_{jk}(z)$ in (4) to $\tau_{jk}(z)$ uniformly in the bandwidth $h \in [h_l, h_u]$, the index $z \in (0,1)$ and the dimension $j,k \in [d]$. We start by establishing the bias-variance decomposition for $\widehat{\tau}_{jk}(z)$ and then show how to bound the bias and variance separately.

Recall that in (40), we define

$$g_{z|(j,k)}(y_i, y_{i'}) = \omega_z(z_i, z_{i'}) \operatorname{sign}(x_{ia} - x_{i'a}) \operatorname{sign}(x_{ib} - x_{i'b}),$$

and $\omega_z(z_i, z_{i'})$ is defined in (5). The estimator $\widehat{\tau}_{jk}(z)$ can be written as a quotient of two $U$-statistics

$$\widehat{\tau}_{jk}(z) = \mathbb{U}_n[g_{z|(j,k)}]/\mathbb{U}_n[\omega_z].$$

Recall that the operator $u_n[\cdot]$ is defined as

$$u_n[H] = \sqrt{n} \cdot \big( \mathbb{U}_n[H] - \mathbb{E}[\mathbb{U}_n[H]] \big)$$

for any bivariate function $H(x, x')$. With this, following an argument similar to that in Equation (3.45) of Pagan and Ullah (1999), we have the following decomposition

$$\begin{aligned}
\widehat{\tau}_{jk}(z) &= \frac{\mathbb{E}[\mathbb{U}_n\left[g_{z|(j,k)}\right]] + \mathbb{U}_n\left[g_{z|(j,k)}\right] - \mathbb{E}[\mathbb{U}_n\left[g_{z|(j,k)}\right]]}{\mathbb{E}[\mathbb{U}_n[\omega_z]]} \left[ 1 + \frac{\mathbb{U}_n[\omega_z] - \mathbb{E}[\mathbb{U}_n[\omega_z]]}{\mathbb{E}[\mathbb{U}_n[\omega_z]]} \right]^{-1} \\
&= \frac{\mathbb{E}\left[\mathbb{U}_n\left[g_{z|(j,k)}\right]\right]}{\mathbb{E}[\mathbb{U}_n[\omega_z]]} + \frac{u_n\left[g_{z|(j,k)}\right]}{\sqrt{n}\,\mathbb{E}[\mathbb{U}_n[\omega_z]]} - \frac{\mathbb{E}\left[\mathbb{U}_n\left[g_{z|(j,k)}\right]\right] \cdot u_n\left[\omega_z\right]}{\sqrt{n}\left(\mathbb{E}[\mathbb{U}_n[\omega_z]]\right)^2} \\
&\quad + n^{-1} O\left( u_n\left[g_{z|(j,k)}\right] \cdot u_n\left[\omega_z\right] + \left(u_n\left[\omega_z\right]\right)^2 \right),
\end{aligned} \tag{77}$$

under the condition that

$$\left|\frac{u_n[\omega_z]}{\sqrt{n}\mathbb{E}[\mathbb{U}_n[\omega_z]]}\right| < 1 \quad \text{and} \quad \mathbb{E}[\mathbb{U}_n[\omega_z]] \neq 0. \tag{78}$$

Taking expectation on both sides of (77), we obtain

$$\mathbb{E}\left[\widehat{\tau}_{jk}(z)\right] = \frac{\mathbb{E}\left[\mathbb{U}_n\left[g_{z|(j,k)}\right]\right]}{\mathbb{E}\left[\mathbb{U}_n[\omega_z]\right]} + n^{-1}O\Bigg(\mathbb{E}\Big[u_n\left[g_{z|(j,k)}\right]\cdot u_n\left[\omega_z\right]\Big] + \mathbb{E}\left[\left(u_n\left[\omega_z\right]\right)^2\right]\Bigg). \tag{79}$$

With this, we are ready to prove Lemma 9 and Lemma 10.

### D.1 Proof of Lemma 9

We first check that the condition in (78) is satisfied under the assumptions. Using (50) and (52), for sufficiently large $n$,

$$\sup_{j,k\in[d]}\sup_{z\in(0,1)}\left|\frac{u_n[\omega_z]}{\sqrt{n}\mathbb{E}[\mathbb{U}_n[\omega_z]]}\right| \lesssim \sqrt{\frac{\log(d/h)\vee\log\left(\delta^{-1}\log\left(h_u h_l^{-1}\right)\right)}{nh}}\cdot\frac{1}{f_Z^2(z)+O(h^2)} < 1.$$

Since $f_Z$ is bounded from below, $\mathbb{E}[\mathbb{U}_n[\omega_z]] = f_Z^2(z) + O(h^2) \geq \underline{f}_Z/2 > 0$ for large enough $n$. Therefore, condition in (78) holds and the expansion in (77) is valid.

From (77) and (79), we have

$$\widehat{\tau}_{jk}(z,h) - \mathbb{E}\left[\widehat{\tau}_{jk}(z,h)\right] = \underbrace{\frac{u_n\left[g_{z|(j,k)}\right]}{\sqrt{n}\,\mathbb{E}[\mathbb{U}_n[\omega_z]]}}_{I_1} - \underbrace{\frac{\mathbb{E}\left[\mathbb{U}_n\left[g_{z|(j,k)}\right]\right]\cdot u_n\left[\omega_z\right]}{\sqrt{n}\left(\mathbb{E}[\mathbb{U}_n[\omega_z]]\right)^2}}_{I_2} + I_3, \tag{80}$$

where

$$I_3 = n^{-1}O\left(u_n\left[g_{z|(j,k)}\right]\cdot u_n\left[\omega_z\right] + \left(u_n\left[\omega_z\right]\right)^2 + \mathbb{E}\Big[u_n\left[g_{z|(j,k)}\right]\cdot u_n\left[\omega_z\right]\Big] + \mathbb{E}\left[\left(u_n\left[\omega_z\right]\right)^2\right]\right).$$

We bound $I_1$, $I_2$ and $I_3$ separately. Using (50) and (52), we have that

$$\begin{aligned}
&\sup_{j,k\in[d]}\sup_{h\in[h_l,h_u]}\sup_{z\in(0,1)}\left|\frac{\sqrt{nh}\cdot|I_1|}{\sqrt{\log(d/h)\vee\log\left(\delta^{-1}\log\left(h_u h_l^{-1}\right)\right)}}\right| \\
&\leq \sup_{j,k\in[d]}\sup_{h\in[h_l,h_u]}\sup_{z\in(0,1)}\left|\frac{\sqrt{h}}{\sqrt{\log(d/h)\vee\log\left(\delta^{-1}\log\left(h_u h_l^{-1}\right)\right)}}\frac{|u_n\left[g_{z|(j,k)}\right]|}{f_Z^2(z)+O(h^2)}\right| \leq C,
\end{aligned}$$

with probability $1-\delta$ for large enough $n$. Similarly, using (50), (51) and (52), we also have

$$\begin{aligned}
&\sup_{j,k\in[d]}\sup_{h\in[h_l,h_u]}\sup_{z\in(0,1)}\left|\frac{\sqrt{nh}\cdot|I_2|}{\sqrt{\log(d/h)\vee\log\left(\delta^{-1}\log\left(h_u h_l^{-1}\right)\right)}}\right| \\
&\leq \sup_{j,k\in[d]}\sup_{h\in[h_l,h_u]}\sup_{z\in(0,1)}\left|\frac{\sqrt{h}}{\sqrt{\log(d/h)\vee\log\left(\delta^{-1}\log\left(h_u h_l^{-1}\right)\right)}}\frac{|u_n\left[\omega_z\right]|(f_Z^2(z)\tau_{jk}(z)+O(h^2))}{(f_Z^2(z)+O(h^2))^2}\right| \leq C,
\end{aligned}$$

with probability $1-\delta$. Finally, using (50), (53) and (54), we have

$$\sup_{j,k\in[d]}\sup_{h\in[h_l,h_u]}\sup_{z\in(0,1)}\left|\frac{nh\cdot|I_3|}{\log(d/h)\vee\log\left(\delta^{-1}\log\left(h_uh_l^{-1}\right)\right)}\right|$$
$$\lesssim \sup_{j,k\in[d]}\sup_{h\in[h_l,h_u]}\sup_{z\in(0,1)}\left|\frac{2h}{\log(d/h)\vee\log\left(\delta^{-1}\log\left(h_uh_l^{-1}\right)\right)}\Big(u_n\left[\omega_z\right]\vee u_n\left[g_{z|(j,k)}\right]\Big)^2\right|$$
$$+\sup_{j,k\in[d]}\sup_{h\in[h_l,h_u]}\sup_{z\in(0,1)}\left|\frac{nh\cdot O((nh)^{-1})}{\log(d/h)\vee\log\left(\delta^{-1}\log\left(h_uh_l^{-1}\right)\right)}\right|\le C,$$

with probability $1-\delta$.

Combining the above three displays with (80) completes the proof.

### D.2 Proof of Lemma 10

It follows from Lemma 13 and the decomposition in (79) that

$$\sup_{j,k\in[d]}\sup_{h\in[h_l,h_u]}\sup_{z\in(0,1)}\frac{|\mathbb{E}[\widehat{\tau}_{jk}(z)]-\tau_{jk}(z)|}{h^2+1/(nh)}$$
$$\le \sup_{j,k\in[d]}\sup_{h\in[h_l,h_u]}\sup_{z\in(0,1)}\left[\frac{f_Z^2(z)\tau_{jk}(z)+O(h^2)}{f_Z^2(z)+O(h^2)}-\tau_{jk}(z)+O((nh)^{-1})\right]\frac{1}{h^2+(nh)^{-1}}$$
$$\le \sup_{j,k\in[d]}\sup_{h\in[h_l,h_u]}\sup_{z\in(0,1)}\left[\frac{O(h^2+(nh)^{-1})}{f_Z^2(z)+O(h^2)}\right]\frac{1}{h^2+(nh)^{-1}}\le C.$$

## Appendix E. Concentration of $U$-statistics

In this section, we study certain properties of the $U$-statistics used in this paper. We state and prove Lemma 12 and Lemma 13, which were used to establish results on the rate of convergence of Kendall's tau estimator in Appendix D. In particular, we will prove the following two results in this section, with the notations on $U$-statistics and empirical processes defined in Appendix D.

### E.1 Proof of Lemma 12

At a high-level, we will use the Hoeffding decomposition to represent the $U$-statistics $\mathbb{U}_n[\omega_z]$ and $\mathbb{U}_n[g_{z|(j,k)}]$ defined in (43) and (46). Next, we will use concentration inequalities for suprema of empirical processes and $U$-statistics to bound individual terms in the decomposition.

Recall that (43) and (46) give

$$n^{-1/2}\cdot u_n\left[g_{z|(j,k)}\right]=\mathbb{U}_n\left[g_{z|(j,k)}\right]-\mathbb{E}\left[\mathbb{U}_n\left[g_{z|(j,k)}\right]\right]=2\mathbb{E}_n\left[g^{(1)}_{z|(j,k)}(Y_i)\right]+\mathbb{U}_n\left[g^{(2)}_{z|(j,k)}\right]$$

and

$$n^{-1/2}\cdot u_n\left[\omega_z\right]=\mathbb{U}_n[\omega_z]-\mathbb{E}\left[\mathbb{U}_n[\omega_z]\right]=2\mathbb{E}_n\left[\omega_z^{(1)}\right]+\mathbb{U}_n\left[\omega_z^{(2)}\right].$$

In order to bound $u_n[\omega_z]$ and $u_n\left[g_{z|(j,k)}\right]$ it is sufficient to bound $g^{(1)}_{z|(j,k)}$, $g^{(2)}_{z|(j,k)}$, $\omega^{(1)}_z$ and $\omega^{(2)}_z$. We do so in the following lemmas.

**Lemma 15** *We assume $n^{-1}\log d = o(1)$ and the bandwidths $0 < h_l < h_u < 1$ satisfy $h_l n/\log(dn) \to \infty$ and $h_u = o(1)$. There exist a universal constant $C > 0$ such that*

$$\sup_{j,k\in[d]}\sup_{h\in[h_l,h_u]}\sup_{z\in(0,1)}\left|\mathbb{G}_n\left[\frac{\sqrt{h}g^{(1)}_{z|(j,k)}}{\sqrt{\log(d/h)\vee\log\left(\delta^{-1}\log\left(h_u h_l^{-1}\right)\right)}}\right]\right| \le C \tag{81}$$

*with probability $1-\delta$.*

**Proof** The general strategy to show (81) can be separated into two steps.

**Step 1.** Splitting the supreme over $h \in [h_l, h_u]$ into smaller intervals such that the empirical process (81) is easier to bound for each interval. Let $S$ be the smallest integer such that $2^S h_l \ge h_u$. Note that $S \le \log_2(h_u/h_l)$ and $[h_l, h_u] \subseteq \cup_{\ell=1}^S [2^{\ell-1}h_l, 2^\ell h_l] =: \mathcal{H}_\ell$. Then

$$\mathbb{P}\left[\sup_{z,j,k}\sup_{h\in[h_l,h_u]}\left|\mathbb{G}_n\left[\sqrt{h}g^{(1)}_{z|(j,k)}\right]\right| \ge t\right] \le \sum_{\ell=1}^S \mathbb{P}\left[\sup_{j,k\in[d]}\sup_{h\in\mathcal{H}_\ell}\sup_{z\in(0,1)}\left|\mathbb{G}_n\left[\sqrt{h}g^{(1)}_{z|(j,k)}\right]\right| \ge t\right]. \tag{82}$$

**Step 2.** We apply Talagrand's inequality (Bousquet, 2002) to each term in the summation on the right hand side of (82).

Consider the class of functions

$$\mathcal{F}_\ell = \left\{\sqrt{h}g^{(1)}_{z|(j,k)} \,\middle|\, h \in \mathcal{H}_\ell, z \in (0,1), j,k \in [d]\right\}.$$

In order to apply Talagrand's inequality to functions in the class $\mathcal{F}_\ell$, we have to check three conditions:

- The class $\mathcal{F}_\ell$ is uniformly bounded;
- Bounding the variance of $g^{(1)}_{z|(j,k)}$;
- Bounding the covering number of $\mathcal{F}_\ell$

We verify the above three conditions next. First, we show that the class $\mathcal{F}_\ell$ is uniformly bounded.

Recall that in (62), we show that

$$\mathbb{E}\left[g_{z|(j,k)}(y',Y)\right] = K_h(z'-z)\int K_h(s-z)\varphi\left(x'_j, x'_k, \mathbf{\Sigma}_{jk}(s)\right)f_Z(s)ds, \tag{83}$$

where $\varphi(\cdot)$ is defined in (61). Since $|\varphi(u,v,\rho)| \le 2$ for any $u$, $v$, and $\rho$, and $h \in \mathcal{H}_\ell$, the above display gives us

$$\begin{aligned}\sup_{z\in(0,1)}\max_{j,k\in[d]}\left\|g^{(1)}_{z|(j,k)}\right\|_\infty &\le \frac{2\|K\|_\infty}{h}\sup_{z\in(0,1)}\int K_h(s-z)f_Z(s)ds + \sup_{z\in(0,1)}\max_{j,k\in[d]}\mathbb{E}\left[\mathbb{U}_n\left[g_{z|(j,k)}\right]\right]\\ &\le \frac{2\|K\|_\infty}{h}\sup_{z\in(0,1)}\left(f_Z(z)+O(h^2)\right) + \bar{\mathrm{f}}_Z^2 + O(h^2) \le \frac{3\|K\|_\infty\bar{\mathrm{f}}_Z}{h},\end{aligned} \tag{84}$$

where the second inequality is due to (51). This results shows that the class $\mathcal{F}_\ell$ is uniformly bounded by $(2^\ell h_l)^{-1/2} 3\|K\|_\infty \bar{\mathrm{f}}_Z$ and has the envelope $F_\ell = 3(2^{\ell-1}h_l)^{-1/2}\|K\|_\infty \bar{\mathrm{f}}_Z$.

Next, we bound the variance of $g^{(1)}_{z|(j,k)}$. Let $\bar{\varphi}_{u,v}(s) = \varphi(u, v, \mathbf{\Sigma}_{jk}(s)) f_Z(s)$. We can rewrite (83) as

$$\mathbb{E}\left[g_{z|(j,k)}(y', Y)\right] = K_h(z' - z) \cdot (K_h * \bar{\varphi}_{u,v})(z),$$

where "$*$" denotes the convolution operator. Then we bound the convolution by

$$\sup_{u,v} \|K_h * \bar{\varphi}_{u,v}\|_\infty \le \sup_{u,v} \|K_h\|_1 \|\bar{\varphi}_{u,v}\|_\infty \le 2\bar{\mathrm{f}}_Z, \tag{85}$$

where the first inequality follows using the Young's inequality for convolution. Now, for any fixed $h \in \mathcal{H}_\ell$ and $z \in (0, 1)$, we have

$$\begin{aligned}
&\sup_{z,j,k} \mathbb{E}\left[\left(\sqrt{h} g^{(1)}_{z|(j,k)}\right)^2\right] \le \sup_{z,j,k} \mathbb{E}\left[\max_{u,v}\left(\sqrt{h} g^{(1)}_{z|(u,v)}\right)^2\right] \\
&\le \sup_z 2\mathbb{E}\left[h\left(K_h(Z - z)\right)^2\right] \sup_{u,v} \|K_h * \bar{\varphi}_{u,v}\|^2_\infty + \sup_{z,j,k} 2h\left\{\mathbb{E}\left[\mathbb{U}_n\left[g_{z|(j,k)}\right]\right]\right\}^2 \\
&\le C\bar{\mathrm{f}}^2_Z \cdot \sup_z \left(\int \frac{1}{h} K^2\left(\frac{x - z}{h}\right) f_Z(x) dx\right) + Ch\bar{\mathrm{f}}^2_Z + O(h^3) \\
&\le C\bar{\mathrm{f}}^2_Z \sup_z (f_Z(z)\|K\|^2_2 + O(h^2)) \le C\bar{\mathrm{f}}^3_Z ||K||^2_2.
\end{aligned} \tag{86}$$

The reason why we show a stronger result above on the expectation of maximal is because we need (86) in the proof of Section F.

Using the above display, we can bound the variance for any $\ell \in [S]$,

$$\sigma^2_\ell := \sup_{f \in \mathcal{F}_\ell} \mathbb{E}\left[f^2\right] \le C\bar{\mathrm{f}}^3_Z ||K||^2_2. \tag{87}$$

Finally, we need a bound on the covering number of $\mathcal{F}_\ell$. Using Lemma 28 we have that

$$\sup_Q N(\mathcal{F}_\ell, L_2(Q), \epsilon) \le \frac{Cd^2}{(2^{\ell-1}h_\ell)^{v+9}\epsilon^{v+6}}.$$

Theorem 3.12 of Koltchinskii (2011) then gives us that for some universal constant $C$,

$$\mathbb{E}\left[\sup_{f \in \mathcal{F}_\ell} |n\mathbb{E}_n[f]|\right] \le C\sqrt{n \log\left(\frac{Cd}{2^{\ell-1}h_\ell}\right)} + \frac{C}{(2^{\ell-1}h_l)^{1/2}} \log\left(\frac{Cd}{2^{\ell-1}h_\ell}\right).$$

Furthermore, since $h_l n / \log(dn) \to \infty$, the above display simplifies to

$$\mathbb{E}\left[\sup_{f \in \mathcal{F}_\ell} |n\mathbb{E}_n[f]|\right] \le C\sqrt{n \log\left(\frac{Cd}{2^{\ell-1}h_l}\right)}. \tag{88}$$

Using Theorem 2.3 of Bousquet (2002) together with (87), (88), and $n^{-1}\log\left(dh^{-1}\right) = o(1)$, we obtain

$$\sup_{j,k \in [d]} \sup_{h \in \mathcal{H}_\ell} \sup_{z \in (0,1)} \left|\mathbb{G}_n\left[\sqrt{h} g^{(1)}_{z|(j,k)}\right]\right| \le C\left(\sqrt{\log\left(\frac{Cd}{2^{\ell-1}h_l}\right)} + \sqrt{\log(1/\delta)}\right) \tag{89}$$

with probability $1-\delta$. Observe that $2^{\ell-1}h_l \geq h/2$ for any $h \in \mathcal{H}_\ell$. Therefore, combining (89), (82), we obtain that for some constant $C > 0$

$$\sup_{j,k\in[d]} \sup_{h\in[h_l,h_u]} \sup_{z\in(0,1)} \left| \mathbb{G}_n \left[ \frac{\sqrt{h} g^{(1)}_{z|(j,k)}}{\sqrt{\log(d/h) \vee \log\left(\delta^{-1}\log\left(h_u h_l^{-1}\right)\right)}} \right] \right| \leq C,$$

with probability $1-\delta$. ■

**Lemma 16** *We assume* $n^{-1}\log d = o(1)$ *and the bandwidths* $0 < h_l < h_u < 1$ *satisfy* $h_l n/\log(dn) \to \infty$ *and* $h_u = o(1)$. *There exists a universal constant* $C > 0$ *such that*

$$\sup_{j,k\in[d]} \sup_{h\in[h_l,h_u]} \sup_{z\in(0,1)} \left| \frac{nh}{\log(d/h) \vee \log\left(\delta^{-1}\log\left(h_u h_l^{-1}\right)\right)} \mathbb{U}_n \left[ g^{(2)}_{z|(j,k)} \right] \right| \leq C$$

*with probability* $1-\delta$.

**Proof** The method to prove this lemma is similar to the proof of Lemma 15. The only difference is that instead of bounding the empirical process, we bound a suprema of a $U$-statistic process. Therefore, we will use Theorem 33 instead of Talagrand's inequality.

We apply the trick of splitting $[h_l, h_u]$ again. Let $S$ be the smallest integer such that $2^S h_l \geq h_u$. We have that $S \leq \log_2(h_u/h_l)$ and $[h_l, h_u] \subseteq \cup_{\ell=1}^S [2^{\ell-1}h_l, 2^\ell h_l]$. For simplicity, we define $\mathcal{H}_\ell = [2^{\ell-1}h_l, 2^\ell h_l]$. Therefore,

$$\mathbb{P}\left[ \sup_{z,j,k} \sup_{h\in[h_l,h_u]} \left| h\mathbb{U}_n \left[ g^{(2)}_{z|(j,k)} \right] \right| \geq t \right] \leq \sum_{j=1}^S \mathbb{P}\left[ \sup_{z,j,k} \sup_{h\in\mathcal{H}_\ell} \left| h\mathbb{U}_n \left[ g^{(2)}_{z|(j,k)} \right] \right| \geq t \right]. \tag{90}$$

As $g^{(2)}_{z|(j,k)}$ is a degenerate kernel, we can use Theorem 33 to bound the right hand side of (90). Consider the class of functions

$$\mathcal{F}^{(2)}_{\ell|(j,k)} = \left\{ (2^{\ell-1}h_l L) h g^{(2)}_{z|(j,k)} \,\middle|\, h \in \mathcal{H}_\ell, z \in (0,1) \right\} \text{ and } \mathcal{F}^{(2)}_\ell = \left\{ f \in \mathcal{F}^{(2)}_{\ell|(j,k)} \mid j,k \in [d] \right\}.$$

where $L^{-1} := 7\|K\|_\infty^2 \bar{\mathrm{f}}_Z^2$. From the expansion in (42), with (84) and (51), we have

$$\sup_{z\in(0,1)} \max_{j,k\in[d]} \left\| g^{(2)}_{z|(j,k)} \right\|_\infty \leq C\|K\|_\infty^2 \bar{\mathrm{f}}_Z^2 h^{-2}, \qquad \text{for } h \in [h_l, h_u].$$

Therefore, the class $\mathcal{F}^{(2)}_\ell$ is uniformly bounded by 1 and has the envelope $F^{(2)}_\ell = 1$, which verifies the condition in (159). Furthermore, by (86) and (51), we can bound the variance

as

$$\begin{aligned}
&\sup_{z,j,k} \mathbb{E}\left[\left(hg^{(2)}_{z|(j,k)}\right)^2\right] \\
&\quad \lesssim \sup_{z,j,k}\left\{4h^2\mathbb{E}\left[\left(g_{z|(j,k)}(Y_i,Y_{i'})\right)^2\right] + h\mathbb{E}\left[\left(\sqrt{h}g^{(1)}_{z|(j,k)}\right)^2\right] + 4\{h\mathbb{E}[\mathbb{U}_n[g_{z|(j,k)}]]\}^2\right\} \\
&\quad \lesssim \sup_{z} h^{-2}\mathbb{E}\left[K\left((Z_i-z)/h\right)^2 K\left((Z_{i'}-z)/h\right)^2\right] + \bar{\mathrm{f}}_Z^3 h + \bar{\mathrm{f}}_Z^2 h^2 \\
&\quad \lesssim \sup_{z}\left(h^{-1}\int K\left((x-z)/h\right)^2 f_Z(x)dx\right)^2 + \bar{\mathrm{f}}_Z^3 h + \bar{\mathrm{f}}_Z^2 h^2 \\
&\quad \lesssim \bar{\mathrm{f}}_Z^2 ||K||_2^4 + \bar{\mathrm{f}}_Z^3 h + \bar{\mathrm{f}}_Z^2 h^2 \lesssim \bar{\mathrm{f}}_Z^2 ||K||_2^4.
\end{aligned}$$

Therefore, for any $\ell \in [S]$,

$$\sup_{f\in\mathcal{F}^{(2)}_\ell} \mathbb{E}\left[f^2\right] \lesssim 2\bar{\mathrm{f}}_Z^2 ||K||_2^4 (2^{\ell-1}h_l L)^2 := \sigma_\ell^2 < 1.$$

According to Lemma 28, we have that

$$\sup_Q N(\mathcal{F}^{(2)}_{\ell|(j,k)}, L_2(Q), \epsilon) \le C\frac{(2^{\ell-1}h_l L)^{-4v-15}}{(2^{\ell-1}h_l)^{2v+18}\epsilon^{4v+15}}.$$

By setting $t = C(2^{\ell-1}h_l)[\log\left(C/(2^{\ell-1}h_l)\right) \vee \log(C/\delta)]$ in (160) for a sufficiently large constant $C$, as $h_l n/\log(dn) \to \infty$, we have

$$\begin{aligned}
n\sigma_\ell^2 = n(2^{\ell-1}h_l)^2 &\ge \log\left(1/(2^{\ell-1}h_l)\right) \vee \log(1/\delta) \\
&= \frac{t}{\sigma_\ell} \ge C'\left(\frac{\log(1/(2^{\ell-1}h_l))}{\log n}\right)^{3/2}\log\left(\frac{2}{2^{\ell-1}h_l}\right),
\end{aligned}$$

for some constant $C' > 0$. This verifies the condition (161). Now (160) gives us that with probability $1 - [(2^{\ell-1}h_\ell) \vee \delta]$,

$$\sup_{h\in\mathcal{H}_\ell}\sup_{z\in(0,1)} (2^{\ell-1}h_l L)h\mathbb{U}_n\left[g^{(2)}_{z|(j,k)}\right] \le Cn^{-1}(2^{\ell-1}h_l)[\log\left(1/(2^{\ell-1}h_l)\right) \vee \log(1/\delta)]. \tag{91}$$

Since $2^{\ell-1}h_l \ge h/2$ for any $h \in \mathcal{H}_\ell$, applying (90) and (91) with the union bound over $j,k \in [d]$, we obtain

$$\sup_{j,k\in[d]}\sup_{h\in[h_l,h_u]}\sup_{z\in(0,1)}\left|\frac{nh}{\log(d/h) \vee \log\left(\delta^{-1}\log\left(h_u h_l^{-1}\right)\right)}\mathbb{U}_n\left[g^{(2)}_{z|(j,k)}\right]\right| \le C, \tag{92}$$

with probability $1-(h_u \vee \delta)$. As $h_u = o(1)$, (92) follows with probability larger than $1-\delta$. ■

**Lemma 17** *We suppose* $n^{-1}\log d = o(1)$ *and the bandwidths* $0 < h_l < h_u < 1$ *satisfy* $h_l n/\log(dn) \to \infty$ *and* $h_u = o(1)$*. There exists a universal constant* $C > 0$ *such that*

$$\sup_{h\in[h_l,h_u]}\sup_{z\in(0,1)}\left|\mathbb{G}_n\left[\frac{\sqrt{h}\omega_z^{(1)}}{\sqrt{\log(1/h)\vee\log\left(\delta^{-1}\log\left(h_u h_l^{-1}\right)\right)}}\right]\right| \le C$$

*with probability* $1-\delta$*.*

**Proof** The proof is similar to that of Lemma 15. We again decompose $[h_l, h_u] \subseteq \cup_{\ell=1}^{S}\mathcal{H}_\ell$, where $\mathcal{H}_\ell = [2^{\ell-1}h_l, 2^\ell h_l]$ and $S$ is the smallest integer such that $2^S h_l \ge h_u$. By the union bound,

$$\mathbb{P}\left[\sup_{h\in[h_l,h_u]}\sup_{z\in(0,1)}\left|\mathbb{G}_n\left[h^2\omega_z^{(1)}\right]\right| \ge t\right] \le \sum_{j=1}^{S}\mathbb{P}\left[\sup_{h\in\mathcal{H}_\ell}\sup_{z\in(0,1)}\left|\mathbb{G}_n\left[h^2\omega_z^{(1)}\right]\right| \ge t\right]. \qquad (93)$$

Talagrand's inequality (Bousquet, 2002) is applied once again to control (93). Consider the class of functions

$$\mathcal{K}_\ell = \left\{\sqrt{h}\omega_z^{(1)} \mid h \in \mathcal{H}_\ell, z \in (0,1)\right\}.$$

According to the definition of $\omega_z^{(1)}$ in (44), we have

$$\omega_z^{(1)}(s) = K_h(s-z)\mathbb{E}[K_h(z-Z)] - (\mathbb{E}[K_h(z-Z)])^2.$$

We can bound the supremum norm by

$$\sup_{z\in(0,1)}\|\omega_z^{(1)}\|_\infty \le 2\bar{\mathrm{f}}_Z\|K\|_\infty h^{-1}, \qquad (94)$$

which implies that the envelope function is $F_\ell = 2\bar{\mathrm{f}}_Z\|K\|_\infty(2^{\ell-1}h_\ell)^{-1/2}$. Similar to (86), we can also bound the variance by

$$\begin{aligned}\sup_z \mathbb{E}\left[\left(\sqrt{h}\omega_z^{(1)}\right)^2\right] &\le \sup_z 2\mathbb{E}\left[h\left(K_h(Z-z)\right)^2\right]\cdot(\mathbb{E}[K_h(z-Z)])^2 + \sup_z 2h\left\{\mathbb{E}\left[\mathbb{U}_n\left[\omega_z\right]\right]\right\}^2\\ &\le 2(\bar{\mathrm{f}}_Z^3\|K\|_2^2 + O(h^2)) + 2h\bar{\mathrm{f}}_Z^2 + O(h^3) \le 2\bar{\mathrm{f}}_Z^3||K||_2^2 := \sigma_\ell^2.\end{aligned} \qquad (95)$$

Using Lemma 29 we have

$$\sup_Q N(\mathcal{K}_\ell, L_2(Q), \epsilon) \le \frac{C}{(2^{\ell-1}h_\ell)^4\epsilon^{2v+1}}\cdot\frac{\|F_\ell\|_{L_2}}{2\|K\|_\infty(2^{\ell-1}h_\ell)^{-1/2}},$$

which combined with Theorem 3.12 of Koltchinskii (2011) with $h_l n/\log(dn) \to \infty$ implies that

$$\mathbb{E}\left[\sup_{f\in\mathcal{K}_\ell}|n\mathbb{E}_n[f]|\right] \le C\sqrt{n\log\left(\frac{C}{2^{\ell-1}h_l}\right)}.$$

Theorem 2.3 of Bousquet (2002) derives that for some constant $C > 0$

$$\sup_{h\in\mathcal{H}_\ell}\sup_{z\in(0,1)}\left|\mathbb{G}_n\left[\sqrt{h}\omega_z^{(1)}\right]\right| \le C\left(\sqrt{\log\left(\frac{C}{2^{\ell-1}h_l}\right)} + \sqrt{\log(1/\delta)}\right) \qquad (96)$$

with probability $1-\delta$. Using the union bound in (96) with (93), we have

$$\sup_{h\in[h_l,h_u]}\sup_{z\in(0,1)}\left|\mathbb{G}_n\left[\frac{\sqrt{h}\omega_z^{(1)}}{\sqrt{\log(1/h)\vee\log\left(\delta^{-1}\log\left(h_u h_l^{-1}\right)\right)}}\right]\right|\le C,$$

for some constant $C>0$ with probability $1-\delta$. ■

**Lemma 18** *We assume $n^{-1}\log d=o(1)$ and the bandwidths $0<h_l<h_u<1$ satisfy $h_l n/\log(dn)\to\infty$ and $h_u=o(1)$. There exists a universal constant $C>0$ such that for sufficiently large $n$,*

$$\sup_{h\in[h_l,h_u]}\sup_{z\in(0,1)}\left|\frac{nh}{\log(1/h)\vee\log\left(\delta^{-1}\log\left(h_u h_l^{-1}\right)\right)}\mathbb{U}_n\left[\omega_z^{(2)}\right]\right|\le C$$

*with probability $1-\delta$.*

**Proof** The proof is similar to that of Lemma 16. Instead of applying Lemma 29, we use Lemma 28. We omit the details of the proof. ■

Let

$$r(d,n,h,\delta,h_u,h_l)=\frac{\log(d/h)\vee\log\left(\delta^{-1}\log\left(h_u h_l^{-1}\right)\right)}{nh}.$$

Applying Lemma 15, Lemma 16, Lemma 17 and Lemma 18 to (43) and (46), we obtain that with probability $1-\delta$, there exists a constant $C>0$ such that

$$\sup_{j,k\in[d]}\sup_{h\in[h_l,h_u]}\sup_{z\in(0,1)}\left|\frac{n^{-1/2}\cdot u_n\left[\omega_z\right]\vee u_n\left[g_{z|(j,k)}\right]}{r^{1/2}(d,n,h,\delta,h_u,h_l)+r(d,n,h,\delta,h_u,h_l)}\right|\le C/2. \tag{97}$$

Since $\log(dn)/(nh_l)=o(1)$ and $h_u=o(1)$, we have $r(d,n,h,\delta,h_u,h_l)\le r^{1/2}(d,n,h,\delta,h_u,h_l)$ for sufficiently large $n$. Using this in (97), we have Lemma 12 proved.

### E.2 Proof of Lemma 13

The high-level idea for proving this lemma is to write the expectations as the integrals and apply Taylor expansions to the density functions and other nonparametric functions. Afterwards, we bound the remainder terms of the Taylor expansions.

We compute $\mathbb{E}\left[\mathbb{U}_n\left[g_{z|(j,k)}\right]\right]$ first. Using Corollary 32, we have

$$\mathbb{E}\left[\mathbb{U}_n\left[g_{z|(j,k)}\right]\right]=\frac{2}{\pi}\mathbb{E}\left[K_h(Z_1-z)K_h(Z_2-z)\arcsin\left(\frac{\boldsymbol{\Sigma}_{jk}(Z_1)+\boldsymbol{\Sigma}_{jk}(Z_2)}{2}\right)\right], \tag{98}$$

where $Z_1,Z_2$ are independent and equal to $Z$ in distribution. After a change of variables, we can further expand the right hand side of (98) as

$$\frac{2}{\pi}\iint K(t_1)K(t_2)f_Z(z+t_1h)f_Z(z+t_2h)\arcsin\left(\frac{\boldsymbol{\Sigma}_{jk}(z+t_1h)+\boldsymbol{\Sigma}_{jk}(z+t_2h)}{2}\right)dt_1dt_2.$$

Using the Taylor series expansion of arcsin$(\cdot)$, we have

$$
\begin{aligned}
&\arcsin\left(\frac{\mathbf{\Sigma}_{jk}(z+t_1h)+\mathbf{\Sigma}_{jk}(z+t_2h)}{2}\right)\\
&=\arcsin\left(\mathbf{\Sigma}_{jk}(z)\right)+\left(\frac{t_1h\dot{\mathbf{\Sigma}}_{jk}(z)+t_2h\dot{\mathbf{\Sigma}}_{jk}(z)}{2}+\frac{(t_1h)^2\ddot{\mathbf{\Sigma}}_{jk}(\bar{z})+(t_2h)^2\ddot{\mathbf{\Sigma}}_{jk}(\bar{z})}{4}\right)\frac{1}{\sqrt{1-\bar{\rho}^2}},
\end{aligned}
$$

where $\bar{z}$ is between $z$ and $z+t_1h$, and $\bar{\rho}$ is between $\mathbf{\Sigma}_{jk}(z)$ and $(\mathbf{\Sigma}_{jk}(z+t_1h)+\mathbf{\Sigma}_{jk}(z+t_2h))/2$. Since the minimum eigenvalue of $\mathbf{\Sigma}(z)$ is strictly positive for any $z$, there exists a $\gamma_\sigma<1$ such that $\sup_z|\mathbf{\Sigma}_{jk}(z)|\le\gamma_\sigma<1$ for any $j\neq k$. We thus have $(1-\bar{\rho}^2)^{-1/2}\le(1-\gamma_\sigma^2)^{-1/2}<\infty$. Similarly, we can expand $f_Z(z+th)=f_Z(z)+th\dot{f}_Z(z)+(th)^2\ddot{f}_Z(\widetilde{z})/2$, where $\widetilde{z}\in(z,z+th)$. Therefore, using regularity conditions on $f_Z$ and $\mathbf{\Sigma}(\cdot)$, we obtain

$$
\max_{j,k\in[d]}\sup_{z\in(0,1)}\left|\mathbb{E}\left[\mathbb{U}_n\left[g_{z|(j,k)}\right]\right]-f_Z^2(z)\tau_{jk}(z)\right|\le\frac{M_\sigma\bar{f}_Z}{\sqrt{1-\gamma_\sigma^2}}h^2 \tag{99}
$$

as desired. Proof of (52) follows in the same way since

$$
\begin{aligned}
\mathbb{E}\left[\mathbb{U}_n\left[K_h(Z_i-z)K_h(Z_{i'}-z)\right]\right]&=\iint K(t_1)K(t_2)f_Z(z+t_1h)f_Z(z+t_2h)dt_1dt_2\\
&=f_Z^2(z)+O(h^2).
\end{aligned}
$$

Similarly, we can also bound the expectation of the cross product term

$$
\begin{aligned}
&n^{-1}\mathbb{E}\left[u_n\left[g_{z|(j,k)}\right]\times u_n\left[K_h(z_i-z)K_h(z_i-z)\right]\right]\\
&=\mathbb{E}[\mathbb{U}_n\left[g_{z|(j,k)}\right]\mathbb{U}_n\left[K_h(z_i-z)K_h(z_{i'}-z)\right]]-\mathbb{E}[\mathbb{U}_n\left[g_{z|(j,k)}\right]]\mathbb{E}[\mathbb{U}_n\left[K_h(z_i-z)K_h(z_{i'}-z)\right]]\\
&=\frac{4}{n^2(n-1)^2}\sum_{i\neq j,s\neq t}\mathbb{E}\left[K_h(z_i-z)K_h(z_j-z)g_{z|(j,k)}(y_s,y_t)\right]-(f_Z^2(z)\tau_{jk}(z)+O(h^2))^2\\
&=\frac{24}{n^2(n-1)^2}\left\{\binom{n}{4}\mathbb{E}^2\left[K_h(z_i-z)\right]\mathbb{E}\left[\mathbb{U}_n\left[g_{z|(j,k)}\right]\right]\right.\\
&\quad+\binom{n}{3}\mathbb{E}\left[K_h(z_i-z)\right]\mathbb{E}\left[\mathbb{U}_n\left[g_{z|(j,k)}(y_i,y_j)K_h(z_i-z)\right]\right]\\
&\quad\left.+\binom{n}{3}\mathbb{E}\left[\mathbb{U}_n\left[g_{z|(j,k)}(y_i,y_j)K_h(z_i-z)K_h(z_j-z)\right]\right]\right\}-(f_Z^2(z)+O(h^2))^2=O\left(\frac{1}{nh}\right).
\end{aligned}
$$

and the expectation of the square term

$$
\begin{aligned}
&n^{-1}\mathbb{E}\left[\left(u_n\left[K_h(z_i-z)K_h(z_i-z)\right]\right)^2\right]\\
&=\mathbb{E}[\mathbb{U}_n^2\left[K_h(z_i-z)K_h(z_{i'}-z)\right]]-\mathbb{E}^2[\mathbb{U}_n\left[K_h(z_i-z)K_h(z_{i'}-z)\right]]\\
&=\frac{4}{n^2(n-1)^2}\sum_{i\neq j,s\neq t}\mathbb{E}\left[K_h(z_i-z)K_h(z_j-z)K_h(z_s-z)K_h(z_t-z)\right]-(f_Z^2(z)+O(h^2))^2\\
&=\frac{24}{n^2(n-1)^2}\left\{\binom{n}{4}\mathbb{E}^4\left[K_h(z_i-z)\right]+\binom{n}{3}\mathbb{E}^2\left[K_h(z_i-z)\right]\mathbb{E}\left[K_h(z_i-z)\right]^2\right.\\
&\qquad\qquad\left.+\binom{n}{3}\mathbb{E}\left[K_h(z_i-z)\right]^4\right\}-(f_Z^2(z)+O(h^2))^2=O\left(\frac{1}{nh}\right).
\end{aligned}
$$

This completes the proof.

## Appendix F. Proof of Theorem 4

The proof is similar to the proof of Theorem 3. Instead of only bounding the maximum over $(j,k) \in E^c$, we also need to control the supremum over $z \in [z_L, z_U]$. Without loss of generality, we consider the case $E = \emptyset$ and $[z_L, z_U] = (0,1)$. We will apply the multiplier bootstrap theory for continuous suprema developed in Chernozhukov et al. (2014a). Let $W_0$ and its bootstrap counterpart $W_0^B$ be defined as

$$W_0 = \max_{j,k\in[d]} \sup_{z\in(0,1)} \frac{1}{\sqrt{n}} \sum_{i=1}^n J_{z|(j,k)}(Y_i) \text{ and } W_0^B = \max_{j,k\in[d]} \sup_{z\in(0,1)} \frac{1}{\sqrt{n}} \sum_{i=1}^n J_{z|(j,k)}(Y_i) \cdot \xi_i.$$

**Step 1.** We aim to approximate $W_0$ by a Gaussian process. In order to apply Theorem A.1 of Chernozhukov et al. (2014a), we need to study the variance of $J_{z|(j,k)}$ and covering number of the function class $\mathcal{J} = \{J_{z|(j,k)} \mid z \in (0,1), j,k \in [d]\}$.

By the definition of $J_{z|(j,k)}$ in (48), we have

$$\begin{aligned}\sup_{z,j,k} \mathbb{E}[J^2_{z|(j,k)}(Y)] &\le \sup_{z,j,k} 4\|\boldsymbol{\Omega}_j(z)\|_1^2 \|\boldsymbol{\Omega}_k(z)\|_1^2 \cdot \pi^2 h \cdot \Big(\mathbb{E}[\max_{u,v} g^{(1)}_{z|(u,v)}(Y_i)]^2 + \mathbb{E}[\omega_z^{(1)}(Z_i)]^2\Big) \\ &\le CM^2 := \sigma_J^2, \end{aligned} \tag{100}$$

where the last inequality is due to (86) and (95). Similar to (68), we also have for some constant $C > 0$,

$$\sup_{z,j,k} \|J^2_{z|(j,k)}(Y)\|_\infty \le C/h := b_J.$$

Furthermore, by Lemma 30, the covering number of $\mathcal{J}$ satisfies

$$\sup_Q N(\mathcal{J}, L_2(Q), b_J\epsilon) \le \left(\frac{Cd}{\epsilon h}\right)^c.$$

Therefore $\mathcal{J}$ is a VC$(b_J, C(d/h)^c, c)$ type class (see, for example, Definition 3.1 Chernozhukov et al., 2014a) and we can apply Theorem A.1 in Chernozhukov et al. (2014a). Let $K_n = C(\log n \vee \log(d/h))$ for some sufficiently large $C > 0$. Using Theorem A.1 of Chernozhukov et al. (2014a), there exists a random variable $W^0$ such that for any $\gamma \in (0,1)$,

$$\mathbb{P}\bigg(|W_0 - W^0| \ge \frac{K_n/\sqrt{h}}{(\gamma n)^{1/2}} + \frac{(\sigma_J/\sqrt{h})^{1/2} K_n^{3/4}}{\gamma^{1/2} n^{1/4}} + \frac{h^{-1/6}\sigma_J^{2/3} K_n^{2/3}}{\gamma^{1/3} n^{1/6}}\bigg) \le C\Big(\gamma + \frac{\log n}{n}\Big).$$

Choosing $\gamma = (\log^4(dn)/(nh))^{1/8}$, we have

$$\mathbb{P}\big(|W_0 - W^0| > C(\log^4(dn)/(nh))^{1/8}\big) \le C(\log^4(dn)/(nh))^{1/8}. \tag{101}$$

**Step 2.** We next bound the difference between $W_0^B$ and $W^0$. Define

$$\psi_n = \sqrt{\frac{\sigma_J^2 K_n}{n}} + \left(\frac{\sigma_J^2 K_n^3}{nh}\right)^{1/4} \qquad \text{and} \qquad \gamma_n(\delta) = \frac{1}{\delta}\left(\frac{\sigma_J^2 K_n^3}{nh}\right)^{1/4} + \frac{1}{n}.$$

Since $K_n/h \lesssim (\log(dn))/h \lesssim n\sigma_J^2$, Theorem A.2 of Chernozhukov et al. (2014a) gives us

$$\mathbb{P}\Big(|W_0^B - W^0| > \psi_n + \delta \,\Big|\, \{Y_i\}_{i\in[n]}\Big) \le C\gamma_n(\delta),$$

with probability $1-3/n$. Choosing $\delta = (\log(dn))^3/(nh))^{1/8}$, with probability $1-3/n$,

$$\mathbb{P}\Big(|W_0^B - W^0| > C((\log(dn))^3/(nh))^{1/8} \mid \{Y_i\}_{i\in[n]}\Big) \le C((\log d)^3/(nh))^{1/8}. \tag{102}$$

**Step 3.** The last step is to assemble the above results to quantify the difference between $W$ and $W^B$. According to Lemma 19 and Lemma 20, there exists a constant $c > 0$ such that

$$\mathbb{P}(|W - W_0| > n^{-c}) \le n^{-c} \qquad \text{and} \qquad \mathbb{P}(\mathbb{P}_\xi\big(|W_G^B - W_0^B| > n^{-c}) > n^{-c}\big) \le n^{-c}. \tag{103}$$

Let $q_n := [(\log(dn))^3/(nh)]^{1/8}$. From (103) and (101), we have

$$\mathbb{P}\big(|W - W^0| > q_n\big) \le q_n. \tag{104}$$

Define the event

$$\mathcal{W} = \Big\{\mathbb{P}\big(|W_G^B - W^0| > q_n \,\Big|\, \{Y_i\}_{i\in[n]}\big) \le q_n\Big\}. \tag{105}$$

Using (103) and (102), $\mathbb{P}(\mathcal{W}) \ge 1 - n^{-c}$. Recall that $\widehat{c}_W(1-\alpha, \{Y_i\}_{i=1}^n)$ is $(1-\alpha)$-quantile of $W_G^B$ conditionally on $\{Y_i\}_{i=1}^n$. Let

$$W_0^B(\{Y_i\}_{i=1}^n) = \max_{j,k\in[d]} \sup_{z\in(0,1)} \; n^{-1/2}\sum_{i=1}^n J_{z|(j,k)}(Y_i)\cdot \xi_i$$

and define $W^B(\{Y_i\}_{i=1}^n)$ similarly. Let $\widehat{\sigma}^2_{z,j,k} = n^{-1}\sum_{i=1}^n J^2_{z|(j,k)}(Y_i)$, $\underline{\sigma}_J = \inf_{z,j,k}\widehat{\sigma}_{z,j,k}$ and $\overline{\sigma}_J = \sup_{z,j,k}\widehat{\sigma}_{z,j,k}$. Using the triangle inequality, we have

$$\begin{aligned}
\mathbb{P}\big(W \le \widehat{c}_W(1-\alpha, \{Y_i\}_{i=1}^n)\big) &\overset{(104)}{\ge} \mathbb{P}\big(W^0 \le \widehat{c}_W(1-\alpha,\{Y_i\}_{i=1}^n) - q_n\big) - q_n \\
&\overset{(105)}{\ge} \mathbb{P}\big(W_0^B(\{Y_i\}_{i=1}^n) \le \widehat{c}_W(1-\alpha,\{Y_i\}_{i=1}^n) - 2q_n\big) - 2q_n \\
&\ge \mathbb{P}\big(W^B(\{Y_i\}_{i=1}^n) \le \widehat{c}_W(1-\alpha,\{Y_i\}_{i=1}^n)\big) \\
&\quad - C(\underline{\sigma}_J, \overline{\sigma}_J) q_n(\mathbb{E}_\xi[W_0^B] + \sqrt{1\vee \log(n\underline{\sigma}_J)}) - Cq_n,
\end{aligned} \tag{106}$$

where the last inequality follows from the anti-concentration for suprema of Gaussian processes, given in Lemma A.1 of Chernozhukov et al. (2014b), and $C(\underline{\sigma}_J, \overline{\sigma}_J)$ is a constant that only depends on $\underline{\sigma}_J$ and $\overline{\sigma}_J$.

In the following of the proof, we will bound the right hand side of (106). We first show the constant $C(\underline{\sigma}_J, \overline{\sigma}_J)$ is independent to $h, n$ or $d$ and then bound $\mathbb{E}_\xi[W^B]$. We bound $\mathbb{E}_\xi[W^B]$ by bounding $\mathbb{E}_\xi[W_0^B]$ and $\mathbb{E}_\xi[|W^B - W_0^B|]$. Since $W_0^B$ is a supreme of a Gaussian process given data, we can bound its expectation by quantifying its conditional variance and related covering number. We first bound its conditional variance $n^{-1}\sum_{i=1}^n J^2_{z|(j,k)}(Y_i)$. Using the notation defined in (67), we have

$$\sup_{z,j,k}\frac{1}{n}\sum_{i=1}^n (J^2_{z|(j,k)}(Y_i) - \mathbb{E}[J^2_{z|(j,k)}(Y_i)]) = \sup_{z,j,k}\frac{1}{n}\sum_{i=1}^n \gamma_{z|(j,k,j,k)}(Y_i).$$

Therefore, (69) and (70) give an upper bound and variance of $(J^2_{z|(j,k)}(Y_i) - \mathbb{E}[J^2_{z|(j,k)}(Y_i)])$. Define the function class $\mathcal{J}_{(2)} = \Big\{J^2_{z|(j,k)} \mid z \in (0,1), j,k \in [d]\Big\}$. Using Lemma 26 and Lemma 30,

$$\sup_Q N(\mathcal{J}_{(2)}, \|\cdot\|_{L_2(Q)}, \epsilon/\sqrt{h}) \leq \left(\frac{Cd}{h\epsilon}\right)^c.$$

As the upper bound, variance and covering number are quantified above, similar to (88), we apply Theorem 3.12 of Koltchinskii (2011) to get

$$\mathbb{E}\left[\sup_{z,j,k} \frac{1}{n}\sum_{i=1}^n (J^2_{z|(j,k)}(Y_i) - \mathbb{E}[J^2_{z|(j,k)}(Y_i)])\right] \lesssim \sqrt{\frac{\log(2d/h)}{nh^2}} + \frac{\log(2d/h)}{nh}.$$

Under the assumptions of the theorem, $\sqrt{\log(2d/h)/(nh^2)} + \log(2d/h)/(nh) = O(n^{-2c})$ and the Markov's inequality give

$$\mathbb{P}\left(\sup_{z,j,k} \frac{1}{n}\sum_{i=1}^n (J^2_{z|(j,k)}(Y_i) - \mathbb{E}[J^2_{z|(j,k)}(Y_i)]) > n^{-c}\right) \leq Cn^{-c}.$$

Combining with (100),

$$\overline{\sigma}^2_J = \sup_{z,j,k} \frac{1}{n}\sum_{i=1}^n J^2_{z|(j,k)}(Y_i) \leq \sigma^2_J + n^{-c} \leq 2\sigma^2_J,$$

with probability $1 - Cn^{-c}$. By the assumption in Theorem 4,

$$\inf_{z,j,k} \mathbb{E}[J^2_{z|(j,k)}(Y)] = \inf_{z,j,k} \mathrm{Var}(\boldsymbol{\Omega}_j(z)\boldsymbol{\Theta}_z\boldsymbol{\Omega}_k(z)) > c > 0.$$

Therefore, we have

$$\begin{aligned}\underline{\sigma}_J = \inf_{z,j,k} \frac{1}{n}\sum_{i=1}^n J^2_{z|(j,k)}(Y_i) \\ \geq \inf_{z,j,k} \mathbb{E}[J^2_{z|(j,k)}(Y)] - \sup_{z,j,k} \frac{1}{n}\sum_{i=1}^n (J^2_{z|(j,k)}(Y_i) - \mathbb{E}[J^2_{z|(j,k)}(Y_i)]) \geq c/2 > 0\end{aligned}$$

with probability $1 - Cn^{-c}$. The constant $C(\underline{\sigma}_J, \overline{\sigma}_J)$ does not depend on $n, d$ and $h$.

Combining Lemma 2.2.8 in van der Vaart and Wellner (1996) and Lemma 30, we have

$$\mathbb{E}_\xi[W^B_0] \leq C\sigma_J\sqrt{\log\left(Cd\sigma^{-1}_J/h\right)} \leq C\sqrt{\log(d/h)}.$$

From Lemma 20, we have $\mathbb{E}_\xi[W^B] \leq \mathbb{E}_\xi[W^B_0] + \mathbb{E}_\xi[|W^B - W^B_0|] \leq C\sqrt{\log(d/h)}$, since $q_n\sqrt{\log(d/h)} = (\log^8(d/h)/(nh))^{1/8} = O(n^{-c})$. Due to (106) and the fact that $\mathbb{P}(\mathcal{W}) \geq 1 - n^{-c}$, we have $\mathbb{P}\big(W \leq c_W(1-\alpha)\big) \geq 1 - \alpha - 3n^{-c}$. Similarly, we can also show that $\mathbb{P}\big(W \geq \widehat{c}_W(1-\alpha, \{Y_i\}_{i=1}^n)\big) \geq \alpha - 3n^{-c}$, which completes the proof.

## Appendix G. Auxiliary Lemmas for Score Statistics

In this section, we provide the auxiliary results for proving auxiliary lemmas on the asymptotic properties on the score statistics.

### G.1 Approximation Error for Score Statistics

In this section, we prove Lemma 19 and Lemma 20, which approximate the score statistics by a leading linear term.

**Lemma 19** *Under the same conditions as Theorem 4, there exists a universal constant $c > 0$ such that*

$$\sup_{j,k\in[d]} \sup_{z\in(0,1)} \left| \sqrt{nh} \cdot \mathbb{U}_n[\omega_z] \widehat{S}_{z|(j,k)}\big(\widehat{\mathbf{\Omega}}_{k\setminus j}(z)\big) - \mathbb{G}_n\big[J_{z|(j,k)}\big] \right| \leq n^{-c}, \tag{107}$$

*with probability $1 - c/d$.*

**Proof** We have

$$\begin{aligned}
&\left| \sqrt{nh} \cdot \mathbb{U}_n[\omega_z] \widehat{S}_{z|(j,k)} \left(\widehat{\mathbf{\Omega}}_{k\setminus j}(z)\right) - \mathbb{G}_n\left[J_{z|(j,k)}\right] \right| \\
&\quad \leq \underbrace{\sqrt{nh} \cdot \mathbb{U}_n[\omega_z] \left| \widehat{S}_{z|(j,k)} \left(\widehat{\mathbf{\Omega}}_{k\setminus j}(z)\right) - \mathbf{\Omega}_j^T(z) \left(\widehat{\mathbf{\Sigma}}(z)\mathbf{\Omega}_k(z) - \mathbf{e}_k^T\right) \right|}_{I} \\
&\qquad + \underbrace{\left| \sqrt{nh} \cdot \mathbb{U}_n[\omega_z] \mathbf{\Omega}_j^T(z) \left(\widehat{\mathbf{\Sigma}}(z)\mathbf{\Omega}_k(z) - \mathbf{e}_k^T\right) - \mathbb{G}_n\left[J_{z|(j,k)}\right] \right|}_{II}
\end{aligned}$$

A bound on $I$ is obtained in Lemma 21. Here, we proceed to obtain a bound on $II$.

To simplify the notation, we sometimes omit the argument $z_0$, that is, we write $\mathbf{\Omega}(z_0)$ as $\mathbf{\Omega}$ and similarly for other parameters indexed by $z$. Applying the Taylor expansion to $\sin(\cdot)$ we obtain

$$\begin{aligned}
&\sqrt{nh} \cdot \mathbf{\Omega}_j^T(z) \left(\widehat{\mathbf{\Sigma}}(z)\mathbf{\Omega}_k(z) - \mathbf{e}_k\right) = \sqrt{nh} \cdot \mathbf{\Omega}_j^T(z) \left(\widehat{\mathbf{\Sigma}}(z) - \mathbf{\Sigma}(z)\right) \mathbf{\Omega}_k(z) \\
&= \sqrt{nh} \sum_{a,b\in[d]} \mathbf{\Omega}_{ja}\mathbf{\Omega}_{bk} \left( \sin\left(\widehat{\tau}_{ab}\frac{\pi}{2}\right) - \sin\left(\tau_{ab}\frac{\pi}{2}\right)\right) \\
&= \underbrace{\sqrt{nh} \sum_{a,b\in[d]} \mathbf{\Omega}_{ja}\mathbf{\Omega}_{bk} \cos\left(\tau_{ab}\frac{\pi}{2}\right) \frac{\pi}{2}\left(\widehat{\tau}_{ab} - \tau_{ab}\right)}_{T_1} \\
&\qquad\qquad\qquad \underbrace{- \frac{\sqrt{nh}}{2} \sum_{a,b\in[d]} \mathbf{\Omega}_{ja}\mathbf{\Omega}_{bk} \sin\left(\widetilde{\tau}_{ab}\frac{\pi}{2}\right) \left(\frac{\pi}{2}\left(\widehat{\tau}_{ab} - \tau_{ab}\right)\right)^2}_{T_2},
\end{aligned} \tag{108}$$

where $\widetilde{\tau}_{ij} = (1-\alpha)\widehat{\tau}_{ij} + \alpha\tau_{ij}$ for some $\alpha \in (0,1)$.

In order to study the properties of $T_1$ and $T_2$, we first analyze the $\sqrt{nh}(\widehat{\tau}_{ab} - \tau_{ab})$ shared by both terms. Let

$$r_{z|(a,b)}(Y_i, Y_{i'}) := g_{z|(a,b)}(Y_i, Y_{i'}) - \tau_{ab}(z)\omega_z(Z_i, Z_{i'}), \tag{109}$$

where $g_{z|(a,b)}$ is defined in (40). From (4), we have

$$\sqrt{nh}\big(\widehat{\tau}_{ab}(z) - \tau_{ab}(z)\big) = \frac{\sqrt{nh} \cdot \sum_{i \neq i'} r_{z|(a,b)}(Y_i, Y_{i'})}{\sum_{i \neq i'} \omega_z(Z_i, Z_{i'})}.$$

We divide the $U$-statistic in the numerator into two parts

$$\frac{1}{\sqrt{nh}(n-1)} \sum_{i \neq i'} r_{z|(a,b)}(Y_i, Y_{i'}) = \sqrt{h} \cdot u_n[r_{z|(a,b)}] + \sqrt{nh} \cdot \mathbb{E}[r_{z|(a,b)}(Y_i, Y_{i'})].$$

For any $z \in (0,1)$ and $a, b \in [d]$, using Lemma 13 we obtain

$$\begin{aligned}
&\sqrt{nh} \cdot \mathbb{E}[r_{z|(a,b)}(Y_i, Y_{i'})] \\
&\qquad = \sqrt{nh} \cdot \mathbb{E}\left[\mathbb{U}_n\left[g_{z|(a,b)}\right]\right] - \sqrt{nh} \cdot \tau_{ab}(z) \mathbb{E}\left[\mathbb{U}_n\left[K_h(Z_i - z) K_h(Z_{i'} - z)\right]\right] \\
&\qquad = \sqrt{nh}\big[f_Z^2(z)\tau_{ab}(z) + O(h^2)\big] - \sqrt{nh}\tau_{ab}(z)\big[f_Z^2(z) + O(h^2)\big] \\
&\qquad = O(\sqrt{nh^5}).
\end{aligned} \tag{110}$$

For the leading term $\sqrt{h} \cdot u_n[r_{z|(a,b)}]$, we combine (109) with (43) and (46), to obtain

$$\sqrt{h} \cdot u_n[r_{z|(a,b)}] = 2\sqrt{nh} \cdot \mathbb{E}_n\left[g^{(1)}_{z|(a,b)} - \tau_{ab}(z)\omega^{(1)}_z\right] + \sqrt{nh} \cdot \mathbb{U}_n\left[g^{(2)}_{z|(a,b)} - \tau_{ab}(z)\omega^{(2)}_z\right],$$

where $g^{(1)}_{z|(a,b)}$, $g^{(2)}_{z|(a,b)}$, $\omega^{(1)}_z$, $\omega^{(2)}_z$ are defined in (41), (42), (44), (45) respectively. With this,

$$\begin{aligned}
T_1 &= [\mathbb{U}_n[\omega_z]]^{-1} \sum_{a,b \in [d]} \mathbf{\Omega}_{ja} \mathbf{\Omega}_{bk} \frac{\pi}{2} \cos\left(\tau_{ab}\frac{\pi}{2}\right) \left(\sqrt{h} \cdot u_n[r_{z|(a,b)}] + \sqrt{nh} \cdot \mathbb{E}[r_{z|(a,b)}(Y_i, Y_{i'})]\right) \\
&= [\mathbb{U}_n[\omega_z]]^{-1}(T_{11} + T_{12} + T_{13}),
\end{aligned}$$

where

$$\begin{aligned}
T_{11} &= \sqrt{nh} \sum_{a,b \in [d]} \mathbf{\Omega}_{ja} \mathbf{\Omega}_{bk} \pi \cos\left(\tau_{ab}\frac{\pi}{2}\right) \mathbb{E}_n\left[g^{(1)}_{z|(a,b)} - \tau_{ab}(z)\omega^{(1)}_z\right], \\
T_{12} &= \sqrt{nh} \sum_{a,b \in [d]} \mathbf{\Omega}_{ja} \mathbf{\Omega}_{bk} \frac{\pi}{2} \cos\left(\tau_{ab}\frac{\pi}{2}\right) \mathbb{U}_n\left[g^{(2)}_{z|(a,b)} - \tau_{ab}(z)\omega^{(2)}_z\right] \quad \text{and} \\
T_{13} &= \sqrt{nh} \sum_{a,b \in [d]} \mathbf{\Omega}_{ja} \mathbf{\Omega}_{bk} \frac{\pi}{2} \cos\left(\tau_{ab}\frac{\pi}{2}\right) \mathbb{E}[r_{z|(a,b)}(Y_i, Y_{i'})].
\end{aligned}$$

From (48), we have that $T_{11} = \mathbb{G}_n[J_{z|(j,k)}]$. We proceed to bound the other terms. Using Lemma 16 and Lemma 18, we have

$$\sup_{z,j,k} |T_{12}| \leq \sup_{z,j,k} \sqrt{nh} \cdot \|\mathbf{\Omega}_j\|_1 \|\mathbf{\Omega}_k\|_1 \max_{a,b \in [d]} \left(\left|\mathbb{U}_n\left[g^{(2)}_{z|(a,b)}\right]\right| + \left|\mathbb{U}_n\left[\omega^{(2)}_z\right]\right|\right) \lesssim \frac{\log(d/h)}{\sqrt{nh}}, \tag{111}$$

with probability $1 - 1/d$. Using (110), we can bound $T_{13}$ as

$$\sup_{z,j,k} |T_{13}| \lesssim \frac{\pi}{2} \|\mathbf{\Omega}_j\|_1 \|\mathbf{\Omega}_k\|_1 \cdot (\sqrt{nh^5}) \leq \pi M^2 \sqrt{nh^5}. \tag{112}$$

The final step is to bound $T_2$. Using Lemma 9 and Lemma 10,

$$\sup_{z,j,k} |T_2| \le \frac{\pi^2}{8} \|\boldsymbol{\Omega}_j\|_1 \|\boldsymbol{\Omega}_k\|_1 \max_{z,a,b} \sqrt{nh} \cdot |\widehat{\tau}_{ab}(z) - \tau_{ab}(z)|^2$$
$$\le CM^2 \sqrt{nh} \cdot \left( \frac{\log(d/h)}{nh} + h^4 + \frac{1}{n^2h^2} \right) \tag{113}$$

with probability $1 - 1/d$.

Combining (55), (111), (112), and (113) with (108), we finally have

$$\sup_{j,k\in[d]} \sup_{z\in(0,1)} \left| \sqrt{nh} \cdot \mathbb{U}_n[\omega_z] \widehat{S}_{z|(j,k)} \big( \widehat{\boldsymbol{\Omega}}_{k\setminus j}(z) \big) - \mathbb{G}_n \big( J_{z|(j,k)} \big) \right|$$
$$\le \sup_{j,k\in[d]} \sup_{z\in(0,1)} \big( |T_{12}| + |T_{13}| + |\mathbb{U}_n[\omega_z]||T_2| \big) \lesssim \frac{\log(d/h)}{\sqrt{nh}} + \sqrt{nh^5}.$$

with probability $1-1/d$. Under the assumptions of the lemma, there exists a constant $c > 0$ such that $\log(d/h)/\sqrt{nh} = o(n^{-c})$ and $\sqrt{nh^5} = o(n^{-c})$, which completes the proof.

■

The following lemma states a result analogous to Lemma 19 for the bootstrap test statistic.

**Lemma 20** *Under the same conditions as Theorem 4, there exists a universal constant $c > 0$ such that for sufficiently large $n$,*

$$\mathbb{P}_\xi \left( \sup_{j,k\in[d]} \sup_{z\in(0,1)} \left| \sqrt{nh} \cdot \mathbb{U}_n[\omega_z^B] \widehat{S}^B_{z|(j,k)} \big( \widehat{\boldsymbol{\Omega}}_{k\setminus j}(z) \big) - \mathbb{G}_n^\xi \left[ J_{z|(j,k)} \right] \right| \le n^{-c} \right) \ge 1 - c/d.$$

**Proof** We have

$$\left| \sqrt{nh} \cdot \mathbb{U}_n[\omega_z^B] \widehat{S}^B_{z|(j,k)} \left( \widehat{\boldsymbol{\Omega}}_{k\setminus j}(z) \right) - \mathbb{G}_n^\xi \left( J_{z|(j,k)} \right) \right|$$
$$\le \underbrace{\sqrt{nh} \cdot \mathbb{U}_n[\omega_z^B] \left| \widehat{S}^B_{z|(j,k)} \left( \widehat{\boldsymbol{\Omega}}^B_{k\setminus j}(z) \right) - \boldsymbol{\Omega}_j^T(z) \left( \widehat{\boldsymbol{\Sigma}}(z) \boldsymbol{\Omega}_k(z) - \mathbf{e}_k^T \right) \right|}_{I}$$
$$+ \underbrace{\left| \sqrt{nh} \cdot \mathbb{U}_n[\omega_z^B] \boldsymbol{\Omega}_j^T(z) \left( \widehat{\boldsymbol{\Sigma}}^B(z) \boldsymbol{\Omega}_k(z) - \mathbf{e}_k^T \right) - \mathbb{G}_n^\xi \left[ J_{z|(j,k)} \right] \right|}_{II}.$$

Lemma 22 gives a bound on $I$. Here, we focus on obtaining a bound on $II$. That is, we show that $\sqrt{nh} \cdot \mathbb{U}_n[\omega_z^B] \boldsymbol{\Omega}_j^T(z) \big( \widehat{\boldsymbol{\Sigma}}^B(z) \boldsymbol{\Omega}_k(z) - \mathbf{e}_k^T \big)$ is close to the linear leading term

$$T_0^B := \mathbb{G}_n^\xi \left[ J_{z|(j,k)} \right] = \frac{1}{\sqrt{n}} \sum_{i=1}^n J_{z|(j,k)}(Y_i) \cdot \xi_i.$$

Similar to (108), we have

$$\max_{j,k\in[d]} \sup_{z\in(0,1)} \sqrt{nh} \cdot \mathbb{U}_n[\omega_z^B] \boldsymbol{\Omega}_j(z_0)^T \left( \widehat{\boldsymbol{\Sigma}}^B(z) \boldsymbol{\Omega}_k(z) - \mathbf{e}_k^T \right) = T_1^B + T_2^B,$$

where

$$T_1^B = \sqrt{nh} \sum_{a,b \in [d]} \mathbf{\Omega}_{ja} \mathbf{\Omega}_{bk} \cos\left(\tau_{ab} \frac{\pi}{2}\right) \frac{\pi}{2} \mathbb{U}_n[\omega_z^B] \left(\widehat{\tau}_{ab}^B - \tau_{ab}\right)$$

and

$$T_2^B = -\frac{\sqrt{nh}}{2} \sum_{a,b \in [d]} \mathbf{\Omega}_{ja} \mathbf{\Omega}_{bk} \sin\left(\tau_{ab} \frac{\pi}{2}\right) \mathbb{U}_n[\omega_z^B] \left(\frac{\pi}{2}\left(\widehat{\tau}_{ab}^B - \tau_{ab}\right)\right)^2.$$

We bound $T_2^B$ first. Note that

$$\underline{\mathrm{f}}_Z^2 - \sup_{z \in (0,1)} |\mathbb{U}_n[\omega_z^B] - f_Z^2(z)| \leq \inf_{z \in (0,1)} \mathbb{U}_n[\omega_z^B] \leq \sup_{z \in (0,1)} \mathbb{U}_n[\omega_z^B] \leq \bar{\mathrm{f}}_Z^2 + \sup_{z \in (0,1)} |\mathbb{U}_n[\omega_z^B] - f_Z^2(z)|.$$

Using Lemma 24 and $\log(h^{-1})/(nh^2) = o(1)$, we have

$$\underline{\mathrm{f}}_Z^2/2 \leq \inf_{z \in (0,1)} \mathbb{U}_n[\omega_z] \leq \sup_{z \in (0,1)} \mathbb{U}_n[\omega_z] \leq 2\bar{\mathrm{f}}_Z^2, \tag{114}$$

with probability at least $1 - 1/n$. Similar to (113), (114) and the Hölder's inequality give us

$$|T_2^B| \leq \frac{\pi^2}{4\underline{\mathrm{f}}_Z^2} \|\mathbf{\Omega}_j\|_1 \|\mathbf{\Omega}_k\|_1 \max_{a,b \in [d]} \sqrt{nh} \cdot |\mathbb{U}_n[\omega_z^B](\widehat{\tau}_{ab}^B - \tau_{ab})|^2.$$

Under the assumptions, $h \asymp n^{-\delta}$ for $\delta \in (1/5, 1/4)$, and Lemma 23 gives us

$$\mathbb{P}_\xi\Big( \max_{z,j,k} |T_2^B| \leq C \log(d/h)/\sqrt{nh^3} \Big) \geq 1 - 1/d \tag{115}$$

for some constant $C > 0$ with probability $1 - c/d$.

Next, we handle the difference between $T_1^B$ and $T_0^B$. We denote

$$\Delta W_{z|(a,b)} = \mathbb{U}_n[\omega_z^B] \left(\widehat{\tau}_{ab}^B(z) - \tau_{ab}(z)\right) - \frac{2}{n} \sum_{i=1}^n \left[g_{z|(a,b)}^{(1)}(Y_i) - \tau_{ab}(z)\omega_z^{(1)}(Z_i)\right]\xi_i. \tag{116}$$

Using the Hölder's inequality, we have

$$|T_1^B - T_0^B| \leq \left| \sqrt{nh} \sum_{a,b \in [d]} \mathbf{\Omega}_{ja} \mathbf{\Omega}_{bk} \cos\left(\tau_{ab} \frac{\pi}{2}\right) \frac{\pi}{2} \cdot \Delta W_{z|(a,b)} \right| \leq C M^2 \sqrt{nh} \max_{a,b} |\Delta W_{z|(a,b)}|.$$

Combining with (123), there exists a constant $c > 0$ such that with probability $1 - c/d$

$$\mathbb{P}_\xi\big( \sup_{z \in (0,1)} \max_{j,k \in [d]} |T_1^B - T_0^B| \geq C\sqrt{\log(d/h)/(nh^2)} \big) \leq 1/d. \tag{117}$$

If $\log(d/h)/(nh^2) \asymp n^{-c}$, (120), (115) and (117) give us with probability $1 - c/d$

$$\mathbb{P}_\xi\Big( \sup_{z \in (0,1)} \max_{j,k \in [d]} |\widehat{S}_{z|(j,k)}^B\big(\widehat{\mathbf{\Omega}}_{k \setminus j}(z)\big) - T_0^B| \leq n^{-c} \Big) \geq 1 - 1/d,$$

which completes the proof of the lemma.

■

### G.2 First Step Approximation Results

Here we establish results needed for the first step in the proofs of Lemma 19 and Lemma 20.

**Lemma 21** *Under the same conditions as Theorem 4, there exists a universal constant $c > 0$ such that*

$$\sup_{j,k\in[d]}\sup_{z\in(0,1)}\sqrt{nh}\cdot\left|\widehat{S}_{z|(j,k)}\left(\widehat{\boldsymbol{\Omega}}_{k\setminus j}(z)\right)-\boldsymbol{\Omega}_j^T(z)\left(\widehat{\boldsymbol{\Sigma}}(z)\boldsymbol{\Omega}_k(z)-\mathbf{e}_k^T\right)\right|\le n^{-c}$$

*with probability $1 - c/d$.*

**Proof** For any matrix $\mathbf{A} = (\mathbf{A}_1, \ldots, \mathbf{A}_d) \in \mathbb{R}^{d\times d}$, we define

$$\mathbf{A}_{-j} := (\mathbf{A}_1, \ldots, \mathbf{A}_{j-1}, \mathbf{A}_{j+1}, \ldots, \mathbf{A}_d) \in \mathbb{R}^{d\times(d-1)},$$

which is a submatrix of $\mathbf{A}$ with column $j$ removed. We also denote

$$\begin{aligned}\boldsymbol{\gamma}^*(z) &= (\boldsymbol{\Omega}_{k1}(z), \ldots, \boldsymbol{\Omega}_{k(j-1)}(z), \boldsymbol{\Omega}_{k(j+1)}(z)\ldots\boldsymbol{\Omega}_{kd}(z))^T \in \mathbb{R}^{d-1} \quad \text{and}\\ \widehat{\boldsymbol{\gamma}}(z) &= (\widehat{\boldsymbol{\Omega}}_{k1}(z), \ldots, \widehat{\boldsymbol{\Omega}}_{k(j-1)}(z), \widehat{\boldsymbol{\Omega}}_{k(j+1)}(z), \ldots, \widehat{\boldsymbol{\Omega}}_{kd}(z))^T \in \mathbb{R}^{d-1}.\end{aligned}$$

To simplify the notation, we sometimes omit the varying variable $z$ in the proof.

We start by decomposing the score function into two parts. We will then identify the leading term and bound the remainder. With the above introduced notation, we have

$$\begin{aligned}&\sqrt{nh}\cdot\widehat{S}_{z|(j,k)}(\widehat{\boldsymbol{\Omega}}_{k\setminus j})\\ &\quad=\sqrt{nh}\cdot\widehat{\boldsymbol{\Omega}}_j^T\left(\widehat{\boldsymbol{\Sigma}}_{-j}\widehat{\boldsymbol{\gamma}}-\mathbf{e}_k^T\right)\\ &\quad=\sqrt{nh}\cdot\boldsymbol{\Omega}_j^T\left(\widehat{\boldsymbol{\Sigma}}_{-j}\boldsymbol{\gamma}^*-\mathbf{e}_k^T\right)+\underbrace{\sqrt{nh}\cdot\left(\widehat{\boldsymbol{\Omega}}_j^T-\boldsymbol{\Omega}_j^T\right)\left(\widehat{\boldsymbol{\Sigma}}_{-j}\boldsymbol{\gamma}^*-\mathbf{e}_k^T\right)}_{I_1}+\underbrace{\sqrt{nh}\cdot\widehat{\boldsymbol{\Omega}}_j^T\widehat{\boldsymbol{\Sigma}}_{-j}\left(\widehat{\boldsymbol{\gamma}}-\boldsymbol{\gamma}^*\right)}_{I_2}.\end{aligned}$$

The first term in the display above is the leading term on the right hand side of (107) as desired. The remaining part of the proof is to bound $I_1$ and $I_2$. By the Hölder's inequality, we have

$$|I_1| = \sqrt{nh}\cdot\left(\widehat{\boldsymbol{\Omega}}_j-\boldsymbol{\Omega}_j\right)^T\left(\widehat{\boldsymbol{\Sigma}}_{-j}-\boldsymbol{\Sigma}_{-j}\right)\boldsymbol{\gamma}^*\le\sqrt{nh}\cdot\left\|\widehat{\boldsymbol{\Omega}}_j-\boldsymbol{\Omega}_j\right\|_1\left\|\widehat{\boldsymbol{\Sigma}}-\boldsymbol{\Sigma}\right\|_{\max}\|\boldsymbol{\Omega}_k\|_1.$$

Assumption 4.4 together with the display above gives

$$\sup_{z\in(0,1)}\max_{j,k\in[d]}|I_1|\le M\sqrt{nh}\cdot r_{1n}r_{2n} \tag{118}$$

with probability $1 - 1/d$. Next, using the Hölder's inequality we have

$$|I_2|\le\sqrt{nh}\cdot\|\widehat{\boldsymbol{\Omega}}_j^T\widehat{\boldsymbol{\Sigma}}_{-j}\|_\infty\left\|\widehat{\boldsymbol{\gamma}}-\boldsymbol{\gamma}^*\right\|_1.$$

From Assumption 4.4, we have $\|\widehat{\boldsymbol{\Omega}}_j^T\widehat{\boldsymbol{\Sigma}}_{-j}\|_\infty\le\|\widehat{\boldsymbol{\Omega}}_j^T\widehat{\boldsymbol{\Sigma}}\|_\infty\le r_{3n}$ with probability $1-1/d$ and, therefore,

$$\sup_{z\in(0,1)}\max_{j,k\in[d]}|I_2|\le\sqrt{nh}\cdot r_{3n}r_{2n} \tag{119}$$

with probability $1-1/d$. Combining (118) and (119), we have

$$\sup_{j,k\in[d]}\sup_{z\in(0,1)}\big(|I_1|+|I_2|\big)\le\sqrt{nh}\cdot r_{2n}(r_{1n}+r_{3n})\le n^{-c}$$

with probability $1-2/d$, which completes the proof.

■

**Lemma 22** *Under the same conditions as Theorem 4, there exists a universal constant $c>0$ such that with probability $1-1/d$,*

$$\mathbb{P}_\xi\bigg(\max_{j,k\in[d]}\sup_{z\in(0,1)}\sqrt{nh}\cdot\Big|\widehat{S}^B_{z|(j,k)}\big(\widehat{\boldsymbol{\Omega}}_{k\backslash j}(z)\big)-\boldsymbol{\Omega}_j^T(z)\Big(\widehat{\boldsymbol{\Sigma}}^B(z)\boldsymbol{\Omega}_k(z)-\mathbf{e}_k^T\Big)\Big|\le n^{-c}\bigg)\ge 1-1/d. \tag{120}$$

**Proof** The proof is similar to the proof of Lemma 21. Compared to the proof in Lemma 21, there are two differences: (1) we need to bound $\mathbb{U}_n[\omega_z^B]\|\widehat{\boldsymbol{\Sigma}}^B-\boldsymbol{\Sigma}\|_{\max}$ instead of $\|\widehat{\boldsymbol{\Sigma}}-\boldsymbol{\Sigma}\|_{\max}$ and (2) obtain a rate for $\mathbb{U}_n[\omega_z^B]\|\widehat{\boldsymbol{\Omega}}_j^T\widehat{\boldsymbol{\Sigma}}^B\|_\infty$ instead of $\|\widehat{\boldsymbol{\Omega}}_j^T\widehat{\boldsymbol{\Sigma}}\|_\infty$.

According to Lemma 23 and (114), we have

$$\begin{aligned}
&\mathbb{P}_\xi\Big(\sup_{z\in(0,1)}\mathbb{U}_n[\omega_z^B]\|\widehat{\boldsymbol{\Sigma}}^B-\boldsymbol{\Sigma}\|_{\max}>C\sqrt{\log(d/h)/(nh^2)}\Big)\\
&\le\mathbb{P}_\xi\bigg(\max_{j,k\in[d]}\sup_{z\in(0,1)}\Big|\mathbb{U}_n[\omega_z^B]\big(\widehat{\tau}^B_{jk}(z)-\tau_{jk}(z)\big)\Big|>C\sqrt{\log(d/h)/(nh^2)}\bigg)\le 1/d.
\end{aligned}\tag{121}$$

Next, (114) and the Hölder's inequality, give us

$$\mathbb{U}_n[\omega_z^B]\|\widehat{\boldsymbol{\Omega}}_j^T\widehat{\boldsymbol{\Sigma}}^B\|_\infty\le 2\bar{\mathrm{f}}_Z\Big(\|\widehat{\boldsymbol{\Omega}}_j^T\widehat{\boldsymbol{\Sigma}}\|_\infty+\big(\|\widehat{\boldsymbol{\Omega}}_j-\boldsymbol{\Omega}_j\|_1+\|\boldsymbol{\Omega}_j\|_1\big)\mathbb{U}_n[\omega_z^B]\|\widehat{\boldsymbol{\Sigma}}^B-\widehat{\boldsymbol{\Sigma}}\|_{\max}\Big).$$

Therefore, by Assumption 4.4, (121) and Theorem 6, with probability $1-1/d$,

$$\mathbb{P}_\xi\Big(\mathbb{U}_n[\omega_z^B]\|\widehat{\boldsymbol{\Omega}}_j\widehat{\boldsymbol{\Sigma}}^B\|_\infty>C\big(M\lambda+M\sqrt{\log(d/h)/(nh^2)}\big)\Big)\le 1/d. \tag{122}$$

Compared to the rate on $I_1$ in (118), we have

$$I_{12}^B:=\sqrt{nh}\cdot\big(\widehat{\boldsymbol{\Omega}}_j-\boldsymbol{\Omega}_j\big)^T\big(\widehat{\boldsymbol{\Sigma}}^B_{-j}-\boldsymbol{\Sigma}_{-j}\big)\boldsymbol{\gamma}^*\le\sqrt{nh}\cdot\big\|\widehat{\boldsymbol{\Omega}}_j-\boldsymbol{\Omega}_j\big\|_1\big\|\widehat{\boldsymbol{\Sigma}}^B-\boldsymbol{\Sigma}\big\|_{\max}\|\boldsymbol{\gamma}^*\|_1.$$

For $\lambda=\kappa\sqrt{\log(dn)}\cdot(h^2+1/\sqrt{nh})$ and $h=n^{-\delta}$, for $1/5<\delta<1/4$, by (121), we have with probability $1-1/d$, there exists a constant $c$ such that

$$\mathbb{P}_\xi\Big(\sup_{z,j,k}|I_{12}^B|>sM^2\lambda\sqrt{\log(d/h)/h}\Big)\le 1/d.$$

Instead of $I_2$ in (119), we define $I_2^B:=\sqrt{nh}\cdot\mathbb{U}_n^B(\omega_z)\|\widehat{\boldsymbol{\Omega}}_j\widehat{\boldsymbol{\Sigma}}^B\|_\infty\big\|\widehat{\boldsymbol{\gamma}}-\boldsymbol{\gamma}^*\big\|_1$. By (122) and Theorem 6, we have with probability $1-1/d$,

$$\mathbb{P}_\xi\Big(\sup_{z,j,k}|I_2^B|\le C\sqrt{nh}\cdot sM^2\lambda(\lambda+\sqrt{\log(d/h)(nh^2)})\Big)\le 1/d.$$

For $\lambda = C_{\mathbf{\Sigma}}\big(h^2 + \sqrt{\log(d/h)/(nh)}\,\big)$, as $s \log d/\sqrt{nh^3} = o(n^{-c})$ and $\sqrt{nh^5} = o(n^{-c})$, we have with probability $1 - 1/d$,

$$\mathbb{P}_\xi \left( \sup_{j,k\in[d]} \sup_{z\in(0,1)} \sqrt{nh} \cdot \Big| \widehat{S}^B_{z|(j,k)}\big(\widehat{\mathbf{\Omega}}_{k\setminus j}(z)\big) - \mathbf{\Omega}_j^T(z)\Big(\widehat{\mathbf{\Sigma}}^B(z)\mathbf{\Omega}_k(z) - \mathbf{e}_k^T\Big)\Big| > n^{-c} \right)$$
$$\le \mathbb{P}_\xi \left( \sup_{j,k\in[d]} \sup_{z\in(0,1)} \big(|I_{12}^B| + |I_2^B|\big) > n^{-c} \right) \le 1/d,$$

following the same proof of Lemma 21. The proof is therefore complete. ∎

### G.3 Properties of Bootstrap Score Statistics

In this section, we focus on establishing certain properties of the Gaussian multiplier bootstrap statistics introduced in this paper. The main goal is to prove Lemma 20, which states the approximation rate of a leading linear term to the bootstrap score statistic. To that end, we establish a rate of convergence for the bootstrap Kendall's tau estimator $\widehat{\tau}^B_{jk}(z)$ parallel to the results for $\widehat{\tau}^B_{jk}(z)$ in Lemma 12.

Recall from (17) and (20) that

$$\widehat{\tau}^B_{jk}(z) = \frac{\sum_{i\neq i'} K_h\left(Z_i - z\right) K_h\left(Z_{i'} - z\right) \operatorname{sign}(X_{ij} - X_{i'j}) \operatorname{sign}(X_{ik} - X_{i'k})(\xi_i + \xi_{i'})}{\sum_{i\neq i'} K_h\left(Z_i - z\right) K_h\left(Z_{i'} - z\right) (\xi_i + \xi_{i'})},$$

and

$$\mathbb{U}_n[\omega^B_z] = \frac{2}{n(n-1)} \sum_{i\neq i'} K_h\left(Z_i - z_0\right) K_h\left(Z_{i'} - z_0\right) (\xi_i + \xi_{i'}).$$

The following lemma presents a convergence rate of the bootstrap Kendall's tau estimator $\widehat{\tau}^B_{jk}(z)$.

**Lemma 23** *Under the conditions of Lemma 20, with probability* $1 - c/d$*,*

$$\mathbb{P}_\xi \left( \max_{j,k\in[d]} \sup_{z\in(0,1)} \Big| [\mathbb{U}_n[\omega^B_z]]\big(\widehat{\tau}^B_{jk}(z) - \tau_{jk}(z)\big) \Big| > C\sqrt{\log(d/h)/(nh^2)} \right) \le 1/d$$

*and*

$$\mathbb{P}_\xi \left( \max_{j,k\in[d]} \sup_{z\in(0,1)} \sqrt{nh} |\Delta W_{z|(j,k)}| > C\sqrt{\log(d/h)/(nh^2)} \right) \le 1/d, \tag{123}$$

*with* $\Delta W_{z|(j,k)}$ *defined in* (116).

**Proof** We first introduce some notation to simplify the proof. Let

$$W_{z|(j,k)}(Y_i) = \frac{2}{n-1} \sum_{i'\neq i} \omega_z(Z_i, Z_{i'}) \Big( \operatorname{sign}(X_{ij} - X_{i'j}) \operatorname{sign}(X_{ik} - X_{i'k}) - \tau_{jk}(z) \Big). \tag{124}$$

From the definition of $\widehat{\tau}_{jk}^{B}(z)$ in (17), conditionally on $\{Y_i\}_{i\in[n]}$, we have

$$\sqrt{n}\cdot \mathbb{U}_n[\omega_{z_0}^B]\Big(\widehat{\tau}_{jk}^{B}(z)-\tau_{jk}(z)\Big)=\frac{1}{\sqrt{n}}\sum_{i=1}^{n}W_{z|(j,k)}(Y_i)\xi_i\sim N\Big(0,\frac{1}{n}\sum_{i=1}^{n}W_{z|(j,k)}^2(Y_i)\Big). \quad (125)$$

Since the bootstrap process in (124) is a Gaussian process, we bound its supreme using the Borell's inequality (see Proposition A.2.1, van der Vaart and Wellner (1996)). The Borell's inequality requires us bound the following three quantities:

1. the variance of $n^{-1}\sum_{i=1}^{n}W_{z|(j,k)}^2(Y_i)$;
2. the supremum norm of $n^{-1}\sum_{i=1}^{n}W_{z|(j,k)}^2(Y_i)$;
3. the $L^2$ norm covering number of the function class

$$\mathcal{F}_W=\Big\{\omega_{z|(j,k)}^{(i)}\mid z\in(0,1),j,k\in[d]\Big\}, \quad (126)$$

under the empirical measure $\mathbb{P}_n=n^{-1}\sum_{i=1}^{n}\delta_{Y_i}$, where

$$\omega_{z|(j,k)}^{(i)}:=\frac{1}{n-1}\sum_{i'\neq i}\omega_z(Z_i,Z_{i'})\,\mathrm{sign}(X_{ij}-X_{i'j})\,\mathrm{sign}(X_{ik}-X_{i'k}).$$

To bound the variance, we first study $W_{z|(j,k)}(Y_i)$ for each single $i\in[n]$. For any bivariate function $f(y_1,y_2)$, define the operator

$$\mathbb{G}_{n-1}^{(i)}[f]=\frac{1}{\sqrt{n-1}}\sum_{i'\neq i}^{n}\left(f(Y_{i'},Y_i)-\mathbb{E}\left[f(Y_{i'},Y_i)\mid Y_i\right]\right).$$

Now,  can be written as

$$\begin{aligned}W_{z|(j,k)}(Y_i)&=\mathbb{E}[W_{z|(j,k)}(Y_i)]+\frac{2}{\sqrt{n-1}}\underbrace{\Big(\mathbb{G}_{n-1}^{(i)}(g_{z|(j,k)})-\tau_{jk}(z)\mathbb{G}_{n-1}^{(i)}(\omega_z)\Big)}_{J^{(1)}(Y_i)}\\&\quad+\underbrace{2\left(g_{z|(j,k)}^{(1)}(Y_i)-\tau_{jk}(z)\omega_z^{(1)}(Z_i)\right)}_{J^{(2)}(Y_i)}.\end{aligned} \quad (127)$$

From (84) and (94), we have that almost surely

$$\max_{i\in[n]}\sup_{j,k\in[d]}\sup_{z\in(0,1)}J^{(2)}(Y_i)\leq Ch^{-1}. \quad (128)$$

Using Lemma 13, we have

$$\sup_{z,j,k}\mathbb{E}[W_{z|(j,k)}(Y_i)]\leq\sup_{z,j,k}2\Big(\left|\mathbb{E}\left[\mathbb{U}_n\left[g_{z|(j,k)}\right]\right]-f_Z^2(z)\tau_{jk}(z)\right|+\left|\mathbb{E}\left[\mathbb{U}_n\left[\omega_z\right]\right]-f_Z^2(z)\right|\Big)\leq Ch^2. \quad (129)$$

Similar to the proof of Lemma 15 and Lemma 17, with probability $1-\delta$, we have

$$\max_{i\in[n]} \sup_{j,k\in[d]} \sup_{z\in(0,1)} \left| \mathbb{G}_{n-1}^{(i)} \left[ \frac{h^{3/2}\big(g_{z|(j,k)} - \tau_{jk}(z)\omega_z\big)}{\sqrt{\log(d/h) \vee \log(n/\delta)}} \right] \right| \le C. \tag{130}$$

Plugging (128), (129) and (130) into (127), with probability $1-1/d$, we have

$$\max_{j,k\in[d]} \sup_{z\in(0,1)} \frac{1}{n}\sum_{i=1}^{n} W_{z|(j,k)}^2(Y_i) \le \max_{i\in[n]} \max_{j,k\in[d]} \sup_{z\in(0,1)} W_{z|(j,k)}^2(Y_i) \le Ch^{-2}, \tag{131}$$

as $\log(d/h)/(nh) = o(1)$ and $h = o(1)$.

Next, we bound the covering number of the function class $\mathcal{F}_W$ defined in (126). For some $M_0$ to be determined later, let $\{K_h(z_\ell - \cdot)\}_{\ell\in[M_0]}$ be the $\epsilon$-net of

$$\mathcal{K} = \left\{ K\left((s-\cdot)/h\right) \mid s \in (0,1) \right\}.$$

That is, for any $z \in (0,1)$, there exists a $z_\ell$ such that $\|K_h(z_\ell - \cdot) - K_h(z-\cdot)\|_{L^2(\mathbb{P}_n)}^2 \le \epsilon$. For this $z_\ell$, we also have

$$\begin{aligned}
\|\omega_{z|(j,k)}^{(i)} - \omega_{z_\ell|(j,k)}^{(i)}\|_{L^2(\mathbb{P}_n)}^2 &\le \frac{1}{n}\sum_{i=1}^{n}\left(\frac{1}{n-1}\sum_{i'\neq i} |\omega_z(Z_i, Z_{i'}) - \omega_{z'}(Z_i, Z_{i'})|\right)^2 \\
&\le \frac{1}{n}\sum_{i=1}^{n}(K_h(z-Z_i))^2\left(\frac{1}{n-1}\sum_{i'\neq i}|K_h(z-Z_{i'}) - K_h(z_\ell - Z_{i'})|^2\right) \\
&\quad + \frac{1}{n}\sum_{i=1}^{n}\big(K_h(z-Z_i) - K_h(z_\ell - Z_i)\big)^2\left(\frac{1}{n-1}\sum_{i'\neq i}|K_h(z_\ell - Z_{i'})|^2\right) \\
&\le C\epsilon^2 h^{-2}.
\end{aligned}$$

Therefore, as $M_0 \le (C/\epsilon)^v$, we have $N(\mathcal{F}_W, \|\cdot\|_{L^2(\mathbb{P}_n)}, h^{-1}\epsilon) \le d^2(C/\epsilon)^{3v}$.

Similarly, for the function class

$$\mathcal{F}_W' = \left\{ W_{z|(j,k)}(Y_i) \mid z \in (0,1), j, d \in [d] \right\}$$

we have

$$N(\mathcal{F}_W', \|\cdot\|_{L^2(\mathbb{P}_n)}, h^{-1}\epsilon) \le N(\mathcal{F}_W, \|\cdot\|_{L^2(\mathbb{P}_n)}, h^{-1}\epsilon) \le d^2(C/\epsilon)^{6v}. \tag{132}$$

This bound follows by combining the the fact that $\tau_{jk}(\cdot)$ is Lipschitz (since $\mathbf{\Sigma}_{jk}(\cdot) \in \mathcal{H}(2,L)$) with Lemma 26 and Lemma 27.

Now, the Dudley's inequality (see Lemma 2.2.8 in van der Vaart and Wellner, 1996), together with the fact that $\mathbb{U}_n[\omega_{z_0}^B](\widehat{\tau}_{jk}^B(z) - \tau_{jk}(z))$ is normally distributed conditionally on data (see (125)), the upper bound and variance bound in (131) and the covering number on $\mathcal{F}_W'$ above, gives us

$$\mathbb{E}\left[\sup_{z\in(0,1)} \max_{j,k\in[d]} \mathbb{U}_n[\omega_{z_0}^B]\Big(\widehat{\tau}_{jk}^B(z) - \tau_{jk}(z)\Big)\right] \le C\sqrt{\frac{\log(d/h)}{nh^2}}.$$

Using the Borell's inequality, on the event that (131) is true, we have

$$\mathbb{P}_\xi\Big(\sup_{z\in(0,1)}\max_{j,k\in[d]}\mathbb{U}_n[\omega^B_{z_0}]\Big(\widehat{\tau}^B_{jk}(z)-\tau_{jk}(z)\Big)\geq C\sqrt{\frac{\log(d/h)}{nh^2}}\Big)\geq 1/d.$$

Since (131) is true with probability $1-1/d$, the first part of the lemma is proved.

Similarly, we can bound $\Delta W_{z|(j,k)}$. By (127), we have

$$\Delta W_{z|(j,k)}=\bar{\xi}\cdot\mathbb{E}[W_{z|(j,k)}(Y)]+\frac{1}{n}\sum_{i=1}^{n}\frac{2J^{(1)}(Y_i)}{\sqrt{n-1}}\xi_i, \tag{133}$$

where $\bar{\xi}=n^{-1}\sum_{i=1}^n\xi_i$. From the concentration of sub-Gaussian random variables, we have $\mathbb{P}(|\bar{\xi}|<C\sqrt{\log d/n})\geq 1-1/d$. Combining with (129), we have

$$\mathbb{P}_\xi\left(\sup_{z\in(0,1)}\max_{j,k\in[d]}\bar{\xi}\cdot\mathbb{E}[W_{z|(j,k)}(Y_i)]\geq\sqrt{\frac{Ch^4\log d}{n}}\right)\leq 1/d. \tag{134}$$

According to (130), with probability $1-1/d$, we have

$$\sup_{z\in(0,1)}\max_{j,k\in[d]}\frac{1}{n}\sum_{i=1}^{n}\left(\frac{2J^{(1)}(Y_i)}{\sqrt{n-1}}\right)^2\leq\max_{i\in[n]}\sup_{z\in(0,1)}\max_{j,k\in[d]}\left(\frac{2J^{(1)}(Y_i)}{\sqrt{n-1}}\right)^2\leq\frac{C\log(d/h)}{nh^3}. \tag{135}$$

Define the function class $\widetilde{F}_W=\{J^{(1)}(\cdot)\mid z\in(0,1),j,k\in[d]\}$. By the definition of $J^{(1)}$ in (127), we apply Lemma 26 to the covering number of function classes consisting of $W_{z|(j,k)}$, $g^{(1)}_{z|(j,k)}$ and $\omega^{(1)}_z$ in (132), (144) and (153) to obtain

$$N(\widetilde{\mathcal{F}}_W,\|\cdot\|_{L^2(\mathbb{P}_n)},h^{-1}\epsilon)\leq d^2(C/\epsilon)^{cv}.$$

The Borell's inequality, on the event that (135) is true, gives us

$$\mathbb{P}_\xi\left(\max_{j,k\in[d]}\sup_{z\in(0,1)}\frac{1}{n}\sum_{i=1}^{n}J^{(1)}(Y_i)\xi_i\geq C\sqrt{\frac{\log(d/h)}{n^2h^3}}\right)\leq 1/d. \tag{136}$$

Plugging (134) and (136) into (133), the proof of the second part is complete. ■

The following lemma presents a convergence rate of $\mathbb{U}^B(\omega_z)$ to $f^2_Z(z)$.

**Lemma 24** *Under the conditions of Lemma 20, with probability* $1-1/n$*,*

$$\mathbb{P}_\xi\left(\sup_{z\in(0,1)}\big|\mathbb{U}_n[\omega^B_z]]-f^2_Z(z)\big|>C\sqrt{\log(1/h)/(nh^2)}\right)\leq 1/n.$$

**Proof** The proof is similar to that of Lemma 23. We define

$$\overline{W}_{z|(j,k)}(Z_i)=\frac{2}{n-1}\sum_{i'\neq i}\Big(\omega_z(Z_i,Z_{i'})-f^2_Z(z)\Big).$$

Conditionally on the data $\{Y_i\}_{i\in[n]}$,

$$\sqrt{n}\cdot(\mathbb{U}_n[\omega_z^B]-f_Z^2(z))=\frac{1}{\sqrt{n}}\sum_{i=1}^n\overline{W}_{z|(j,k)}(Z_i)\xi_i\sim N\Big(0,\frac{1}{n}\sum_{i=1}^n\overline{W}^2_{z|(j,k)}(Z_i)\Big).$$

Note that

$$\overline{W}_{z|(j,k)}(Y_i)=\mathbb{E}[\overline{W}_{z|(j,k)}(Y_i)]+2(n-1)^{-1/2}\mathbb{G}_{n-1}^{(i)}[\omega_z]+2\omega_z^{(1)}(Z_i).$$

Similar to the proof of Lemma 17 and (130), with probability $1-\delta$,

$$\max_{j,k\in[d]}\sup_{z\in(0,1)}\frac{1}{n}\sum_{i=1}^n\overline{W}^2_{z|(j,k)}(Y_i)\leq\max_{i\in[n]}\max_{j,k\in[d]}\sup_{z\in(0,1)}\overline{W}^2_{z|(j,k)}(Y_i)\leq Ch^{-2}. \tag{137}$$

Using Lemmas 29, 26 and 27, we can bound the covering number of the function class

$$\mathcal{F}''_W=\{\overline{W}_{z|(j,k)}(Y_i)|z\in(0,1)\}$$

by $N(\mathcal{F}''_W,\|\cdot\|_{L^2(\mathbb{P}_n)},h^{-1}\epsilon)\leq(C/\epsilon)^{6v}$.

The remained of the proof follows the proof of Lemma 23. Using the Dudley's and Borell's inequality (see Lemma 2.2.8 and Proposition A.2.1 van der Vaart and Wellner, 1996)), on the event that (137) is true, we have

$$\mathbb{P}_\xi\left(\sup_{z\in(0,1)}\big(\mathbb{U}_n[\omega_z^B]-f_Z^2(z)\big)\geq C\sqrt{\frac{\log(n/h)}{nh^2}}\right)\geq 1/n.$$

The lemma follows since (137) holds with probability $1-1/n$. ■

### G.4 Proof of Lemma 2

Recall that we defined the matrix $\mathbf{\Theta}^{(i)}$ in (57) with elements

$$\mathbf{\Theta}^{(i)}_{jk}(z)=\pi\cos\big((\pi/2)\tau_{jk}(z)\big)\cdot\frac{1}{n-1}\sum_{i'\neq i}\tau^{(1)}_{jk}(Y_{i'}),$$

and $\tau^{(1)}_{jk}$ defined in (12). The strategy of the proof is to establish

$$\frac{1}{n}\sum_{i=1}^n\Big(\widehat{\mathbf{\Omega}}_j^T(z_0)\widehat{\mathbf{\Theta}}^{(i)}(z_0)\widehat{\mathbf{\Omega}}_k(z_0)\Big)^2\xrightarrow{P}\mathrm{Var}(\mathbf{\Omega}_j^T(z_0)\mathbf{\Theta}_{z_0}\mathbf{\Omega}_k(z_0))\quad\text{and} \tag{138}$$

$$[\mathbb{U}_n[\omega_{z_0}]]^2\xrightarrow{P}f_Z^4(z_0). \tag{139}$$

The lemma then follows from the Slutsky's theorem.

We first establish (138). Let

$$\Delta_1=\frac{1}{n}\sum_{i=1}^n\Big(\widehat{\mathbf{\Omega}}_j^T(z_0)\widehat{\mathbf{\Theta}}^{(i)}(z_0)\widehat{\mathbf{\Omega}}_k(z_0)\Big)^2-\frac{1}{n}\sum_{i=1}^n\Big(\mathbf{\Omega}_j^T(z_0)\widehat{\mathbf{\Theta}}^{(i)}(z_0)\widehat{\mathbf{\Omega}}_k(z_0)\Big)^2\text{ and}$$

$$\Delta_2=\frac{1}{n}\sum_{i=1}^n\Big(\mathbf{\Omega}_j^T(z_0)\widehat{\mathbf{\Theta}}^{(i)}(z_0)\widehat{\mathbf{\Omega}}_k(z_0)\Big)^2-\frac{1}{n}\sum_{i=1}^n\Big(\mathbf{\Omega}_j^T(z_0)\widehat{\mathbf{\Theta}}^{(i)}(z_0)\mathbf{\Omega}_k(z_0)\Big)^2.$$

We can bound $\Delta_1$ as

$$
\begin{aligned}
|\Delta_1| &= \left| (\widehat{\boldsymbol{\Omega}}_j(z_0) - \boldsymbol{\Omega}_j(z_0))^T \cdot \frac{1}{n}\sum_{i=1}^n \widehat{\boldsymbol{\Theta}}^{(i)}(z_0)\widehat{\boldsymbol{\Omega}}_k(z_0)\widehat{\boldsymbol{\Omega}}_k^T(z_0)\widehat{\boldsymbol{\Theta}}^{(i)}(z_0) \cdot (\widehat{\boldsymbol{\Omega}}_j(z_0) + \boldsymbol{\Omega}_j(z_0))^T \right| \\
&\le \big\|\widehat{\boldsymbol{\Omega}}_j(z_0) - \boldsymbol{\Omega}_j(z_0)\big\|_1 \big\|\widehat{\boldsymbol{\Omega}}_j(z_0) + \boldsymbol{\Omega}_j(z_0)\big\|_1 \big\|\widehat{\boldsymbol{\Omega}}_k(z_0)\big\|_1^2 \big\|\widehat{\boldsymbol{\Theta}}^{(i)}(z_0)\big\|_{\max}^2 \\
&= O_P\big(M^4 r_{2n}(2\pi)^2 h^{-1}\big) = o_P(1),
\end{aligned} \tag{140}
$$

where the second equality is by Assumption 4.4 and $\big\|\widehat{\boldsymbol{\Theta}}^{(i)}(z_0)\big\|_{\max} \le 2\pi h^{-1/2}$ with probability $1 - c/d$. Similarly,

$$
|\Delta_2| = O_P\big(M^4 r_{2n}(2\pi)^2 h^{-1}\big) = o_P(1). \tag{141}
$$

Next, $\max_{i\in[n]} h^{1/2}\|\widehat{\boldsymbol{\Theta}}^{(i)} - \boldsymbol{\Theta}^{(i)}\|_{\max} \le \Delta_{31} + \Delta_{32}$, where

$$
\Delta_{31} = \max_{i\in[n],j,k\in[d]} \pi \left|\cos\left((\pi/2)\widehat{\tau}_{jk}(z_0)\right) - \cos\left((\pi/2)\tau_{jk}(z_0)\right)\right| \cdot \left| \frac{h^{1/2}}{n-1}\sum_{i'\neq i} \tau_{jk}^{(1)}(Y_{i'}) \right|;
$$

$$
\Delta_{32} = \max_{i\in[n],j,k\in[d]} \pi \left|\cos\left((\pi/2)\tau_{jk}(z_0)\right)\right| \cdot h^{1/2} \left| q_{i,jk}(z_0) - \frac{1}{n-1}\sum_{i'\neq i} \tau_{jk}^{(1)}(Y_{i'}) \right|,
$$

where $q_{i,jk}$ is defined in (13). Using Lemma 9 and Lemma 10, with probability $1 - c/d$

$$
\Delta_{31} \le \pi^2 \|K\|_\infty \cdot \max_{j,k} |\widehat{\tau}_{jk}(z_0) - \widehat{\tau}_{jk}(z_0)| \lesssim \left(h^2 + \sqrt{\log(dn)/(nh)}\right).
$$

Let $\bar{q}_{i,jk} = q_{i,jk} + \widehat{\tau}_{jk}(z_0)$. By the Hoeffding's inequality and union bound, we have

$$
\mathbb{P}\left( \max_{i,j,k} h^{3/2}|\bar{q}_{i,jk} - \mathbb{E}[\bar{q}_{i,jk}]| > t \,\middle|\, Y_i \right) \le 2nd^2 \exp\left(-(n-1)t^2/2\right).
$$

Setting $t = \sqrt{\log(dn)/n}$ and taking the expectation above, with probability $1 - c/d$,

$$
\max_{i,j,k} h^{1/2}|\bar{q}_{i,jk} - \mathbb{E}[\bar{q}_{i,jk}]| \le \sqrt{\log(dn)/n}.
$$

Similarly, with probability $1 - c/d$,

$$
\Delta_{32} \le \pi h^{1/2} \max_{j,k} |\widehat{\tau}_{jk}(z_0) - \widehat{\tau}_{jk}(z_0)| + \max_{i,j,k} h^{1/2}|\bar{q}_{i,jk} - \mathbb{E}[\bar{q}_{i,jk}]| \lesssim \sqrt{\log(dn)/n} + h^{3/2}.
$$

Therefore,

$$
\begin{aligned}
\Delta_3 &:= \frac{1}{n}\sum_{i=1}^n \left(\boldsymbol{\Omega}_j^T(z_0)\widehat{\boldsymbol{\Theta}}^{(i)}(z_0)\boldsymbol{\Omega}_k(z_0)\right)^2 - \frac{1}{n}\sum_{i=1}^n \left(\boldsymbol{\Omega}_j^T(z_0)\boldsymbol{\Theta}^{(i)}(z_0)\boldsymbol{\Omega}_k(z_0)\right)^2 \\
&\le 4M^4\pi h^{-1/2} \max_{i\in[n]} \|\widehat{\boldsymbol{\Theta}}^{(i)} - \boldsymbol{\Theta}^{(i)}\|_{\max} = O_P\Big(h + \sqrt{\log(dn)/(nh^3)}\Big) = o_P(1).
\end{aligned} \tag{142}
$$

Finally, by the law of large numbers,

$$\frac{1}{n}\sum_{i=1}^{n}\Big(\mathbf{\Omega}_j^T(z_0)\mathbf{\Theta}^{(i)}(z_0)\mathbf{\Omega}_k(z_0)\Big)^2 \xrightarrow{P} \mathbb{E}[\mathbf{\Omega}_j^T(z_0)\mathbf{\Theta}^{(i)}(z_0)\mathbf{\Omega}_k(z_0)]^2 = \mathrm{Var}(\mathbf{\Omega}_j^T(z_0)\mathbf{\Theta}_{z_0}\mathbf{\Omega}_k(z_0)).$$

Combining (140), (141) and (142), we prove (138).

Using Lemma 24 and the continuous mapping theorem, we also prove (139). By the Slutsky's theorem, we have $\widehat{\sigma}_{jk}(z_0) \xrightarrow{P} \sigma_{jk}(z_0)$.

## Appendix H. Results on Covering Number

In this section, we present several results on the covering number of certain function classes. The first two lemmas, Lemmas 26 and 27, are preliminary technical lemmas that will be used to prove Lemmas 28, 29 and 30. Lemma 26 provides bounds on the covering numbers for function classes generated from products and additions of two function classes. Lemma 27 provides a bound on the covering number of a class of constant functions.

Before presenting these lemmas, we first state a result on the covering number of kernel functions.

**Lemma 25** *(Lemma 22 of Nolan and Pollard, 1987). Let $K : \mathbb{R} \mapsto \mathbb{R}$ be a bounded variation function. The following function class*

$$\mathcal{K} = \left\{ K\left(\frac{s-\cdot}{h}\right) \,\Big|\, h > 0, s \in \mathbb{R} \right\}$$

*indexed by the kernel bandwidth satisfies the uniform entropy condition*

$$\sup_Q N(\mathcal{K}, L_2(Q), \epsilon) \le C\epsilon^{-v}, \quad \text{for all } \epsilon \in (0,1), \tag{143}$$

*for some $C > 0$ and $v > 0$.*

The following lemma is about the covering number of summation and product of functions.

**Lemma 26** *Let $\mathcal{F}_1$ and $\mathcal{F}_2$ be two function classes satisfying*

$$N(\mathcal{F}_1, \|\cdot\|_{L_2(Q)}, a_1\epsilon) \le C_1\epsilon^{-v_1} \qquad \text{and} \qquad N(\mathcal{F}_2, \|\cdot\|_{L_2(Q)}, a_2\epsilon) \le C_2\epsilon^{-v_2}$$

*for some $C_1, C_2, a_1, a_2, v_1, v_2 > 0$ and any $0 < \epsilon < 1$. Define $\|\mathcal{F}_\ell\|_\infty = \sup\{\|f\|_\infty, f \in \mathcal{F}_\ell\}$ for $\ell = 1, 2$ and $U = \|\mathcal{F}_1\|_\infty \vee \|\mathcal{F}_2\|_\infty$. For the function classes $\mathcal{F}_\times = \{f_1 f_2 \mid f_1 \in \mathcal{F}_1, f_2 \in \mathcal{F}_2\}$ and $\mathcal{F}_+ = \{f_1 + f_2 \mid f_1 \in \mathcal{F}_1, f_2 \in \mathcal{F}_2\}$, we have for any $\epsilon \in (0,1)$,*

$$N(\mathcal{F}_\times, \|\cdot\|_{L_2(Q)}, \epsilon) \le C_1 C_2 \left(\frac{2a_1 U}{\epsilon}\right)^{v_1}\left(\frac{2a_2 U}{\epsilon}\right)^{v_2};$$

$$N(\mathcal{F}_+, \|\cdot\|_{L_2(Q)}, \epsilon) \le C_1 C_2 \left(\frac{2a_1}{\epsilon}\right)^{v_1}\left(\frac{2a_2}{\epsilon}\right)^{v_2}.$$

**Proof** For any $\epsilon \in (0,1)$, let $N_1 = \{f_{11}, \dots, f_{1N_1}\}$ and $N_2 = \{f_{21}, \dots, f_{2N_2}\}$ be the $\epsilon/(2U)$-net of $\mathcal{F}_1$ and $\mathcal{F}_2$ respectively with

$$N_1 \le C_1 \left(\frac{2a_1U}{\epsilon}\right)^{v_1} \text{ and } N_2 \le C_2 \left(\frac{2a_2U}{\epsilon}\right)^{v_2}.$$

Define the set $N = \{f_{1j}f_{2k} \mid f_{1j} \in N_1, f_{2k} \in N_2\}$. We now show that $N$ is an $\epsilon$-net for $\mathcal{F}_\times$. For any $f_1 f_2 \in \mathcal{F}$, there exist two functions $f_{1j} \in N_1$ and $f_{2k} \in N_2$ such that $\|f_1 - f_{1j}\|_{\|\cdot\|_{L_2(Q)}} \le \epsilon/(2U)$ and $\|f_2 - f_{2k}\|_{\|\cdot\|_{L_2(Q)}} \le \epsilon/(2U)$. Moreover, we have $f_{1j}f_{2k} \in N$ and

$$\|f_1 f_2 - f_{1j}f_{2k}\|_{\|\cdot\|_{L_2(Q)}} \le \|\mathcal{F}_2\|_\infty \|f_1 - f_{1j}\|_{\|\cdot\|_{L_2(Q)}} + \|\mathcal{F}_1\|_\infty \|f_2 - f_{2k}\|_{\|\cdot\|_{L_2(Q)}} \le \epsilon.$$

Therefore $N$ is the $\epsilon$-net for $\mathcal{F}_\times$. Similarly, we also have

$$\|(f_1 + f_2) - (f_{1j} + f_{2k})\|_{\|\cdot\|_{L_2(Q)}} \le \|f_1 - f_{1j}\|_{\|\cdot\|_{L_2(Q)}} + \|f_2 - f_{2k}\|_{\|\cdot\|_{L_2(Q)}} \le \epsilon/U.$$

So $N' = \{f_{1j} + f_{2k} \mid f_{1j} \in N_1, f_{2k} \in N_2\}$ is the $\epsilon/U$-net of $\mathcal{F}_+$. We finally complete the proof by showing that

$$|N'| = |N| = N_1 N_2 \le C_1 C_2 \left(\frac{2a_1U}{\epsilon}\right)^{v_1} \left(\frac{2a_2U}{\epsilon}\right)^{v_2}.$$

■

**Lemma 27** *Let $f(s)$ be a Lipschitz function defined on $[a,b]$ such that $|f(s) - f(s')| \le L_f|s - s'|$ for any $s, s' \in [a,b]$. We define the constant function class $\mathcal{F}_c = \{g_s(\cdot) \equiv f(s) \mid s \in [a,b]\}$. For any probability measure $Q$, the covering number of $\mathcal{F}_c$ satisfies for any $\epsilon \in (0,1)$,*

$$N(\mathcal{F}_c, \|\cdot\|_{L_2(Q)}, \epsilon) \le L_f \cdot \frac{|b-a|}{\epsilon}.$$

**Proof** Let $N = \{a + i\epsilon/L_f \mid i = 0, \dots, \lfloor L_f|b-a|/\epsilon \rfloor\}$. For any $g_{s_0} \in \mathcal{F}_c$, there exists a $s \in N$ such that $|s_0 - s| \le \epsilon/L_f$ and we have

$$\|g_{s_0} - g_s\|_{L^2(Q)} = |f(s_0) - f(s)| \le L_f|s_0 - s| \le \epsilon.$$

Therefore $\{g_s \mid s \in N\}$ is the $\epsilon$-net of $\mathcal{F}_c$. As $|N| \le L_f|b-a|/\epsilon$, the lemma is proved. ■

The following lemma presents the covering number of function classes consisting of $g^{(1)}_{z|(j,k)}(\cdot)$ or $g^{(2)}_{z|(j,k)}(\cdot)$ defined in (41) and (42).

**Lemma 28** *For some $0 < \underline{\mathrm{h}} < \overline{\mathrm{h}} < 1$, we consider the class of functions*

$$\mathcal{F}^{(1)} = \left\{\sqrt{h} \cdot g^{(1)}_{z|(j,k)} \,\middle|\, h \in [\underline{\mathrm{h}}, \overline{\mathrm{h}}], z \in (0,1), j,k \in [d]\right\}; \tag{144}$$
$$\mathcal{F}^{(2)} = \left\{h \cdot g^{(2)}_{z|(j,k)} \,\middle|\, h \in [\underline{\mathrm{h}}, \overline{\mathrm{h}}], z \in (0,1), j,k \in [d]\right\},$$

*where* $g^{(1)}_{z|(j,k)}$ *and* $g^{(2)}_{z|(j,k)}$ *are defined in* (41) *and* (42). *There exist constants* $C_{(1)}$ *and* $C_{(2)}$ *such that for any* $\epsilon \in (0,1)$

$$\sup_Q N(\mathcal{F}^{(1)}, \|\cdot\|_{L_2(Q)}, \epsilon) \leq \frac{d^2 C_{(1)}}{\underline{\mathrm{h}}^{v+9}\epsilon^{v+6}} \quad \text{and} \quad \sup_Q N(\mathcal{F}^{(2)}, \|\cdot\|_{L_2(Q)}, \epsilon) \leq \frac{d^2 C_{(2)}}{\underline{\mathrm{h}}^{2v+18}\epsilon^{4v+15}}.$$

**Proof** Recall that

$$g^{(1)}_{z|(j,k)}(y) = \mathbb{E}[g_{z|(j,k)}(y, Y)] - \mathbb{E}\left[\mathbb{U}_n(g_{z|(j,k)})\right] \qquad \text{and}$$
$$g^{(2)}_{z|(j,k)}(y_1, y_2) = g_{z|(j,k)}(y_1, y_2) - g^{(1)}_{z|(j,k)}(y_1) - g^{(1)}_{z|(j,k)}(y_2) - \mathbb{E}\left[\mathbb{U}_n(g_{z|(j,k)})\right].$$

Our proof strategy for bounding the covering numbers of $\mathcal{F}^{(1)}$ and $\mathcal{F}^{(2)}$ is to decompose them into the following three auxiliary function classes:

$$\mathcal{F}^{(1)}_{1,jk} = \left\{\mathbb{E}[g_{z|(j,k)}(y, Y)] \mid h \in [\underline{\mathrm{h}}, \overline{\mathrm{h}}], z \in (0,1)\right\};$$
$$\mathcal{F}^{(1)}_{2} = \{\mathbb{E}\left[\mathbb{U}_n(g_{h,z})\right] \mid h \in [\underline{\mathrm{h}}, \overline{\mathrm{h}}], z \in (0,1)\};$$
$$\mathcal{F}^{(2)}_{jk} = \{g_{z|(j,k)}(y_1, y_2) \mid h \in [\underline{\mathrm{h}}, \overline{\mathrm{h}}], z \in (0,1)\}.$$

Observe that we can write

$$\mathcal{F}^{(1)} = \{\sqrt{h}\cdot(f_1 - f_2) \mid h \in [\underline{\mathrm{h}}, \overline{\mathrm{h}}], f_1 \in \mathcal{F}^{(1)}_{1,jk}, f_2 \in \mathcal{F}^{(1)}_{2}, j,k \in [d]\};$$
$$\mathcal{F}^{(2)} = \{h\cdot(f_1 - f_2 - f_3 - f_4) \mid h \in [\underline{\mathrm{h}}, \overline{\mathrm{h}}], f_1 \in \mathcal{F}^{(2)}_{jk}, f_2, f_3 \in \mathcal{F}^{(1)}_{1,jk}, f_4 \in \mathcal{F}^{(1)}_{2}, j,k \in [d]\}.$$

Therefore, we can apply Lemma 26 on the addition and product of functions classes and Lemma 27 on the constant functions to bound the covering numbers of $\mathcal{F}^{(1)}$ and $\mathcal{F}^{(2)}$.

**Covering number of $\mathcal{F}^{(1)}_{1,jk}$.** We bound the covering number of $\mathcal{F}^{(1)}_{1,jk}$ first. Recall from (62) that for $y' = (z', \boldsymbol{x}')$

$$\mathbb{E}\left[g_{z|(j,k)}(y', Y)\right] = K_h(z' - z)\int K_h(s - z)\varphi\left(x'_j, x'_k, \boldsymbol{\Sigma}_{jk}(s)\right) f_Z(s)ds.$$

We have $\mathcal{F}^{(1)}_{1,jk} = \left\{f_1 \cdot f_2 \mid f_1 \in \{K_h(\cdot - z)\}, f_2 \in \mathcal{F}^{(1)}_{3,jk}\right\}$ where

$$\mathcal{F}^{(1)}_{3,jk} = \left\{q_{z,h}(x,y) = \int K_h(s-z)\varphi\left(x, y, \boldsymbol{\Sigma}_{jk}(s)\right) f_Z(s)ds, h \in [\underline{\mathrm{h}}, \overline{\mathrm{h}}], z \in (0,1)\right\}. \tag{145}$$

Let $\bar{\varphi}_{x,y}(s) = \varphi\left(x, y, \boldsymbol{\Sigma}_{jk}(s)\right) f_Z(s)$. Then $\mathcal{F}^{(1)}_{3,jk}$ is the class of functions generated by the convolution $q_{z,h}(x,y) = (K_h * \bar{\varphi}_{x,y})(z)$. The $L_1$ norm of the derivative of $K_h$ can be bounded by

$$\|K'_h\|_1 = \int \frac{1}{h^2}\left|K'\left(\frac{t}{h}\right)\right| dt = h^{-1}\int |K'(t)|dt = h^{-1}\mathrm{TV}(K), \tag{146}$$

where $\mathrm{TV}(K)$ is the total variation of the kernel $K$. Similarly we have for any $h \in [\underline{\mathrm{h}}, \overline{\mathrm{h}}]$,

$$\left\|\frac{\partial}{\partial h}K_h\right\|_1 \leq \int h^{-2}|K(t/h)|dt + \int h^{-3}|K'(t/h)|dt = h^{-1}\|K\|_1 + h^{-2}\mathrm{TV}(K).$$

We can apply a similar argument as in (85) and derive that

$$\sup_{z_0,h,x,y}\left|\frac{\partial}{\partial z}q_{z,h}(x,y)\Big|_{z=z_0}\right| = \sup_{h,x,y}\|K_h'*\bar{\varphi}_{x,y}\|_\infty \le \sup_{h,x,y}\|K_h'\|_1\|\bar{\varphi}_{x,y}\|_\infty \le 2\underline{\mathrm{h}}^{-1}\mathrm{TV}(K)\bar{\mathrm{f}}_Z, \tag{147}$$

where the first equality is due to the property of the derivative of a convolution, the first inequality is because of Young's inequality and the last inequality is by (146). Similarly, we have

$$\begin{aligned}\sup_{z,h_0,x,y}\left|\frac{\partial}{\partial h}q_{z,h}(x,y)\Big|_{h=h_0}\right| &= \sup_{h_0,x,y}\|\nabla_{h=h_0}K_h*\bar{\varphi}_{x,y}\|_\infty \\ &\le \sup_{h_0,x,y}\|\nabla_{h=h_0}K_h\|_1\|\bar{\varphi}_{x,y}\|_\infty \\ &\le 2\bar{\mathrm{f}}_Z(\underline{\mathrm{h}}^{-1}\|K\|_1+\underline{\mathrm{h}}^{-2}\mathrm{TV}(K)).\end{aligned} \tag{148}$$

Therefore for any $z_1,z_2\in(0,1)$, $h_1,h_2\in[\underline{\mathrm{h}},\bar{\mathrm{h}}]$, denoting $C_h:=2\bar{\mathrm{f}}_Z[(\underline{\mathrm{h}}^{-1}+\underline{\mathrm{h}}^{-2})\mathrm{TV}(K)+\underline{\mathrm{h}}^{-1}\|K\|_1]$, we have

$$\sup_{x,y}|q_{z_1,h_1}(x,y)-q_{z_2,h_2}(x,y)|\le C_h\max(|z_1-z_2|,|h_1-h_2|).$$

Given any measure $Q$ on $\mathbb{R}^2$, let $\mathcal{Z}$ be the $\epsilon/C_h$-net of $(0,1)\times[\underline{\mathrm{h}},\bar{\mathrm{h}}]$ under $\|\cdot\|_\infty$. For any $z\in(0,1),h\in[\underline{\mathrm{h}},\bar{\mathrm{h}}]$, choose $(z_0,h_0)\in\mathcal{Z}$ such that $\max(|z-z_0|,|h-h_0|)\le\epsilon/C_h$ and we have

$$\|q_{z,h}-q_{z_0,h_0}\|_{\|\cdot\|_{L_2(Q)}}\le\|q_{z,h}-q_{z_0,h_0}\|_\infty\le\epsilon.$$

This shows that $\{q_{z,h}\mid(z,h)\in\mathcal{Z}\}$ is the $\epsilon$-net of $\mathcal{F}_{3,jk}^{(1)}$ and

$$\sup_Q N(\mathcal{F}_{3,jk}^{(1)},\|\cdot\|_{L_2(Q)},\epsilon)\le|\mathcal{Z}|\le\left(\frac{C_h}{\epsilon}\right)^2. \tag{149}$$

According to the formulation in (62) and Lemma 26 and (143), we have

$$\sup_Q N(\mathcal{F}_{1,jk}^{(1)},\|\cdot\|_{L_2(Q)},\epsilon)\le C\left(\frac{1}{\underline{\mathrm{h}}\epsilon}\right)^v\left(\frac{\bar{\mathrm{h}}-\underline{\mathrm{h}}}{\underline{\mathrm{h}}^2\epsilon}\right)\left(\frac{C_h}{\underline{\mathrm{h}}\epsilon}\right)^2\le\frac{CC_h^2}{\underline{\mathrm{h}}^{v+4}\epsilon^{v+3}}. \tag{150}$$

**Covering number of $\mathcal{F}_2^{(1)}$.** According to (51) and Assumptions (**T**) and (**D**), we have that for any $z_1,z_2\in(0,1),h_1,h_2\in[\underline{\mathrm{h}},\bar{h}]$

$$\begin{aligned}\left|\mathbb{E}\left[\mathbb{U}_n(g_{z_1|(j,k)}^{(1)})\right]-\mathbb{E}\left[\mathbb{U}_n(g_{z_2|(j,k)}^{(1)})\right]\right| &\le\left|f_Z^2(z_1)\tau_{jk}(z_1)-f_Z^2(z_2)\tau_{jk}(z_2)\right|+C|h_1^2-h_2^2| \\ &\le\bar{\mathrm{f}}_Z^2C_\tau|z_1-z_2|+2C\bar{\mathrm{h}}|h_1-h_2|.\end{aligned}$$

Using Lemma 27, we have

$$\sup_Q N(\mathcal{F}_2^{(1)},\|\cdot\|_{L_2(Q)},\epsilon)\le\left(\frac{\bar{\mathrm{f}}_Z^2C_\tau+2C\bar{\mathrm{h}}}{\epsilon}\right)^2. \tag{151}$$

**Covering number of $\mathcal{F}^{(1)}$.** Observe that the function $g(h) = \sqrt{h}$ is Lipschitz on $[\underline{\mathrm{h}}, \overline{\mathrm{h}}]$ with Lipschitz constant $(\underline{\mathrm{h}})^{-1/2}/2$. Combining (150) and (151) with Lemma 26, we have

$$\sup_Q N(\mathcal{F}^{(1)}, \|\cdot\|_{L_2(Q)}, \epsilon) \le d^2 \cdot \frac{2CC_h^2(\overline{\mathrm{f}}_Z^2 C_\tau + 2C\overline{\mathrm{h}})^2}{\underline{\mathrm{h}}^{v+5}\epsilon^{v+6}}. \tag{152}$$

Here the additional $d^2$ on the right hand side of (152) is because we also take supreme over $j,k \in [d]$ in the definition of $\mathcal{F}^{(1)}$. As $C_h \le 2\overline{\mathrm{f}}_Z(2\mathrm{TV}(K) + \|K\|_1) \cdot \underline{\mathrm{h}}^{-2}$, defining

$$C_{(1)} := 4C\overline{\mathrm{f}}_Z(2\mathrm{TV}(K) + \|K\|_1)(\overline{\mathrm{f}}_Z^2 C_\tau + 2C)^2$$

and the first part of lemma in (144) is proved.

**Covering numbers of $\mathcal{F}_{jk}^{(2)}$ and $\mathcal{F}^{(2)}$.** Let $N_K = \{K_h(x-z)K_h(y-z) \mid z \in \mathcal{Z}_K, h \in \mathcal{H}_K\}$ be the $\epsilon$-net of the function class $\mathcal{K}^2 = \{K_h(x-z)K_h(y-z) \mid h \in [\underline{\mathrm{h}}, \overline{\mathrm{h}}], z \in (0,1)\}$. According to Lemma 26 and (143), we have $|N_K| \le C^2(2\|K\|_\infty/\epsilon)^{2v}$. Given any $g_{h_0,z_0} \in \mathcal{F}_{jk}^{(2)}$, there exist $h_1 \in \mathcal{Z}_K, z_1 \in \mathcal{H}_K$ such that

$$\|g_{h_0,z_0} - g_{h_1,z_1}\|_{\|\cdot\|_{L_2(Q)}} \le \|K_{h_0}(\cdot - z_0)K_{h_0}(\cdot - z_0) - K_{h_1}(\cdot - z_1)K_{h_1}(\cdot - z_1)\|_{\|\cdot\|_{L_2(Q)}} \le \epsilon.$$

Therefore $N_g = \{g_{z|(j,k)}(y_1, y_2) \mid z \in \mathcal{Z}_K, h \in \mathcal{H}_K\}$ is an $\epsilon$-net of $\mathcal{H}$ and $|N_g| = |N_K| \le C^2(2\|K\|_\infty/\epsilon)^{2v}$. Applying Lemma 26 again with (151) and (152), we have

$$\begin{aligned}\sup_Q N(\mathcal{F}^{(2)}, \|\cdot\|_{L_2(Q)}, \epsilon) &\le d^2C^2\left(\frac{\overline{\mathrm{h}} - \underline{\mathrm{h}}}{\epsilon}\right)\left(\frac{2\|K\|_\infty}{\epsilon}\right)^{2v}\left(\frac{C_{(1)}}{\underline{\mathrm{h}}^{v+9}\epsilon^{v+6}}\right)^2\left(\frac{\overline{\mathrm{f}}_Z^2 C_\tau + 2C\overline{\mathrm{h}}}{\epsilon}\right)^2 \\ &\le \frac{d^2 C_{(2)}}{\underline{\mathrm{h}}^{2v+18}\epsilon^{4v+15}},\end{aligned}$$

where $C_{(2)} := C^2 4^v \|K\|_\infty^{2v} C_{(1)}^2 (\overline{\mathrm{f}}_Z^2 C_\tau + 2C)^2$. Therefore, we complete the proof of the lemma. ■

Similar to Lemma 28, we can also establish the covering number for function classes consisting of $\omega_z^{(1)}$ or $\omega_z^{(2)}$ in the following lemma.

**Lemma 29** *For some $0 < \underline{\mathrm{h}} < \overline{\mathrm{h}} < 1$, we consider the class of functions*

$$\begin{aligned}\mathcal{K}^{(1)} &= \left\{\sqrt{h} \cdot \omega_z^{(1)} \,\middle|\, h \in [\underline{\mathrm{h}}, \overline{\mathrm{h}}], z \in (0,1)\right\}; \\ \mathcal{K}^{(2)} &= \left\{h \cdot \omega_z^{(2)} \,\middle|\, h \in [\underline{\mathrm{h}}, \overline{\mathrm{h}}], z \in (0,1)\right\},\end{aligned} \tag{153}$$

*where $\omega_z^{(1)}$ and $\omega_z^{(2)}$ are defined in* (44) *and* (45)*. There exist constants $C'_{(1)}$ and $C'_{(2)}$ such that for any $\epsilon \in (0,1)$*

$$\sup_Q N(\mathcal{K}^{(1)}, \|\cdot\|_{L_2(Q)}, \epsilon) \le \frac{C'_{(1)}}{\underline{\mathrm{h}}^{2v+7}\epsilon^{v+3}} \quad \text{and} \quad \sup_Q N(\mathcal{K}^{(2)}, \|\cdot\|_{L_2(Q)}, \epsilon) \le \frac{C'_{(2)}}{\underline{\mathrm{h}}^{4v+14}\epsilon^{4v+8}}.$$

**Proof** The proof is similar to Lemma 28. We first bound $\sup_Q N(\mathcal{K}^{(1)}, \|\cdot\|_{L_2(Q)}, \epsilon)$. We have

$$\begin{aligned}\omega_z^{(1)}(s) &= \mathbb{E}[K_h(s-z)K_h(Z-z)] - \mathbb{E}\left[\mathbb{U}_n(K_h(Z_i - z)K_h(Z_{i'} - z))\right] \\ &= K_h(s-z)\mathbb{E}[K_h(z-Z)] - \{\mathbb{E}[K_h(z-Z)]\}^2.\end{aligned}$$

We first study the covering number of the function class

$$\mathcal{C}_K = \{\mathbb{E}[K_h(z-Z)] = (K_h * f_Z)(z) \mid h \in [\underline{\mathrm{h}}, \overline{\mathrm{h}}], z \in (0,1)\},$$

where "$*$" denotes the convolution. Just as (145), $\mathcal{C}_K$ is also generated by convolutions. Similar to (147) and (148), we have

$$\begin{aligned}&\sup_{z_0,h}\left|\nabla_z \mathbb{E}[K_h(z-Z)]\Big|_{z=z_0}\right| \le 2\underline{\mathrm{h}}^{-1}\mathrm{TV}(K)\overline{\mathrm{f}}_Z \text{ and} \\ &\sup_{z,h_0}\left|\nabla_h \mathbb{E}[K_h(z-Z)]\Big|_{h=h_0}\right| \le 2\overline{\mathrm{f}}_Z(\underline{\mathrm{h}}^{-1}\|K\|_1 + \underline{\mathrm{h}}^{-2}\mathrm{TV}(K)).\end{aligned}$$

Therefore, following the derivation of (149), for any $z_1, z_2 \in (0,1)$, $h_1, h_2 \in [\underline{\mathrm{h}}, \overline{\mathrm{h}}]$, we have

$$\left|\mathbb{E}[K_{h_1}(z_1 - Z)] - \mathbb{E}[K_{h_2}(z_2 - Z)]\right| \le C_h \max(|z_1 - z_2|, |h_1 - h_2|),$$

where $C_h := 2\overline{\mathrm{f}}_Z[(\underline{\mathrm{h}}^{-1} + \underline{\mathrm{h}}^{-2})\mathrm{TV}(K) + \underline{\mathrm{h}}^{-1}\|K\|_1]$. From Lemma 27, we have

$$\begin{aligned}&\sup_Q N(\mathcal{C}_K, \|\cdot\|_{L_2(Q)}, \epsilon) \le C_h/\epsilon \quad \text{and} \\ &\sup_Q N\big(\{(\mathbb{E}[K_h(z-Z)])^2 \mid z, h\}, \|\cdot\|_{L_2(Q)}, \epsilon\big) \le C_h/\epsilon.\end{aligned}$$

Using the fact that the function $q(h) = 1/\sqrt{h}$ is Lipschitz on $[\underline{\mathrm{h}}, \overline{\mathrm{h}}]$ with Lipschitz constant $(\underline{\mathrm{h}})^{-3/2}/2$ together with Lemma 26, Lemma 27 and (143), we have

$$\sup_Q N(\mathcal{K}^{(1)}, \|\cdot\|_{L_2(Q)}, \epsilon) \le C\left(\frac{1}{\underline{\mathrm{h}}^2\epsilon}\right)^v \left(\frac{C_h}{\underline{\mathrm{h}}\epsilon}\right)\left(\frac{C_h}{\epsilon}\right)\left(\frac{\overline{\mathrm{h}} - \underline{\mathrm{h}}}{\underline{\mathrm{h}}^{3/2}\epsilon}\right) \le \frac{C'_{(1)}}{\underline{\mathrm{h}}^{2v+7}\epsilon^{v+3}},$$

where $C'_{(1)} := [2\overline{\mathrm{f}}_Z(2\mathrm{TV}(K) + \|K\|_1)C]^2$. Function class $\mathcal{K}^{(2)}$ contains functions in the form

$$\omega_z^{(2)}(s,t) = K_h(s-z)K_h(t-z) - \omega_z^{(1)}(s) - \omega_z^{(1)}(t) - \mathbb{E}\left[\mathbb{U}_n(K_h(Z_i - z)K_h(Z_i - z))\right].$$

By Lemma 26, it suffices to study the covering number of

$$\mathcal{C}'_K = \{K_h(s-z)K_h(t-z) \mid h \in [\underline{\mathrm{h}}, \overline{\mathrm{h}}], z \in (0,1)\}.$$

Using Lemma 26 and (143) again, we have

$$\sup_Q N(\mathcal{C}'_K, \|\cdot\|_{L_2(Q)}, \epsilon) \le C^2\left(\frac{2\|K\|_\infty}{\epsilon}\right)^{2v},$$

and therefore combining with the covering number in (153) and (151)

$$\sup_Q N(\mathcal{K}^{(2)}, \|\cdot\|_{L_2(Q)}, \epsilon) \le C^2\left(\frac{2\|K\|_\infty}{\epsilon}\right)^{2v}\left(\frac{C'_{(1)}}{\underline{\mathrm{h}}^{2v+7}\epsilon^{v+3}}\right)^2\left(\frac{C_h}{\epsilon}\right)\left(\frac{\overline{\mathrm{h}} - \underline{\mathrm{h}}}{\epsilon}\right) \le \frac{C'_{(2)}}{\underline{\mathrm{h}}^{4v+14}\epsilon^{4v+8}},$$

where $C'_{(2)} := C^2 4^v \|K\|_\infty^{2v}(C'_{(1)})^2(\overline{\mathrm{f}}_Z^2 C_\tau + 2C)^2$. This completes the proof. ∎

**Lemma 30** *Suppose $\mathbf{\Omega}(z) \in \mathcal{U}(c, M, \rho)$ for all $z \in (0,1)$. Consider the class of functions $\mathcal{J} = \{J_{z|(j,k)} \mid z \in (0,1), j,k \in [d]\}$, where $J_{z|(j,k)}$ is defined in (48). There exists positive constants $C$ and $c$ such that*

$$\sup_{Q} N(\mathcal{J}, \|\cdot\|_{L_2(Q)}, \epsilon/\sqrt{h}) \leq \left(\frac{Cd}{h\epsilon}\right)^{c}.$$

**Proof** We denote for any $u, v \in [d]$ and $z \in (0,1)$ that

$$\mathbf{\Phi}_{uv}(z; Y) = \pi \cos\left(\tau_{uv}(z)\frac{\pi}{2}\right)\sqrt{h} \cdot \left[g^{(1)}_{z|(u,v)}(Y) - \tau_{uv}(z)\omega^{(1)}_{z}(Z)\right] \tag{154}$$

and the matrix $\mathbf{\Phi}(z;Y) = [\mathbf{\Phi}_{uv}(z;Y)]_{u,v\in[d]}$. In order to bound the covering number of $\mathcal{J}$, we define a larger function class

$$\mathcal{J}' = \left\{\mathbf{\Omega}_j^T(z)\mathbf{\Phi}(x;\cdot)\mathbf{\Omega}_k(z) \mid z, x \in (0,1), j,k \in [d]\right\}.$$

Given any measure $Q$, $j,k \in [d]$ and $x_1, x_2, z_1, z_2 \in (0,1)$, we first bound the difference

$$\begin{aligned}
&\|\mathbf{\Omega}_j^T(z_1)\mathbf{\Phi}(x_1;Y)\mathbf{\Omega}_k(z_1) - \mathbf{\Omega}_j^T(z_2)\mathbf{\Phi}(x_2;Y)\mathbf{\Omega}_k(z_2)\|^2_{L_2(Q)} \\
&\leq 3\|\mathbf{\Omega}_j(z_1) - \mathbf{\Omega}_j(z_2)\|_1^2 \max_{u,v} \|\mathbf{\Phi}_{uv}(x_1;Y)\|^2_{L_2(Q)}\|\mathbf{\Omega}_k(z_1)\|_1^2 \\
&\quad + 3\|\mathbf{\Omega}_j(z_2)\|_1^2 \max_{u,v} \|\mathbf{\Phi}_{uv}(z_1;Y) - \mathbf{\Phi}_{uv}(z_2;Y)\|^2_{L_2(Q)}\|\mathbf{\Omega}_k(z_1)\|_1^2 \\
&\quad\quad + 3\|\mathbf{\Omega}_j(z_2)\|_1^2 \max_{u,v} \|\mathbf{\Phi}_{uv}(x_2;Y)\|^2_{L_2(Q)}\|\mathbf{\Omega}_k(z_1) - \mathbf{\Omega}_k(z_2)\|_1^2.
\end{aligned} \tag{155}$$

Since $\mathbf{\Omega}(z) \in \mathcal{U}(c, M, \rho)$ for any $z \in (0,1)$, we have $\sup_z \|\mathbf{\Omega}_j(z)\|_1 \leq M$. Next, using Theorem 2.5 of Stewart et al. (1990), for any $z_1, z_2 \in (0,1)$,

$$\|\mathbf{\Omega}(z_1) - \mathbf{\Omega}(z_2)\|_2 \leq \|\mathbf{\Omega}(z_1)\|_2\|\mathbf{\Omega}(z_2)(\mathbf{\Sigma}(z_1) - \mathbf{\Sigma}(z_2))\|_2.$$

Since $\mathbf{\Omega}(z) \in \mathcal{U}(c, M, \rho)$ for any $z \in (0,1)$, we further have

$$\|\mathbf{\Omega}(z_1) - \mathbf{\Omega}(z_2)\|_1 \leq \sqrt{d}\|\mathbf{\Omega}(z_1) - \mathbf{\Omega}(z_2)\|_2 \leq \rho^2\|\mathbf{\Sigma}(z_1) - \mathbf{\Sigma}(z_2)\|_2 \leq \rho^2 d^{3/2}\|\mathbf{\Sigma}(z_1) - \mathbf{\Sigma}(z_2)\|_{\max}.$$

Since $\mathbf{\Sigma}_{jk}(\cdot) \in \mathcal{H}(2, M_\sigma)$, we have

$$\begin{aligned}
\|\mathbf{\Omega}(z_1) - \mathbf{\Omega}(z_2)\|_1 &\leq \rho^2 d^{3/2}\|\mathbf{\Sigma}(z_1) - \mathbf{\Sigma}(z_2)\|_{\max} \\
&\leq \rho^2 d^{3/2}\|\mathbf{T}(z_1) - \mathbf{T}(z_2)\|_{\max} \leq \rho^2 M_\sigma d^{3/2}|z_1 - z_2|.
\end{aligned} \tag{156}$$

We next study the covering number of the function class $\mathcal{J}_{uv} = \{\mathbf{\Phi}_{uv}(z;\cdot) \mid z \in (0,1)\}$. By (84) and (94), we have

$$\max_{u,v,x} \|\mathbf{\Phi}_{uv}(x;Y)\|^2_{L_2(Q)} \leq \max_{u,v,x} \|\mathbf{\Phi}_{uv}(x;Y)\|^2_{\infty} \leq Ch^{-1}. \tag{157}$$

According to the definition in (154), $\mathbf{\Phi}_{uv}(z;\cdot)$ is obtained from products and summations of functions with known covering numbers, quantified in Lemmas 28 and 29. By Lemmas 27 and 26 and fixing the bandwidth $h = \underline{\mathrm{h}} = \overline{\mathrm{h}}$, $\sup_Q N(\mathcal{J}_{uv}, \|\cdot\|_{L_2(Q)}, \epsilon) \leq C/(h\epsilon)^{v_1}$ for any $u, v \in [d]$. Notice that the construction of covering sets in the proofs of Lemmas 28,

and 29 is independent to the indices $j, k$. Therefore, we can construct a set $N_{(2)} \subset (0,1)$ with $|N_{(2)}| \leq C/(h\epsilon)^c$ such that for any $x \in (0,1)$, there exists a $x_\ell \in N_{(2)}$ with

$$\max_{u,v} \|\boldsymbol{\Phi}_{uv}(x;Y) - \boldsymbol{\Phi}_{uv}(x_\ell;Y)\|_{L_2(Q)} \leq \epsilon. \tag{158}$$

With this, we construct the covering set for $\mathcal{J}'$ as

$$N_{(3)} = N_{(2)} \times \{\ell\epsilon\sqrt{h} | \ell = 0, \ldots, \lfloor 1/(\epsilon\sqrt{h}) \rfloor\}.$$

For any $(x,z) \in (0,1)^2$, we select $(x_\ell, z_\ell) \in N_{(3)}$ such that (158) holds and $|z - z_\ell| \leq \epsilon\sqrt{h}$. Therefore, by (155), (156) and (157), we have

$$\|\boldsymbol{\Omega}_j^T(z)\boldsymbol{\Phi}(x;Y)\boldsymbol{\Omega}_k(z) - \boldsymbol{\Omega}_j^T(z_\ell)\boldsymbol{\Phi}(x_\ell;Y)\boldsymbol{\Omega}_k(z_\ell)\|_{L_2(Q)}^2 \leq Cd^3M^4\epsilon^2$$

and $\sup_Q N(\mathcal{J}', \|\cdot\|_{L_2(Q)}, d^{3/2}M^2\epsilon) \leq d^2|N_{(3)}| = (Cd/(h\epsilon))^c$.

■

## Appendix I. Some Useful Results

**Lemma 31** *Let $(Y_1, Y_2, Y_3, Y_4)^T \sim N_4(\mathbf{0}, \mathbf{K})$ with $\mathbf{K} = [K_{ab}]_{ab}$. We then have*

$$\mathbb{E}\left[\operatorname{sign}(Y_1 - Y_2)\operatorname{sign}(Y_3 - Y_4)\right] = \frac{2}{\pi}\arcsin\left(\frac{K_{13} + K_{24} - K_{23} - K_{14}}{\sqrt{(K_{11} + K_{22} - 2K_{12})(K_{33} + K_{44} - 2K_{34})}}\right).$$

**Proof** Observe that $(Y_1 - Y_2, Y_3 - Y_4)^T$ is distributed according to a bivariate Gaussian distribution with mean zero and Pearson correlation coefficient

$$\operatorname{Corr}\left[\mathrm{Y}_1 - \mathrm{Y}_2, \mathrm{Y}_3 - \mathrm{Y}_4\right] = \frac{\mathrm{K}_{13} + \mathrm{K}_{24} - \mathrm{K}_{23} - \mathrm{K}_{14}}{\sqrt{(\mathrm{K}_{11} + \mathrm{K}_{22} - 2\mathrm{K}_{12})(\mathrm{K}_{33} + \mathrm{K}_{44} - 2\mathrm{K}_{34})}}.$$

The result follows directly from the correspondence between Pearson correlation and Kendall's tau (Fang et al., 1990). ■

**Corollary 32** *Let $(\boldsymbol{X}_1, Z_1), (\boldsymbol{X}_2, Z_2)$ be independently distributed according to the model in (1). Then we have*

$$\mathbb{E}\left[\operatorname{sign}(X_{1a} - x_{2a})\operatorname{sign}(X_{1b} - x_{2b}) \mid Z_1 = z_1, Z_2 = z_2\right] = \frac{2}{\pi}\arcsin\left(\frac{\boldsymbol{\Sigma}_{ab}(z_1) + \boldsymbol{\Sigma}_{ab}(z_2)}{2}\right).$$

**Proof** Follows directly from Lemma 31 by observing that

$$\operatorname{sign}(X_{1a} - X_{2a}) = \operatorname{sign}(f(X_{1a}) - f(X_{2a})),$$

since $f$ is monotone, and using the fact that $f(\boldsymbol{X}_i)$ follows a Gaussian distribution. ■

Let $H : S^2 \mapsto \mathbb{R}$ be a symmetric kernel function. In the setting of our paper, we have $S^2 = \mathbb{R}^2 \times (0, 1)$. A kernel is completely degenerate if

$$\mathbb{E}[H(Y_1, Y_2) \mid Y_2] = 0.$$

$U$-statistic based on the kernel $H$ is called degenerate of order 1. See, for example, Serfling (2001).

**Theorem 33 (Theorem 2, Major (2006))** *Let $\{Y_i\}_{i\in[n]}$ be independent and identically distributed random variables on a probability space $(S, \mathcal{S}, \mu)$. Let $\mathcal{F}$ be a separable space (with respect to $\mu$) of $\mathcal{S}$-measurable $\mu$-degenerate kernel functions that satisfies*

$$N(\epsilon, \mathcal{F}, L_2(\mu)) \leq A\epsilon^{-v}, \quad \text{for all } 1 \geq \epsilon > 0,$$

*where $A$ and $v$ are some fixed constants. Furthermore, we assume that the envelope function is bounded by 1, that is,*

$$\sup_{y_1,y_2} |H(y_1, y_2)| \leq 1, \quad \text{for all } H \in \mathcal{F} \text{ and} \tag{159}$$

$$\sup_{H\in\mathcal{F}} \mathbb{E}\left[H^2(Y_1, Y_2)\right] \leq \sigma^2$$

*for some $0 < \sigma \leq 1$. Then there exist constants $C_1, C_2, C_3$ that depend only on $A$ and $v$, such that*

$$\mu\left[\sup_{H\in\mathcal{F}} \left|\frac{1}{n}\sum_{i\neq i'} H(Y_i, Y_{i'})\right| \geq t\right] \leq C_1 \exp\left(-C_2 \frac{t}{\sigma}\right) \tag{160}$$

*for all $t$ such that*

$$n\sigma^2 \geq \frac{t}{\sigma} \geq C_3\Big(v + \frac{\log A}{\log n}\Big)^{3/2} \log\left(\frac{2}{\sigma}\right). \tag{161}$$

**Lemma 34 (Lemma A.1, van de Geer 2008)** *Let $X_1, \ldots, X_n$ be independent random variables. The $\gamma_1, \ldots, \gamma_d$ be real-valued bounded functions satisfying*

$$\mathbb{E}[\gamma_j(Z_i)] = 0, \text{ for any } i \in [n], j \in [d]; \quad \|\gamma_k\|_\infty \leq \eta_n \quad \text{and} \quad \max_{j\in[d]} \frac{1}{n}\sum_{i=1}^{n} \mathbb{E}\gamma_j^2(Z_i) \leq \tau_n^2.$$

*We have*

$$\mathbb{E}\left[\max_{j\in[d]} \left|\frac{1}{n}\sum_{i=1}^{n} \gamma_j(Z_i)\right|\right] \leq \sqrt{\frac{2\tau_n^2 \log(2d)}{n}} + \frac{\eta_n \log(2m)}{n}.$$

## Appendix J. Notation Table

The following table collects notation used in the paper, their meanings and where they first appear.

| Notation | Meaning | Page No. |
|---|---|---|
| $\boldsymbol{X}$ | $d$-dimensional nonparanormal random vector | 2 |
| $Z$ | index variable | 2 |
| $Y$ | $Y = (\boldsymbol{X}, Z)$ denotes the full data variable | 2 |
| $\boldsymbol{\Sigma}(z)$ | true correlation matrix | 2 |
| $f$ | marginal transform | 2 |
| $\boldsymbol{\Omega}(z)$ | inverse correlation matrix | 2 |
| $[n], [d]$ | discrete sets $[n] = \{1, \ldots, n\}$, $[d] = \{1, \ldots, d\}$ | 5 |
| $\mathbf{A} \circ \mathbf{B}$ | Hadamard product $(\mathbf{A} \circ \mathbf{B})_{jk} = \mathbf{A}_{jk} \cdot \mathbf{B}_{jk}$ | 5 |
| $f * g$ | convolution $(f * g)(t) = \int f(t-x)g(x)dx$ | 5 |
| $\mathrm{TV}(f)$ | total variation of $f$ | 5 |
| $\vee$, $\wedge$ | $a \vee b = \max(a, b)$ and $a \wedge b = \min(a, b)$ | 5 |
| $c, C, c_1, C_1, \ldots$ | universal constants whose values may change from line to line | 5 |
| $\rightsquigarrow$ | converge in distribution | 5 |
| $\xrightarrow{P}$ | converge in probability | 5 |
| $\mathbb{1}(\cdot)$ | indicator function | 5 |
| $\mathbb{E}_n[f]$ | $= n^{-1} \sum_{i \in [n]} f(X_i)$ | 5 |
| $\mathbb{G}_n[f]$ | $= n^{-1/2} \sum_{i \in [n]} (f(X_i) - \mathbb{E}[f(X_i)])$ | 5 |
| $\mathbb{U}_n[H]$ | $= [n(n-1)]^{-1} \sum_{i \neq i'} H(X_i, X_{i'})$ | 5 |
| $G^*(z)$ | true graph | 6 |
| $E^*(z)$ | true edge set | 6 |
| $\mathbf{e}_j$ | the $j$-th canonical basis in $\mathbb{R}^d$ | 6 |
| $\widehat{\boldsymbol{\Omega}}(z)$ | a generic inverse correlation estimator | 6 |
| $\widehat{\boldsymbol{\Omega}}^{\mathrm{CLIME}}(z)$ | CLIME inverse correlation estimator | 6 |
| $\widehat{\boldsymbol{\Omega}}^{\mathrm{CCLIME}}(z)$ | calibrated CLIME inverse correlation estimator | 6 |
| $\lambda$ | $\\|\cdot\\|_\infty$-tuning parameter calibrated CLIME | 6 |
| $\gamma$ | $\\|\cdot\\|_1$-tuning parameter calibrated CLIME | 6 |
| $\tau_{jk}(z)$ | Kendall's tau correlation | 7 |
| $\widehat{\tau}_{jk}(z)$ | Kendall's tau estimator | 7 |
| $h$ | bandwidth | 7 |
| $K_h$ | The rescaled kernel $K_h(\cdot) = h^{-1}K(\cdot/h)$ | 7 |
| $\omega_z$ | double kernel function $\omega_z(Z_i, Z_{i'}) = K_h\left(Z_i - z\right) K_h\left(Z_{i'} - z\right)$ | 7 |
| $\widehat{\boldsymbol{\Sigma}}(z)$ | correlation matrix estimator | 7 |
| $\widehat{\mathbf{T}}(z)$ | Kendall's tau matrix estimator | 7 |
| $[z_L, z_U]$ | interval for uniform edge test | 8 |
| $\widehat{S}_{z\mid(j,k)}(\boldsymbol{\beta})$ | $= \widehat{\boldsymbol{\Omega}}_j^T(z)\big(\widehat{\boldsymbol{\Sigma}}(z)\boldsymbol{\beta} - \mathbf{e}_k\big)$ | 8 |
| $f_Z$ | the density of $Z$ | 9 |
| $\sigma_{jk}^2$ | variance of the score statistic | 9 |
| $\boldsymbol{\Theta}_z$ | random matrix for the definition of $\sigma_{jk}^2$ | 9 |
| $\tau_{jk}^{(1)}(y)$ | $= \sqrt{h} \cdot \mathbb{E}\Big[K_h\left(z - z_0\right) K_h\left(Z - z_0\right) \big(\mathrm{sign}(X_j - x_j)\, \mathrm{sign}(X_k - x_k) - \tau_{jk}(z_0)\big)\Big]$ | 9 |
| $\xi_i$ | i.i.d. sample of standard normal $N(0, 1)$ | 10 |
| $\widehat{\tau}_{jk}^B(z)$ | bootstrap Kendall's tau estimator | 10 |
| $\widehat{\boldsymbol{\Sigma}}_{jk}^B(z)$ | bootstrap correlation estimator | 10 |
| $\mathcal{H}(\gamma, L)$ | the Hölder class on $(0, 1)$ | 12 |
| $\underline{\mathrm{f}}_Z, \overline{\mathrm{f}}_Z$ | lower and upper bounds of $f_Z$ | 12 |
| $M_\sigma$ | the Hölder constant such that $\boldsymbol{\Sigma}_{jk}(\cdot) \in \mathcal{H}(2, M_\sigma)$ | 12 |
| $r_{1n}, r_{2n}, r_{3n}$ | the statistical rates of $\widehat{\boldsymbol{\Sigma}}(z)$ and $\widehat{\boldsymbol{\Omega}}(z)$ | 12 |
| $M$ | upper bound of column $\ell_1$-norm of $\boldsymbol{\Omega}(z)$ | 13 |
| $\rho$ | maximum eigenvalue of $\boldsymbol{\Omega}(z)$ | 13 |
| $\mathcal{U}_s(M, \rho)$ | the class of matrices $\mathbf{\Omega} \succ 1/\rho, \lVert\mathbf{\Omega}\rVert_2 \leq \rho, \max_{j \in [d]} \lVert\mathbf{\Omega}_j\rVert_0 \leq s, \lVert\mathbf{\Omega}\rVert_1 \leq M$ | 13 |
| $h_l, h_u$ | lower and upper bounds of bandwidth | 15 |
| $C_{\mathbf{\Sigma}}$ | universal constant in the rate of $\widehat{\mathbf{\Sigma}}(z)$ | 16 |
| $\overline{\mathcal{U}}_s(M, \rho, L)$ | time-varying matrix class $\mathbf{\Omega}(z) \in \mathcal{U}_s(M, \rho)$ | 17 |
| $g_{z\mid(j,k)}(y_i, y_{i'})$ | $= \omega_z(z_i, z_{i'}) \operatorname{sign}(x_{ia} - x_{i'a}) \operatorname{sign}(x_{ib} - x_{i'b})$ | 27 |
| $g^{(1)}_{z\mid(j,k)}(y)$ | $= \mathbb{E}[g_{z\mid(j,k)}(y, Y)] - \mathbb{E}\left[\mathbb{U}_n[g_{z\mid(j,k)}]\right]$ | 27 |
| $g^{(2)}_{z\mid(j,k)}(y_1, y_2)$ | $= g_{z\mid(j,k)}(y_1, y_2) - g^{(1)}_{z\mid(j,k)}(y_1) - g^{(1)}_{z\mid(j,k)}(y_2) - \mathbb{E}\left[\mathbb{U}_n[g_{z\mid(j,k)}]\right]$ | 27 |
| $\omega^{(1)}_z(s)$ | $= \mathbb{E}[\omega_z(s, Z)] - \mathbb{E}\left[\mathbb{U}_n[\omega_z]\right]$ | 27 |
| $\omega^{(2)}_z(s, t)$ | $= \omega_z(s, t) - \omega^{(1)}_z(s) - \omega^{(1)}_z(t) - \mathbb{E}\left[\mathbb{U}_n[\omega_z]\right]$ | 27 |
| $\mathbb{G}^{\xi}_n[f]$ | $= n^{-1/2} \sum_{i=1}^{n} f(X_i) \cdot \xi_i$ | 26 |
| $\mathbb{P}_\xi$, $\mathbb{E}_\xi$ | $\mathbb{P}_\xi(\cdot) := \mathbb{P}(\cdot \mid \{Y_i\}_{i \in [n]})$ and $\mathbb{E}_\xi[\cdot] := \mathbb{E}[\cdot \mid \{Y_i\}_{i \in [n]}]$ | 26 |
| $J_{z\mid(j,k)}(y')$ | $= \sum_{u,v \in [d]} \mathbf{\Omega}_{ju}(z) \mathbf{\Omega}_{kv}(z) \pi \cos\left(\tau_{uv}(z) \frac{\pi}{2}\right) \sqrt{h} \cdot \left[g^{(1)}_{z\mid(u,v)}(y') - \tau_{uv}(z) \omega^{(1)}_z(z')\right]$ | 27 |
| $u_n[H]$ | $= \sqrt{n} \cdot \left(\mathbb{U}_n[H] - \mathbb{E}[\mathbb{U}_n[H]]\right)$ | 28 |
| $N(\mathcal{F}, d, \epsilon)$ | covering number of function class $\mathcal{F}$ of $\epsilon$-ball in metric $d$ | 61 |